\documentclass{article}

\PassOptionsToPackage{numbers, compress}{natbib}
\usepackage[preprint]{neurips_2026}

\usepackage[utf8]{inputenc} % allow utf-8 input
\usepackage[T1]{fontenc}    % use 8-bit T1 fonts
\usepackage{hyperref}       % hyperlinks
\usepackage{url}            % simple URL typesetting
\usepackage{booktabs}       % professional-quality tables
\usepackage{amsfonts}       % blackboard math symbols
\usepackage{nicefrac}       % compact symbols for 1/2, etc.
\usepackage{microtype}      % microtypography
\usepackage{xcolor}         % colors
\usepackage{graphicx}
\usepackage{subcaption}
\usepackage{booktabs} % for professional tables
\usepackage{dblfloatfix}

\usepackage{amsmath}
\usepackage{amssymb}
\usepackage{mathtools}
\usepackage{amsthm}
\usepackage{makecell}
\usepackage{multirow}
\usepackage[utf8]{inputenc}
\usepackage[T1]{fontenc}
\usepackage{wrapfig}
\usepackage{subcaption}

\theoremstyle{plain}
\newtheorem{theorem}{Theorem}[section]
\newtheorem{proposition}[theorem]{Proposition}

\newtheorem*{lemma*}{Lemma} 

\theoremstyle{definition}

\theoremstyle{remark}

\usepackage{kotex} % 한글 패키지
\usepackage[textsize=tiny]{todonotes}

\title{Focusing Influence Mechanism for\\ Multi-Agent Reinforcement Learning}

\author{%
  Yisak Park \quad
  Sunwoo Lee \quad
  Seungyul Han\thanks{Correspondence to: Seungyul Han} \\
  Graduate School of Artificial Intelligence \\
  UNIST \\
  Ulsan, South Korea 44919 \\
  \texttt{\{isaac1018,sunwoolee,syhan\}@unist.ac.kr}
}

\begin{document}

\maketitle

\begin{abstract}
Cooperative multi-agent reinforcement learning (MARL) under sparse rewards remains fundamentally challenging because agents often fail to concentrate their influence, leading to insufficiently coordinated exploration. To address this, we propose the Focusing Influence Mechanism (FIM), a framework that encourages agents to focus their influence on under-explored parts of the state space through an entropy-based criterion, while leveraging eligibility traces to enable multiple agents to consistently align and sustain their influence on the same parts of the state space when beneficial, thereby promoting coordinated and persistent joint behavior. By emphasizing under-explored regions of the state space, FIM facilitates more efficient and structured exploration even under extremely sparse rewards. Across diverse MARL benchmarks, FIM consistently improves cooperative performance over strong baselines.
\end{abstract}

\section{Introduction}
\label{sec:introduction}

Cooperative multi-agent reinforcement learning (MARL) is a powerful framework for sequential decision-making problems involving multiple agents, with applications in autonomous driving~\cite{shalev2016safe}, multi-robot coordination~\cite{perrusquia2021multi}, and real-time strategy games~\cite{vinyals2019grandmaster}. These environments involve partial observability, making decentralized partially observable Markov decision processes (Dec-POMDPs)~\cite{oliehoek2016concise} a natural modeling choice. To address the challenges arising from limited observability, the centralized training with decentralized execution (CTDE)~\cite{oliehoek2008optimal,yu2022mappo} paradigm has been widely adopted. In CTDE, policies are trained using access to the global state and all agents’ observations, but are executed using only local observations. Prominent CTDE methods such as VDN~\cite{sunehag2018value}, QMIX~\cite{rashid2018qmix}, and QPLEX~\cite{wangqplex} leverage value decomposition to promote coordinated policy learning.

Despite the success of CTDE, many CTDE-based MARL methods struggle in sparse-reward settings where effective exploration is essential~\cite{jaques2019social,Wang2020Influence-Based,liu2021cooperative}. Several approaches have been proposed to address this challenge, including maximizing mutual influence between agents~\cite{Wang2020Influence-Based}, prioritizing under-visited but important states~\cite{zheng2021episodic}, and diversifying trajectory~\cite{li2021celebrating}. While promising, we observe that these methods can still fail in challenging sparse-reward environments, particularly under extremely sparse rewards, because they do not explicitly guide how agents should coordinate and allocate their influence to escape local optima. Liu et al.~\cite{liu2023lazy} improves exploration efficiency by concentrating an agent's influence on an external state; however, selecting external-state dimensions relies on human knowledge and prior task information, which can be difficult to scale as the dimensionality of the external state increases or when such information is unavailable. 

To address these issues, we identify two factors that make sparse-reward MARL difficult. First, even when agents can influence the environment, their influence tends to remain scattered across multiple targets without converging on shared targets, leading to diluted and ineffective exploration. Second, agents lack a principled criterion for prioritizing parts of the state space that are under-explored relative to the current behavior policy. Motivated by these observations, we propose the \textbf{Focusing Influence Mechanism (FIM)}, a MARL framework that addresses both challenges through two complementary components: (i) an \textit{agent focusing influence (AFI)} module that concentrates multiple agents' influence on shared targets by accumulating collective influence via eligibility traces and amplifying intrinsic rewards for components that receive consistent joint attention, thereby enabling persistent coordinated behavior, and (ii) a \textit{state focusing influence (SFI)} module that uses an entropy-based criterion to prioritize under-explored parts of the state space, guiding AFI toward more meaningful targets. As a result, FIM promotes more efficient and structured exploration in challenging sparse-reward environments, and extensive experiments across diverse MARL benchmarks demonstrate improved performance over existing methods.

\section{Related Works}

\paragraph{Intrinsic Motivation in Sparse Reward MARL.} 
Intrinsic motivation is widely used to promote exploration in sparse-reward environments. Curiosity-driven objectives seek novel states~\cite{iqbal2019coordinated, zheng2021episodic, li2023two, zhang2023self, yang2024cmbe, xu2024population,tao2024multi}, while trajectory diversity methods expand state-space coverage~\cite{zhang2023expode, li2025toward, li2025learning}. Committed exploration is induced via a shared latent variable~\cite{mahajan2019maven}, and spatial formation reduce redundant exploration~\cite{jo2024fox}. Subgoal-based methods decompose tasks into smaller objectives~\cite{tang2018hierarchical, jeon2022maser}. Exploration can also be focused in low-dimensional subspaces~\cite{liu2021cooperative, xu2023subspace, he2024saeir}, and expectation alignment enables agents to adapt to anticipated peer behaviors~\cite{ma2022elign}.

\vspace{-1em}
\paragraph{Counterfactual Reasoning Based Credit Assignment.}
Counterfactual reasoning enables credit assignment by measuring each agent’s contribution to the team reward. COMA estimates individual action advantages using counterfactual baselines~\cite{foerster2018counterfactual, cohen2021use, wang2021towards, hoppe2024global}, while predictive counterfactual models support value decomposition~\cite{zhou2022pac, chai2024aligning}. Shapley value–based methods assign local credit by marginalizing individual contributions to the global reward~\cite{wang2020shapley, li2021shapley, wang2022shaq}. In offline settings, counterfactual conservatism~\cite{shao2023counterfactual} and sample averaging~\cite{ma2023learning} improve stability. Counterfactual reasoning also aids in identifying important agents~\cite{chen2025understanding} and salient states~\cite{cheng2023statemask}.

\vspace{-1em}
\paragraph{Influence-Driven Coordination.}
Influence-based methods aim to promote coordination by inducing causally significant changes. Social influence quantifies how an agent’s actions affect the behaviors of its teammates~\cite{jaques2019social,li2022pmic,hou2025cooperative} and guide communication decisions~\cite{ding2020learning}. Opponent modeling enables agents to influence policy updates of others~\cite{foerster2018learning,letcher2018stable,xie2021learning,kim2022influencing}. Influence-aware exploration affects future dynamics~\cite{Wang2020Influence-Based,liu2024imagine} or induces novel observations~\cite{jiang2024settling}. Influence has been extended to incentivize beneficial behaviors in others~\cite{yang2020learning}, discourage undesirable actions~\cite{schmid2021learning}, or shape the returns of other agents~\cite{zhou2024reciprocal}, as well as to affect external states~\cite{liu2023lazy} or latent representations of the environment~\cite{li2024individual}. Our work builds on this line by aligning and concentrating influence across agents on key parts of the state space, enabling more persistent and structured coordination.

\section{Preliminary}
\label{sec:preliminary}

\paragraph{Decentralized POMDP and CTDE Setup.} 
In MARL, the environment is typically modeled as a Dec-POMDP~\cite{oliehoek2016concise}, defined by the tuple $\langle \mathcal{N}, S, A, P, R, O, \mathcal{O}, \gamma \rangle$, where $\mathcal{N}$ is a set of $n$ agents, $S$ is the global state space,  $A = A^0 \times \cdots \times A^{n-1}$ is the joint action space, and $\gamma$ is the discount factor. At each timestep $t$, each agent $i \in \mathcal{N}$ receives a local observation $o^i_t = \mathcal{O}(s_t, i)$ and chooses an action $a^i_t$ from its policy $\pi^i$, based on its trajectory $\tau^i_t = (o^i_0, a^i_0, \ldots, o^i_t)$. The state $s_t$ is defined as a $D$-dimensional vector, i.e., $s_t = (s_t^0,\cdots,s_t^{D-1})$, and for given $(s_t,\mathbf{a}_t)$ pair, the environment transitions to $s_{t+1} \sim P(\cdot \mid s_t, \mathbf{a}_t)$ and returns a shared reward $r_t = R(s_t, \mathbf{a}_t)$. The goal is to learn a joint policy $\boldsymbol{\pi} = \prod_{i=1}^n \pi^i$ that maximizes the expected return $\sum_{t=0}^{\infty} \gamma^tr_t$. In this paper, we adopt the CTDE paradigm~\cite{rashid2018qmix}, where agents are trained using global state to optimize a total value function $Q^{\text{tot}}$, while each agent executes actions based solely on local observations during deployment. 
\vspace{-0.6em}
\paragraph{Credit Assignment via Counterfactual Reasoning.}  
In the CTDE paradigm, credit assignment mechanisms~\cite{rashid2018qmix,shao2023counterfactual, liu2023lazy} estimate each agent’s contribution to team performance, supporting not only the optimization of a global value function but also promoting effective exploration~\cite{li2021celebrating}, information sharing~\cite{jo2024fox}, and communication~\cite{Wang2020Learning}. A widely adopted technique is counterfactual reasoning~\cite{foerster2018counterfactual, shao2023counterfactual, liu2023lazy}, which quantifies causal influence by comparing the actual outcome to a counterfactual one where only an individual agent’s action is replaced. COMA~\cite{foerster2018counterfactual}, for example, defines credit for agent $i$ as:
\begin{equation} \label{eq:credit}
\mathrm{credit}^i_t = f(s_t, \boldsymbol{\tau}_t, \mathbf{a}_t) - \mathbb{E}_{a^i_t \sim \pi^i(\cdot \mid o^i_t)
} \left[ f(s_t, \boldsymbol{\tau}_t, a^i_t, \mathbf{a}^{-i}_t) \right],
\end{equation}
where $f = Q^{tot}$. This formulation has been leveraged not only for advantage estimation but also for shaping exploration and coordination via intrinsic rewards.
\vspace{-0.6em}
\paragraph{Eligibility Trace.}  
Eligibility traces are used to implement TD($\lambda$) online by propagating the current TD error to future timesteps for value updates~\cite{sutton2018reinforcement}. At each timestep $t$, the trace $e_t(s)$ is updated as: \begin{equation} e_t(s) = \begin{cases} \gamma \lambda e_{t-1}(s) + 1, & \text{if } s = s_t, \\ \gamma \lambda e_{t-1}(s), & \text{otherwise}, \end{cases} \end{equation} where $\lambda$ is the decay factor. This mechanism accumulates eligibility for recently visited states and decays it over time, thereby concentrating value updates on recently visited states. In this work, we adapt this concept to sustain coordinated influence, encouraging multiple agents to persistently focus on under-explored state dimensions.

\vspace{-0.5em}
\section{Methodology}
\label{sec:method}

\subsection{The Need for Advanced Focusing Influence}
\label{subsec:motiv}

\begin{figure}[!h]
  \centering
  \begin{subfigure}[T]{0.18\linewidth}
    \centering
    \includegraphics[width=\linewidth]{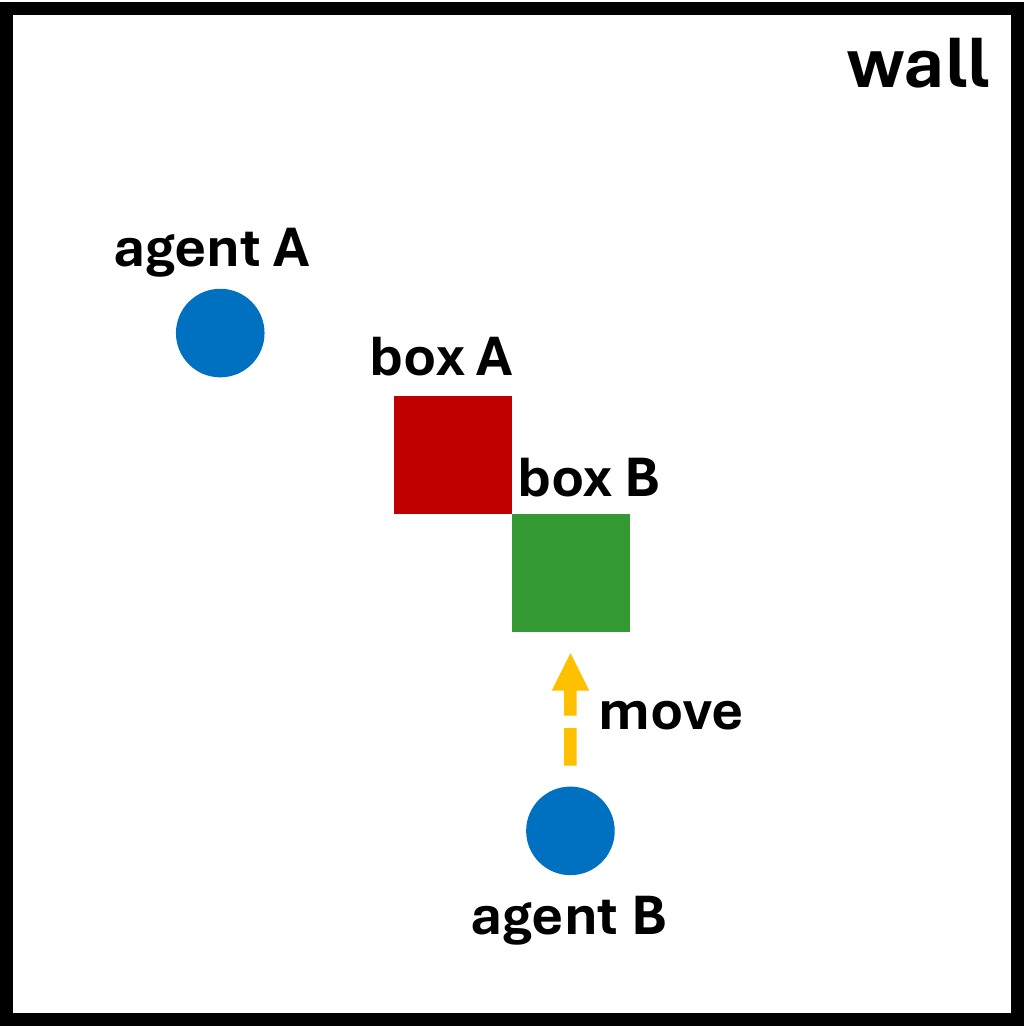}
    \caption{Push-2-Box}
  \end{subfigure}
  \hfill
  \begin{subfigure}[T]{0.18\linewidth}
    \centering
    \includegraphics[width=\linewidth]{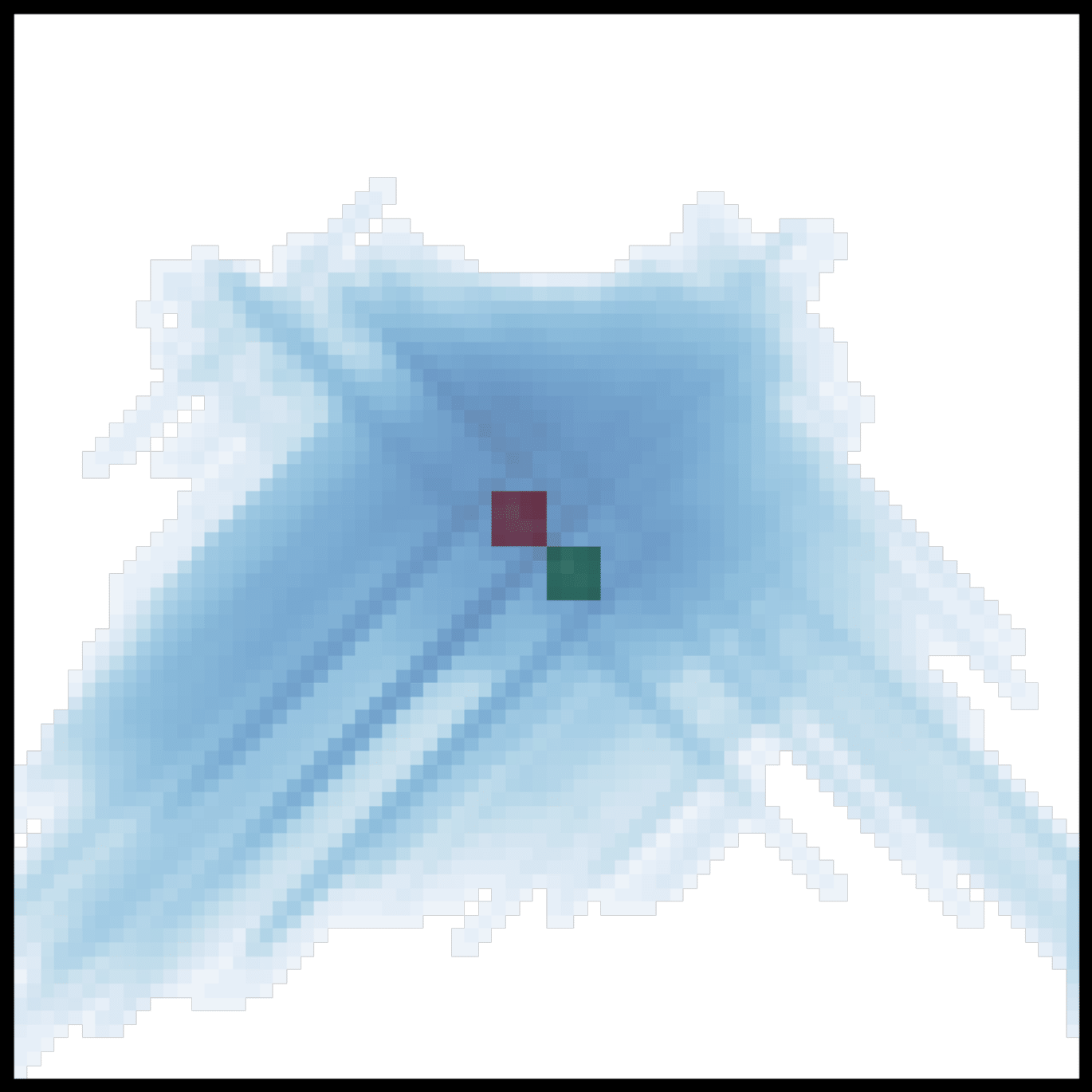}
    \caption{Vanilla QMIX}
  \end{subfigure}
  \hfill
  \begin{subfigure}[T]{0.18\linewidth}
    \centering
    \includegraphics[width=\linewidth]{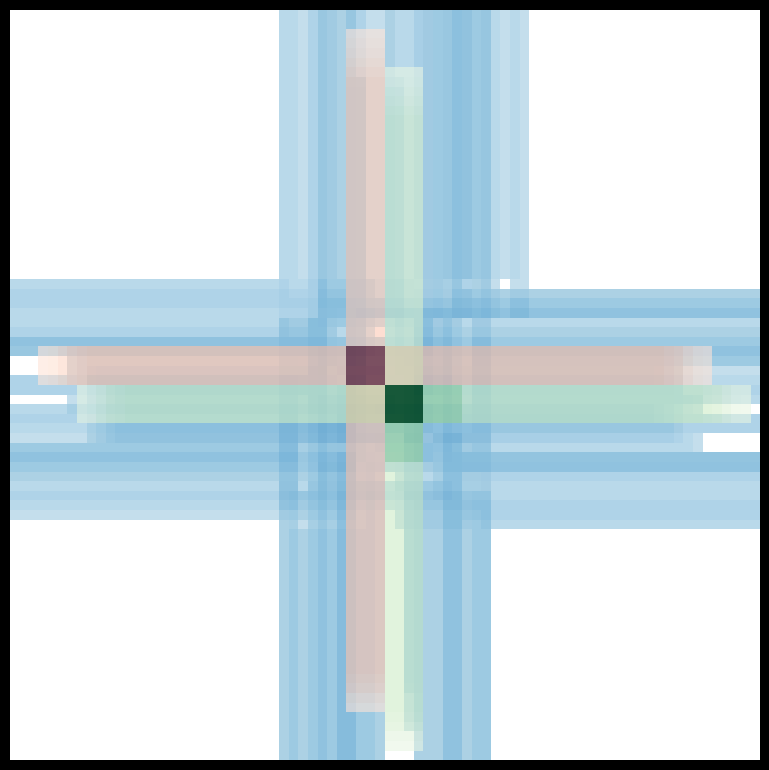}
    \caption{QMIX with AFI}
  \end{subfigure}
  \hfill
  \begin{subfigure}[T]{0.22\linewidth}
    \centering
    \includegraphics[width=0.83\linewidth]{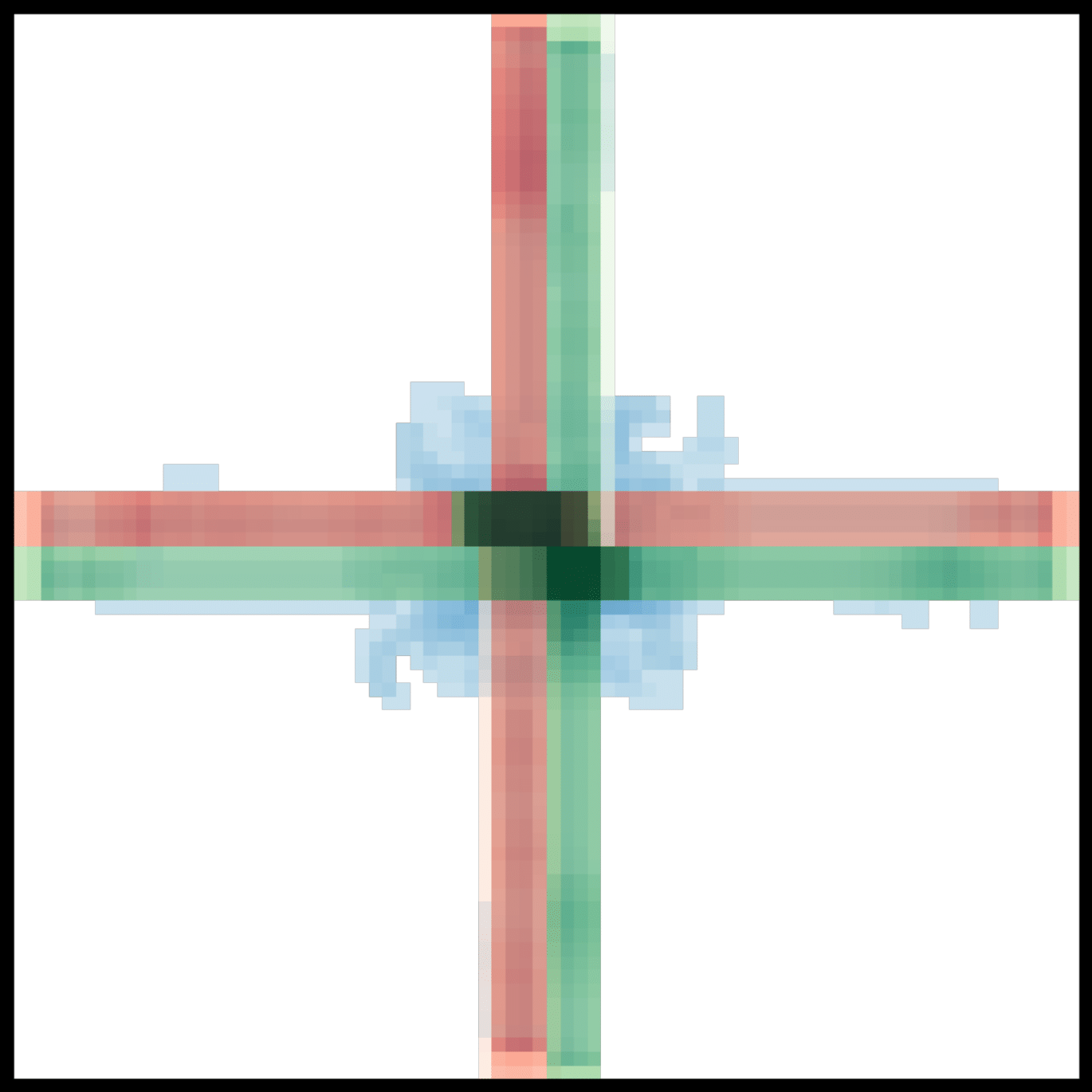}
    \caption{QMIX with AFI and SFI (proposed FIM)}
  \end{subfigure}
  \caption{Comparative results in the Push-2-Box environment: (a) shows an enlarged view of the environment, and (b–d) show average visitation counts of two agents (blue) and two boxes (Box A: red, Box B: green) over 3M timesteps across 100 seeds. Darker areas indicate more frequent visits.}
  \label{fig:pushbox}
\end{figure}

In cooperative MARL, agents must often discover coordinated behaviors through exploration, particularly in tasks where progress depends on cooperative joint actions~\cite{jaques2019social,Wang2020Influence-Based,liu2021cooperative}. While each agent individually possesses the capacity to influence the environment, sparse rewards provide insufficient learning signal to coordinate their influences. As a result, agents act in an uncoordinated manner with scattered, non-cooperative influence, failing to converge on shared targets. To illustrate, we consider the Push-2-Box environment~\cite{Wang2020Influence-Based} in Fig.~\ref{fig:pushbox}(a), which involves two agents and two boxes. The task requires both agents to jointly push a single box to a wall, as individual pushing produces negligible movement. Rewards are provided only when a box reaches the wall, and thus success requires persistent, synchronized pushing of the same box. Under such reward sparsity, standard CTDE methods often fail to discover the need for coordinated pushing during training, resulting in limited variation in the box position and task failure. Fig.~\ref{fig:pushbox}(b) illustrates this failure mode for QMIX, where scattered exploration and weak coordination prevent success.

These observations motivate the need to (i) focus multi-agent influence on shared targets, and (ii) prioritize under-explored state dimensions under the current behavior policy. To this end, we propose the \textbf{Focusing Influence Mechanism (FIM)}, which promotes cooperation through two components: \textit{agent focusing influence (AFI)} and \textit{state focusing influence (SFI)}. AFI encourages agents to align their influence on the same targets, for example by focusing on the same box, by accumulating collective influence via eligibility traces and amplifying intrinsic rewards for dimensions associated with consistent joint attention, thereby driving coordinated behavior. As shown in Fig.~\ref{fig:pushbox}(c), incorporating AFI into QMIX increases collective influence on shared targets; however, since AFI operates across all state dimensions, this influence can be spread too broadly, leaving insufficient effort on the dimensions that truly require coordination, such as box positions. SFI addresses this by weighting state dimensions using an entropy-based criterion, assigning higher priority to under-explored ones. Fig.~\ref{fig:pushbox}(d) shows that combining AFI and SFI further concentrates joint influence on box positions, enabling successful task completion. Several prior methods also incorporate state influence or exploration objectives~\cite{li2021celebrating,liu2023lazy,jo2024fox}. In Section~\ref{sec:exp}, we compare these methods on Push-2-Box and show that, despite the simplicity of the task, they fail to solve it reliably under sparse rewards, whereas FIM succeeds by providing the coordination and persistence required in this setting.

\subsection{Agent Focusing Influence}
\label{subsec:afi}

\begin{figure}[!t]
  \centering
  \begin{subfigure}[T]{0.22\linewidth}
    \centering
    \includegraphics[width=\linewidth]{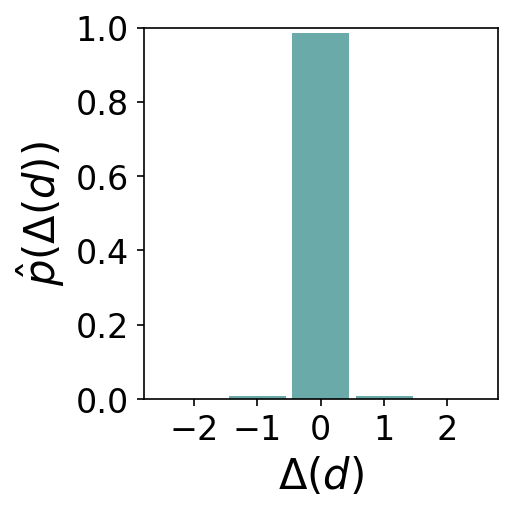}
    \caption{Box A (QMIX)}
  \end{subfigure}
  \hfill
  \begin{subfigure}[T]{0.22\linewidth}
    \centering
    \includegraphics[width=\linewidth]{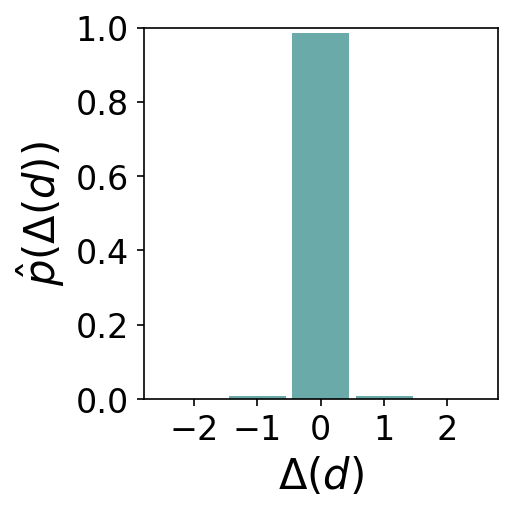}
    \caption{Box B (QMIX)}
  \end{subfigure}
  \hfill
  \begin{subfigure}[T]{0.22\linewidth}
    \centering
    \includegraphics[width=\linewidth]{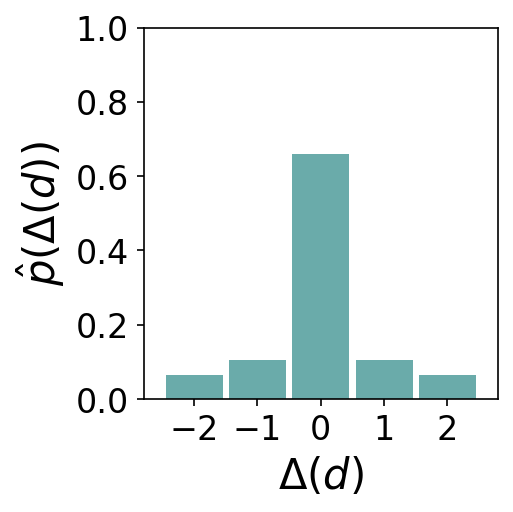}
    \caption{Box A (with AFI)}
  \end{subfigure}
  \hfill
  \begin{subfigure}[T]{0.22\linewidth}
    \centering
    \includegraphics[width=\linewidth]{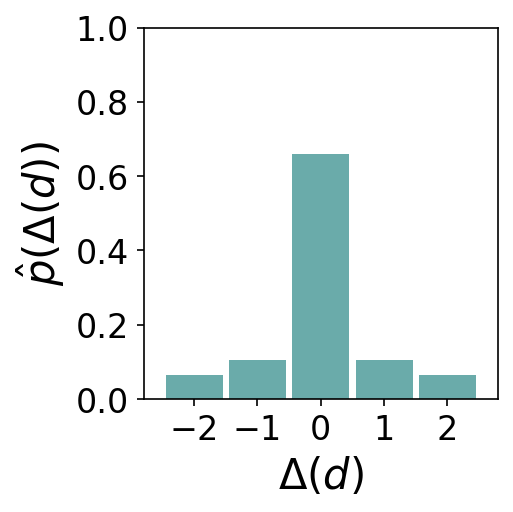}
    \caption{Box B (with AFI)}
  \end{subfigure}
  \caption{Empirical distribution $\hat{p}$ of temporal changes $\Delta(d)$ for box positions in Push-2-Box. (a–b): Without AFI. (c–d): With AFI.}
  \label{fig:pushbox-dist-change}
\end{figure}

To enable coordinated behavior, agents must not only act but also exert \emph{aligned influence} on shared state dimensions. We therefore quantify how much a joint action contributes to changes in each state dimension beyond what would arise without coordination. To this end, we adopt a counterfactual formulation~\cite{foerster2018counterfactual} to measure the collective influence on dimension $d$ at timestep $t$:
\begin{equation}
    \label{eq:inf}
    \mathrm{Inf}_t^{d}(s_t, a_t) = \sum_{i=0}^{n-1}\left\{\left|\hat{s}_{t+1}^d(s_t, a_t) - s_t^d\right| - \mathbb{E}_{a_t^i \sim \rho}\!\left[\left|\hat{s}_{t+1}^d(s_t, a_t^i, a_t^{-i}) - s_t^d\right|\right]\right\},
\end{equation}
where $\hat{s}(\cdot)$ approximates the environment dynamics, and $\rho$ denotes a random policy that removes coordinated contribution from agent $i$. This formulation isolates the additional state change attributable to coordinated joint actions, rather than independent or uncoordinated behavior.

While $\mathrm{Inf}_t^d$ captures instantaneous coordination, effective cooperation requires \emph{sustained} and temporally consistent influence. To account for temporal accumulation, we introduce an eligibility trace:
\begin{equation}
e_t^d = \gamma \cdot e_{t-1}^d + \mathrm{Inf}_t^d,
\end{equation}
which aggregates past influence signals and reflects how consistently a dimension has been jointly affected over time. Using this trace, we define the \textbf{Agent Focusing Influence (AFI)} intrinsic reward:
\begin{equation}
\label{eq:afi_reward}
r^{\mathrm{AFI}}_{t} = \sum_{d \in \mathcal{D}} \mathrm{Inf}_t^d \cdot \mathrm{max}(e_{t-1}^d,1),
\end{equation}
where $\mathrm{max}(e_{t-1}^d,1)$ prevents vanishing rewards, while reinforcing dimensions that have received consistent collective attention. By weighting the current influence $\mathrm{Inf}_t^d$ with the accumulated trace $e_{t-1}^d$, AFI amplifies dimensions where coordinated influence is repeatedly exerted, encouraging agents to sustain coherent joint behavior rather than transient and unstructured interactions.

To illustrate the effect of AFI, Fig.~\ref{fig:pushbox-dist-change} shows the empirical distribution of temporal changes in box positions in Push-2-Box, with and without AFI. Without AFI, panels~(a) and~(b) show that both boxes exhibit similarly low variation, reflecting the dispersed influence of vanilla QMIX. With AFI, panels~(c) and~(d) show increased variation in both box positions, indicating stronger collective influence. However, since AFI aggregates influence across all state dimensions, this effect is distributed broadly, including agent positions that do not require coordination. As a result, insufficient influence is concentrated on box positions, limiting effective task completion. This limitation motivates a mechanism that selectively prioritizes which dimensions to focus on, which we address in Section~\ref{subsec:sfi_fim}.

\subsection{State Focusing Influence and FIM Framework}
\label{subsec:sfi_fim}

\begin{figure}[!t]
  \centering
  \makebox[1\linewidth][c]{
    \begin{subfigure}[b]{0.27\linewidth}
      \centering
      \includegraphics[width=\linewidth]{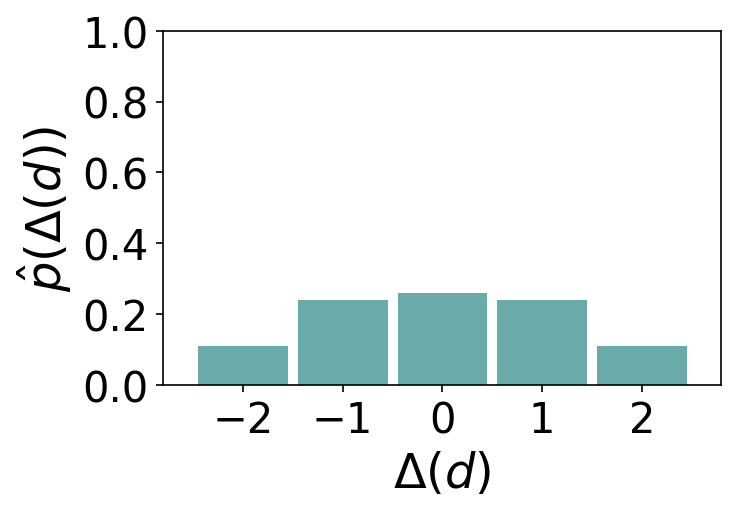}
      \caption{Agent positions}
    \end{subfigure}
    \hfill
    \begin{subfigure}[b]{0.27\linewidth}
      \centering
      \includegraphics[width=\linewidth]{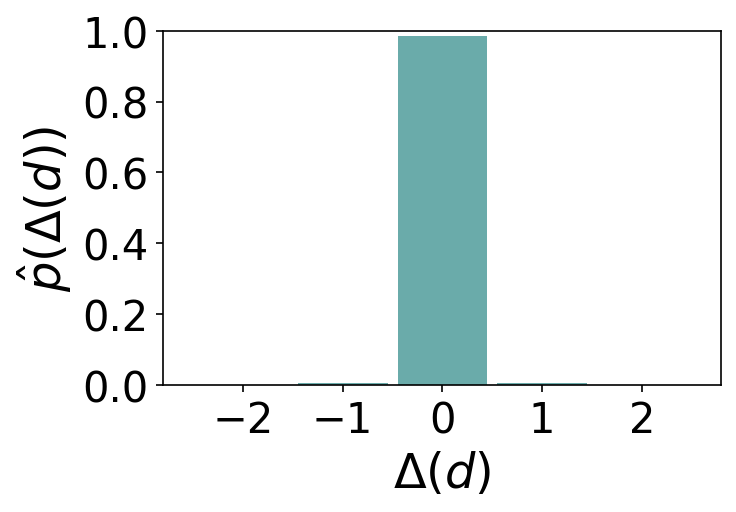}
      \caption{Box positions}
    \end{subfigure}
    \hfill
    \begin{subfigure}[b]{0.27\linewidth}
      \centering
      \includegraphics[width=\linewidth]{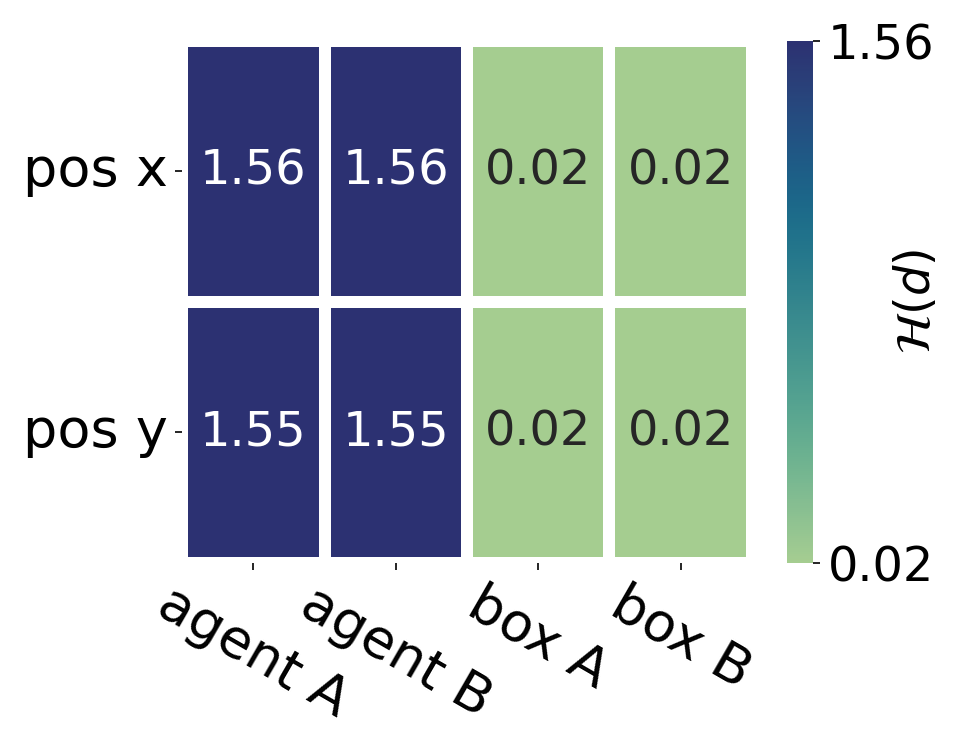}
      \caption{Entropy $\mathcal{H}(d)$}
    \end{subfigure}
  }  
  \caption{(a–b) Empirical distribution $\hat{p}$ of temporal changes $\Delta(d)$ for agent and box positions. (c) Entropy $\mathcal{H}(d)$ for each state dimension in Push-2-Box.}
  \label{fig:cog}
  \vspace{-1.5em}
\end{figure}

To concentrate the coordinated influence induced by AFI on dimensions that matter for exploration, we introduce \textbf{State Focusing Influence (SFI)}, which identifies and prioritizes state dimensions that are under-explored under the current joint behavior policy $\boldsymbol{\beta}$. Our key observation is that under-explored dimensions exhibit low diversity in their induced transitions, which can be characterized by low conditional entropy $\mathbb{E}_{\boldsymbol{\beta}}\!\left[\mathcal{H}(s^d_{t+1} \mid s_t)\right]$. Intuitively, if a dimension rarely changes under the current policy, it receives little effective exploration, even over long horizons.

\begin{proposition}
\label{prop:horizon}
Let $\mathcal{E}_d := \sup_{t \geq 0} \mathcal{H}(s^d_{t+1} \mid s^d_t)$ be positive. To achieve a target exploration level $H^d_{\mathrm{target}}$ such that $\mathcal{H}(s^d_0, \ldots, s^d_T) \geq H^d_{\mathrm{target}}$, the required horizon satisfies
\begin{equation}
    T \geq \frac{H^d_{\mathrm{target}} - \mathcal{H}(s^d_0)}{\mathcal{E}_d}.
\end{equation}
\end{proposition}

\textbf{Proof.} Proof of Proposition~\ref{prop:horizon} is provided in Appendix~\ref{secapp:proof}.

This result shows that when $\mathcal{H}(s^d_{t+1} \mid s^d_t)$ is small, each transition contributes little to exploration, requiring disproportionately long horizons. Therefore, such dimensions must be explicitly targeted to achieve sufficient coverage. However, directly comparing entropy across dimensions is inappropriate due to scale differences, since those with larger natural variation trivially yield higher entropy. To address this issue, we normalize each dimension by its expected change magnitude under $\boldsymbol{\beta}$:
\begin{equation}
\tilde{s}^d_t := \frac{s^d_t}{\mathbb{E}_{\boldsymbol{\beta}}[|s^d_{t+1} - s^d_t|] + \epsilon},
\end{equation}
which removes scale effects and enables fair comparison across dimensions, where $\epsilon$ is a small constant added for numerical stability.

We then approximate the conditional entropy using the marginal entropy of normalized changes:
\begin{equation}
    \mathcal{H}(d) = \mathbb{E}_{\boldsymbol{\beta}}\!\left[-\log p\!\left(
    \Delta^d(s_t, s_{t+1})\right)\right],
\end{equation}
where $\Delta^d := \tilde{s}^d_{t+1} - \tilde{s}^d_t$ and $p$ is estimated empirically. This approximation captures how diverse the observed transitions are along each dimension under the current policy.

Based on this, SFI assigns higher importance to dimensions with lower entropy, i.e., those that are insufficiently explored. Integrating SFI with AFI yields the \textbf{Focusing Influence Mechanism (FIM)}:
\begin{equation}
    r^{\mathrm{FIM}}_t = \sum_{d \in \mathcal{D}} w_d \cdot \mathrm{Inf}_t^d \cdot \mathrm{max}(e_{t-1}^d,1),
\end{equation}
where $w_d = \mathrm{Softmax}(-\mathcal{H}(d))$ emphasizes under-explored dimensions. To ensure stable adaptation to the evolving behavior policy, the weights $w_d$ are updated using an exponential moving average:
\begin{equation}
    w_d \leftarrow (1 - \phi)\, w_d + \phi\, w_d^{\text{new}},
\end{equation}
where $\phi$ is the update rate. Crucially, FIM does not explicitly maximize entropy. Instead, it promotes exploration by allocating coordinated influence to under-explored dimensions, increasing their variability through joint behavior rather than stochasticity alone.

To visualize the effect of SFI, Fig.~\ref{fig:cog} shows the empirical distribution of temporal changes for agent and box positions, along with the corresponding entropy $\mathcal{H}(d)$ for each state dimension in Push-2-Box. Agent positions exhibit high entropy due to frequent changes, whereas box positions have low entropy because they require coordinated actions. Consequently, SFI assigns higher weight to box-position dimensions, directing AFI's influence toward targets that require joint effort.

Finally, the overall reward is defined as
\[
r_t = r_{\mathrm{ext},t} + \alpha \cdot r^{\mathrm{FIM}}_t,
\]
where $\alpha$ is the intrinsic reward scale. We use QMIX~\cite{rashid2018qmix} as the backbone, although FIM is agnostic to the underlying CTDE algorithm. Further implementation details are provided in Appendix~\ref{subsecapp:full-imp-detail}.

\vspace{-0.7em}
\section{Experiment}
\label{sec:exp}

In this section, we evaluate the effectiveness of FIM. We begin with the
Push-2-Box task introduced in Section~\ref{subsec:motiv}, comparing
various combinations of FIM components. We then extend the evaluation to more complex benchmarks, including the StarCraft Multi-Agent Challenge
(SMAC)~\cite{samvelyan2019starcraft}, SMACv2~\cite{ellis2023smacv2}, and Google Research Football
(GRF)~\cite{kurach2020google}. In all performance plots, the mean across 5 random seeds is shown as a solid line, and the standard deviation is represented by a shaded area.

\vspace{-0.5em}
\subsection{Performance Comparison on Push-2-Box}
\label{subsec:pushbox}

We revisit the Push-2-Box task introduced in Fig.~\ref{fig:pushbox}(a). In
Push-2-Box, an external reward of $+100$ is given when either box reaches the wall, and $-1$ is applied if the task fails. The task is solved when agents push a box to the wall within the limited episode length, and detailed settings are provided in Appendix~\ref{secapp:env-details}. For comparison, Fig.~\ref{fig:pushbox-result} reports success rates for several MARL baselines: \textbf{LAIES}~\cite{liu2023lazy}, which encourages influence over heuristic external state features; \textbf{CDS}~\cite{li2021celebrating}, which promotes trajectory diversity; \textbf{FoX}~\cite{jo2024fox}, which leverages formation-aware exploration; and \textbf{vanilla QMIX}, trained using only extrinsic rewards. We also include component-wise variants: \textbf{QMIX+AFI}, which applies $r^{\mathrm{AFI}}_t$ with uniform weights across all state dimensions, and \textbf{QMIX+SFI}, which applies $\sum_{d \in \mathcal{D}} w_d \cdot \mathrm{Inf}^d_t$ without eligibility traces.

\begin{wrapfigure}{r}{0.38\textwidth}
\vspace{-1.2em}
\centering
\includegraphics[width=\linewidth]{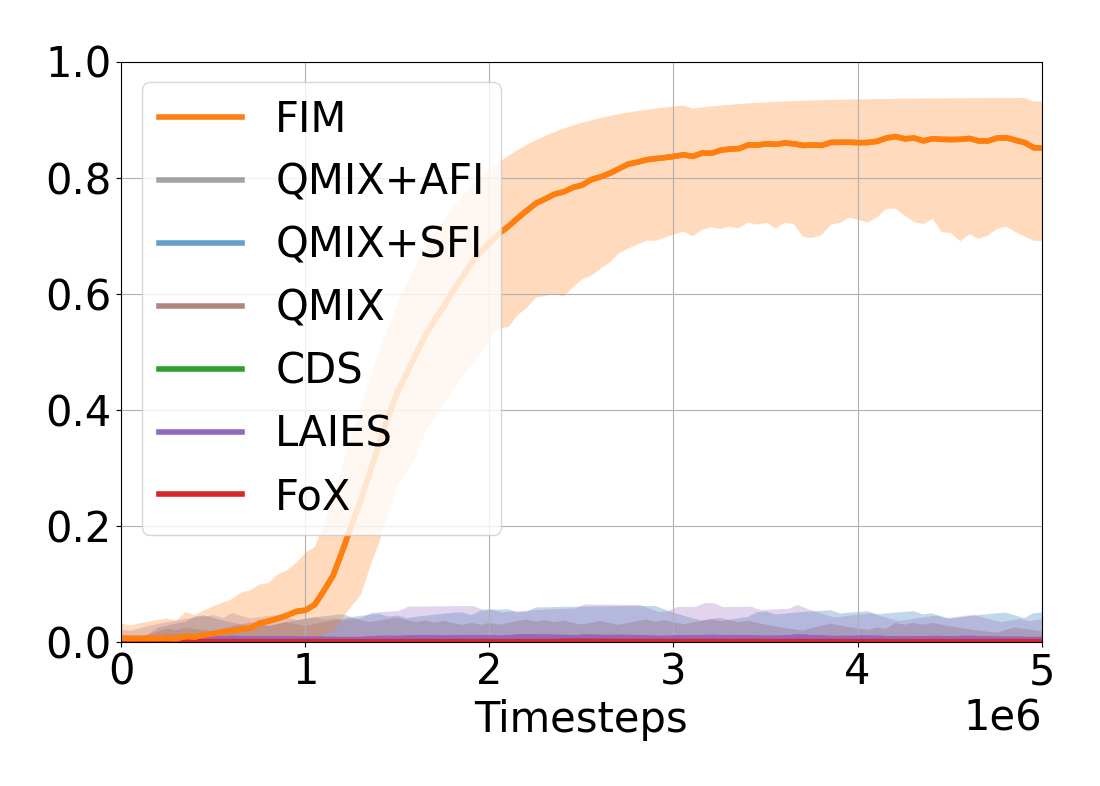}
\caption{Performance comparison on the Push-2-Box environment.}
\label{fig:pushbox-result}
\vspace{-1em}
\end{wrapfigure}

We observe that, despite the simplicity of the environment, only FIM solves the task. All other MARL baselines, including vanilla QMIX and exploration-oriented methods, fail on Push-2-Box. Notably, LAIES also fails even though it targets the same state dimensions in Push-2-Box, suggesting that identifying the correct dimensions is insufficient without AFI. Among our ablations, QMIX+AFI alone fails because concentrated influence without targeted dimension weighting is spread across all state dimensions, leaving insufficient joint effort on box positions to achieve the task, as shown in Fig.~\ref{fig:pushbox}(c). QMIX+SFI increases interaction with hard-to-change components that require joint effort, but still fails due to inconsistent focus on a single target. Together, these results show that both SFI and AFI are necessary, and that their combination enables effective influence focusing.

\subsection{Performance Comparison on Diverse Multi-Agent Environments}

Next, we evaluate our method on three standard MARL benchmarks: SMAC, SMACv2, and GRF.  
SMAC is a multi-agent combat environment built on StarCraft II, where agents must coordinate to defeat enemy units. We use a truly sparse reward setting in which agents receive +1 for a win, 0 for a draw, and -1 for a loss. Evaluation is conducted on five challenging scenarios: one hard map (\texttt{3s\_vs\_5z}) and four super-hard maps (\texttt{corridor}, \texttt{MMM2}, \texttt{6h\_vs\_8z}, \texttt{3s5z\_vs\_3s6z}), where \texttt{s}, \texttt{z}, and \texttt{h} refer to stalker, zealot, and hydralisk units, respectively. We further evaluate on SMACv2, which introduces randomized initial configurations and diverse unit types, using three scenarios (\texttt{zerg\_5\_vs\_5}, \texttt{terran\_5\_vs\_5}, \texttt{protoss\_5\_vs\_5}) under the same sparse reward setting.
GRF is a multi-agent soccer environment where teams compete to score goals under sparse rewards: +100 for a win and -1 for a loss. We evaluate on eight scenarios, including four half-field settings (\texttt{academy\_2\_vs\_2}, \texttt{academy\_3\_vs\_2}, \texttt{academy\_4\_vs\_3}, \texttt{academy\_counterattack}) and their corresponding full-field versions, which are more challenging due to the increased field size. Further environment details and visualizations are provided in Appendix~\ref{secapp:env-details}.

For SMAC and SMACv2, in addition to {\bf Vanilla QMIX}, {\bf LAIES}, {\bf CDS}, and {\bf FoX}, we include baselines compared with FIM variants: {\bf COMA}~\cite{foerster2018counterfactual}, which uses counterfactual baselines for credit assignment; {\bf MASER}~\cite{jeon2022maser}, which identifies subgoals based on $Q$-values; {\bf RODE}~\cite{wang2021rode}, which assigns roles to agents; {\bf MA}$^2${\bf E}~\cite{kangma}, which applies masking to reconstruct global information; and {\bf QPLEX}~\cite{wangqplex}, which adopts a dueling architecture. For GRF, we compare against Vanilla QMIX, LAIES, CDS, FoX, COMA, MA$^2$E, and QPLEX, omitting methods without publicly available GRF results. We also include QMIX-DR for SMAC, trained with dense rewards, as an upper-bound reference. All baselines are evaluated using author-provided implementations, while our method uses the best hyperparameters identified via ablations. Further details, including algorithm descriptions and hyperparameter configurations, are provided in Appendix~\ref{subsecapp:baseline} and Appendix~\ref{subsecapp:hyperparam}, respectively. Here, SFI in FIM typically assigns higher weights to opponent state dimensions that are difficult to change, whereas LAIES heuristically targets all enemy-related dimensions, highlighting a clear difference from our approach.

\begin{figure}[!t]
  \vspace{-0.7em}
  \centering
  \includegraphics[width=0.85\linewidth]{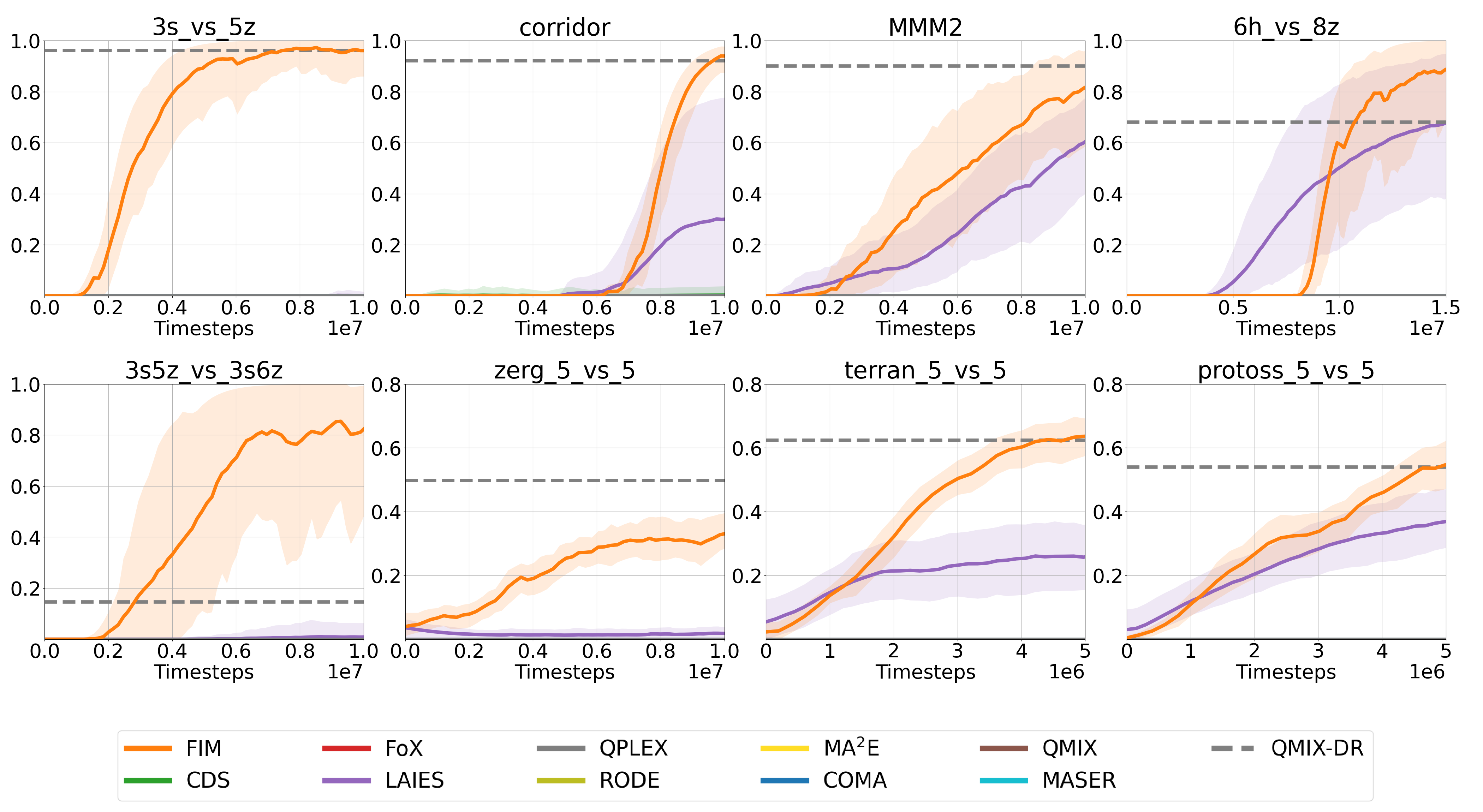}
  \caption{Performance comparison on SMAC and SMACv2 environments}
  \label{fig:SMAC-results}
  \centering
  \vspace{0.5em}
  \includegraphics[width=0.85\linewidth]{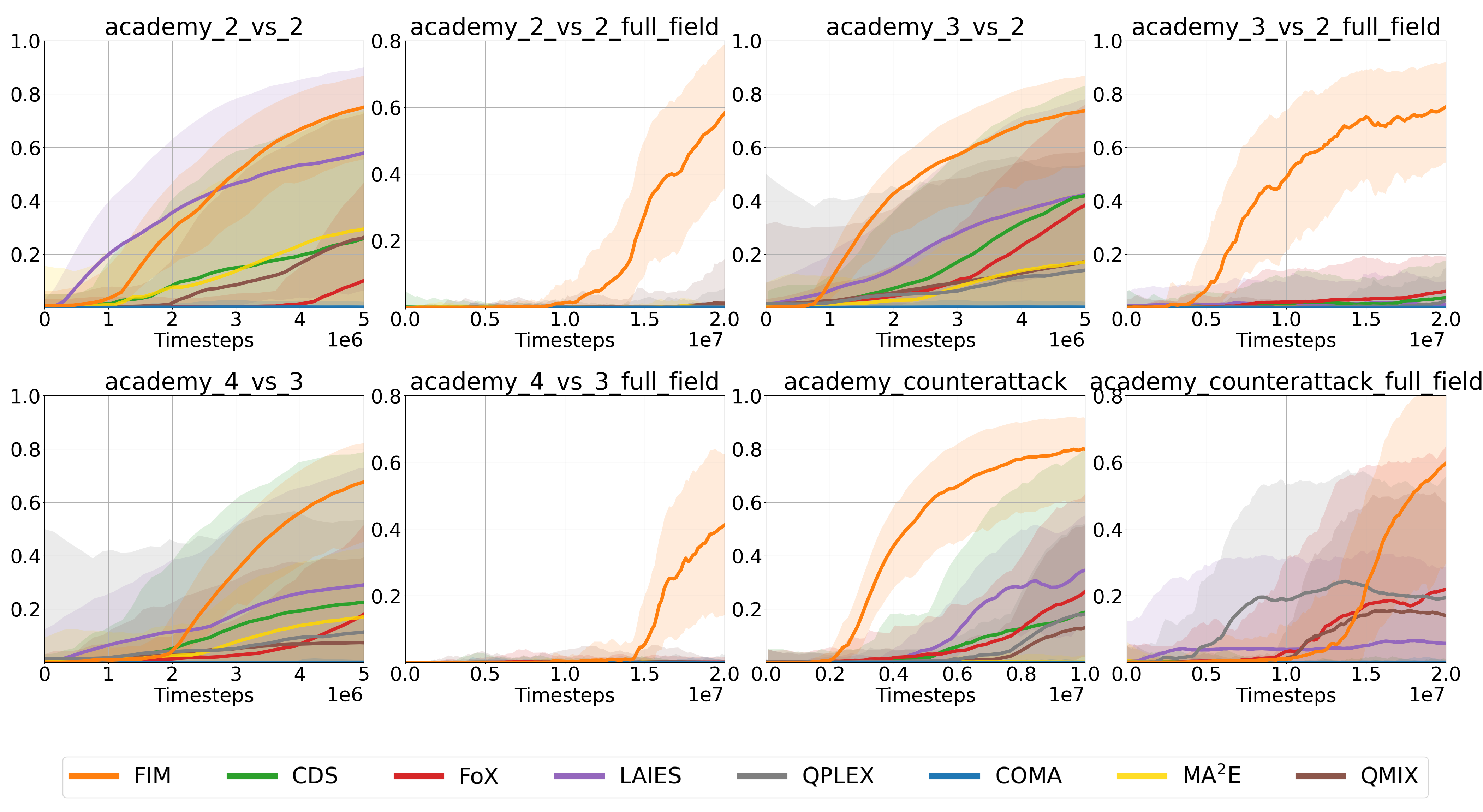}
  \caption{Performance comparison on GRF environments}
  \label{fig:GRF-results}
  \vspace{-1em}
\end{figure}

Fig.~\ref{fig:SMAC-results} and Fig.~\ref{fig:GRF-results} present success rate comparisons on the SMAC, SMACv2, and GRF benchmarks. Across these environments, FIM consistently achieves the highest success rates among all baselines. In SMAC and SMACv2, sparse rewards pose a significant challenge because agents must eliminate all enemies without intermediate feedback. While LAIES remains competitive in some scenarios, it struggles on complex maps such as \texttt{3s\_vs\_5z} and \texttt{corridor}, since it treats all enemies as external state dimensions without focusing influence. In contrast, FIM remains robust by weighting under-explored dimensions. In GRF, although some baselines perform well on simpler half-field tasks, they largely fail on full-field maps with rare scoring opportunities. By concentrating influence on hard-to-change elements, FIM maintains strong performance across all scenarios. Overall, these results show that FIM promotes effective cooperation, enabling agents to solve challenging tasks even under highly sparse rewards.

To demonstrate scalability beyond structured state representations, we further evaluate FIM on Overcooked~\cite{carroll2019utility} in Appendix~\ref{subsecapp:overcooked}, which involves pixel-based observations. In this setting, instead of predefined state dimensions, FIM implicitly focuses on relevant regions of the image, demonstrating that it remains effective beyond structured state settings.

\subsection{In-depth Analysis and Ablation Studies}
\label{sec:ablations-strategy-analysis}

\begin{figure}[!h]
    \centering
    \includegraphics[width=1\linewidth]{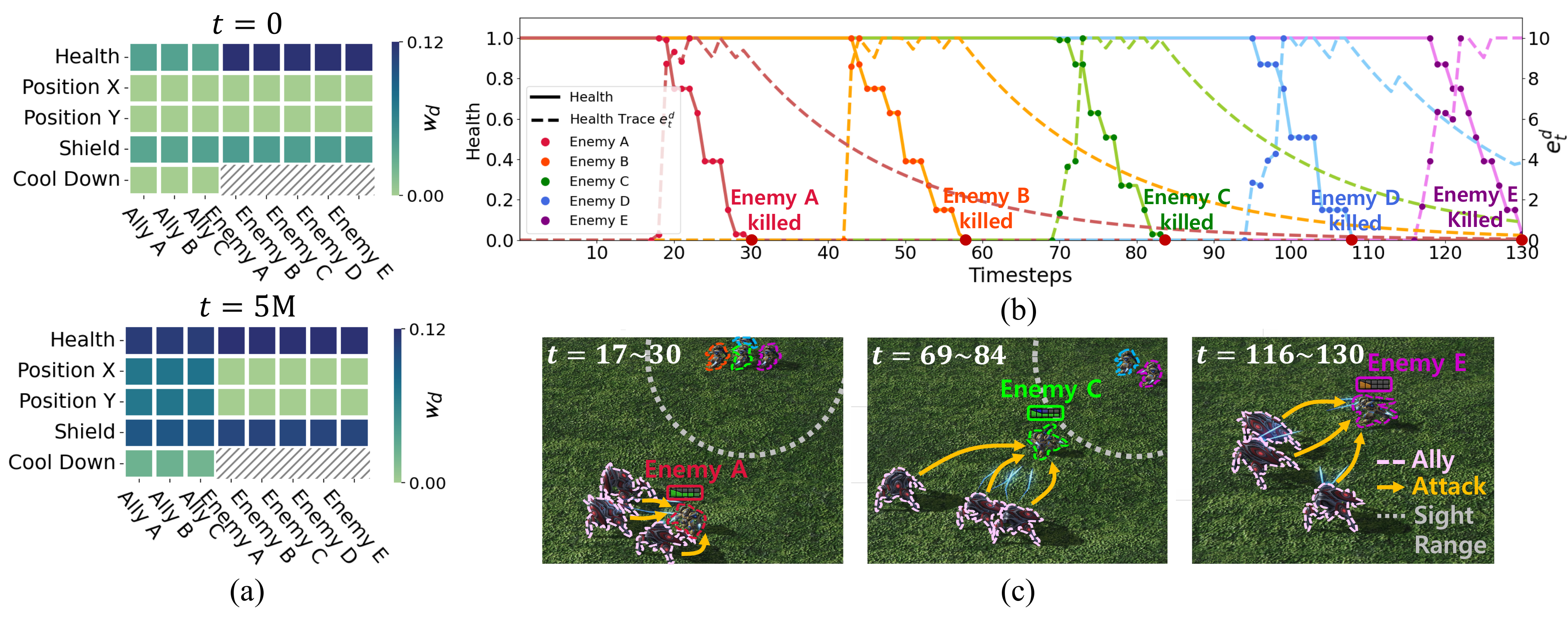}
    \caption{Behavior analysis in \texttt{3s\_vs\_5z}: (a) Dimensional weight $w_d$ across iterations (b) Changes in enemy health and its trace $e_t^d$. (c) Rendered frames for highlighting agents' coordination.}
    \label{fig:SMAC-analysis}
\end{figure}

To better understand FIM's focusing mechanisms, we conduct detailed analyses and ablations in the environments where it shows the largest gains: SMAC \texttt{3s\_vs\_5z} and GRF \texttt{academy\_3\_vs\_2\_full\_field}. In SMAC \texttt{3s\_vs\_5z}, SFI highlights enemy features such as health as under-explored state dimensions, as they remain relatively stable without coordination and thus become targets for joint influence. As shown in Appendix~\ref{subsecapp:w_d-vis}, similar state dimensions exhibit low entropy in other SMAC scenarios. In GRF, the keeper's position is frequently up-weighted by SFI, since it is difficult for agents to manipulate. Overall, these results show that FIM focuses influence on meaningful state dimensions, improving performance by prioritizing under-explored features such as health in SMAC and the keeper's position in GRF.

\paragraph{Behavior Analysis:} To further illustrate the effect of FIM, Fig.~\ref{fig:SMAC-analysis}(a) reports the dimensional weights $w_d$ across training iterations ($t = 0$ and $5$M) in \texttt{3s\_vs\_5z}. Throughout training, the health of enemy units consistently receives high weights from SFI, since these dimensions change substantially only when agents coordinate their attacks to focus fire on the same target. Notably, as training progresses, ally health dimensions are also gradually up-weighted, since agents tend to avoid battle to preserve their own health. By assigning higher weight to ally health, SFI encourages agents to engage more actively with enemies. Fig.~\ref{fig:SMAC-analysis}(b) shows how the corresponding eligibility traces evolve over the enemy-health dimensions, while Fig.~\ref{fig:SMAC-analysis}(c) highlights key timesteps at which enemy units are eliminated within the trajectory. Agents trained with FIM learn to pull enemies into sight and focus fire sequentially: around $t \approx 20$, they concentrate on the first red enemy, increase its health trace, and eliminate it by $t \approx 30$; once removed, its influence drops to zero and attention shifts to the next enemy (e.g., orange at $t \approx 40$), repeating the same process. This strategy resembles human gameplay in StarCraft II. We provide a complementary analysis for GRF in Appendix~\ref{subsecapp:grf-strategy-analysis}, showing that agents learn to disrupt the keeper's behavior, thereby increasing goal-scoring opportunities. Together, these results demonstrate that FIM promotes structured and effective cooperation even in sparse-reward environments.

\begin{figure}[!t]
  \centering
  \makebox[1\linewidth][c]{
    \begin{subfigure}[b]{0.29\linewidth}
      \centering
      \includegraphics[width=\linewidth]{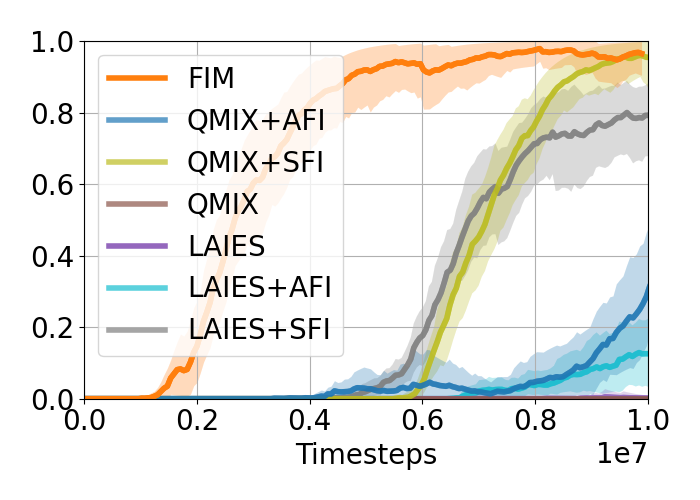}
      \caption{Component evaluation}
    \end{subfigure}
    \hfill
    \begin{subfigure}[b]{0.29\linewidth}
      \centering
      \includegraphics[width=\linewidth]{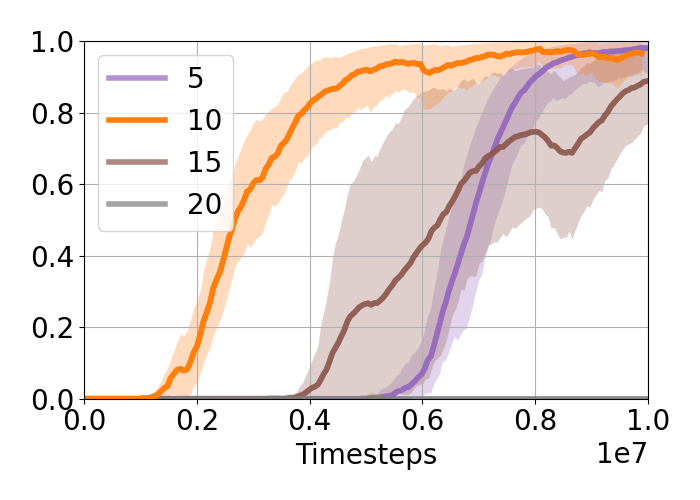}
      \caption{Reward scale $\alpha$}
    \end{subfigure}
    \hfill
    \begin{subfigure}[b]{0.29\linewidth}
      \centering
      \includegraphics[width=\linewidth]{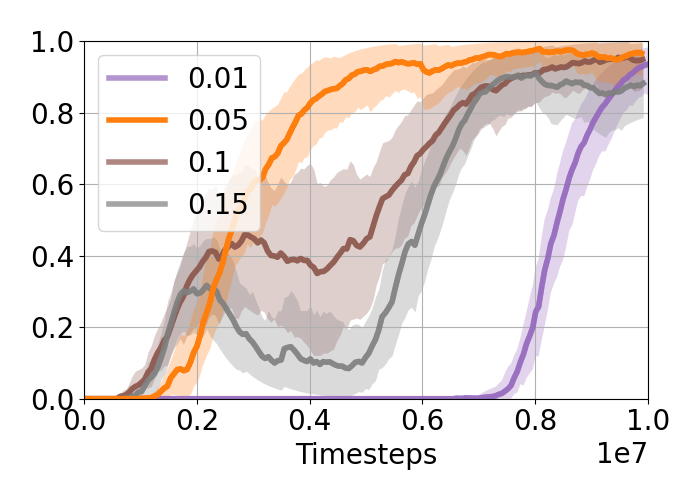}
      \caption{Update rate $\phi$}
    \end{subfigure}
  }  
  \caption{Ablation studies on SMAC \texttt{3s\_vs\_5z}}
  \label{fig:SMAC-3s5z-abl}
  \vspace{-1.5em}
\end{figure}

\paragraph{Component Ablations:} We evaluate the contribution of each component on \texttt{3s\_vs\_5z}. Fig.~\ref{fig:SMAC-3s5z-abl}(a) compares \textbf{Vanilla QMIX}, \textbf{QMIX+AFI}, \textbf{QMIX+SFI}, \textbf{LAIES}, \textbf{LAIES+AFI}, which rescales LAIES's influence using per-dimension eligibility traces, and \textbf{LAIES+SFI}, which applies $w_d$ as a per-dimension weight on LAIES's influence signal, and \textbf{FIM}. The results show that Vanilla QMIX fails to solve the task, while SFI and AFI each provide consistent improvements regardless of the backbone: both QMIX+AFI and LAIES+AFI benefit from enforcing consistent focus on shared targets via eligibility traces, and both QMIX+SFI and LAIES+SFI benefit from concentrating influence on a small number of under-explored dimensions. Combining SFI and AFI yields faster convergence and higher final success rates, confirming that the two components play essential and synergistic roles in influence-based exploration.

\paragraph{Hyperparameter Analysis:} Fig.~\ref{fig:SMAC-3s5z-abl}(b)-(c) examine the effects of the intrinsic reward scale $\alpha$ and under-explore update rate $\phi$. For $\alpha$, too-small values provide insufficient incentive for agents to concentrate their influence, while overly large values cause agents to over-prioritize intrinsic rewards over extrinsic feedback, leading to unstable training. In practice, $\alpha$ is environment-dependent, but can be effectively chosen by scaling it as large as possible without dominating the extrinsic reward signal. For $\phi$, too-small values cause the dimension weights $w_d$ to adapt too slowly to the evolving behavior policy, whereas too-large values result in frequent weight shifts that destabilize the focus target and prevent sustained coordinated influence. Although larger $\phi$ leads to faster adaptation, the weights eventually converge over time, suggesting that $\phi$ primarily controls the adaptation timescale rather than the final behavior. We use $\alpha = 10$ and $\phi = 0.05$ as default hyperparameters.

Additional analyses, including comparisons with alternative dimension-weighting strategies and ablations on more scenarios, are provided in Appendix~\ref{subsecapp:abl}. The learned dynamics model is further analyzed in Appendix~\ref{subsecapp:dynamics-model}, where we show that its evolving prediction error correlates with exploratory effectiveness and overall performance. Additionally, Appendix~\ref{subsecapp:complexity} reports computational complexity, showing that FIM achieves superior performance with training time comparable to QMIX.

% \vspace{-0.8em}
% \subsection{Generalization to Unstructured State Spaces: Overcooked}
% \begin{wrapfigure}{r}{0.35\textwidth}
%     \vspace{-1.5em}
%     \centering    \includegraphics[width=\linewidth]{fig/overcook.png}
%     \caption{Test return score in Overcooked \texttt{coord-ring}}
%     \vspace{-1em}
%     \label{fig:overcooked}
% \end{wrapfigure}
% \tcr{To address the concern regarding applicability beyond structured vector states, we conduct an additional experiment in an unstructured state setting using Overcooked~\cite{carroll2019utility} (coord-ring, JaxMARL~\cite{flair2024jaxmarl} implementation). Unlike standard structured representations that explicitly provide object coordinates, the state is represented as a height $\times$ width $\times$ components tensor, where the positions of components (e.g., agents, ingredients) are represented via binary encodings over spatial grids. Even in this unstructured representation, we observe that certain dimensions exhibit notably low entropy. In particular, the soup location remains under-explored, since it cannot be moved until it is fully prepared. As training progresses, agents explore diverse positions with the soup and eventually converge to efficient goal-directed trajectories. As shown in Fig.~\ref{fig:overcooked}, FIM achieves a performance improvement of approximately 10\% over the baseline (IPPO)}.

\vspace{-0.5em}
\section{Limitation and Future Work}
\label{sec:limitation}

Although FIM achieves strong performance, it still has some limitations. FIM incurs additional computational cost due to training the dynamics model, as analyzed in Appendix~\ref{subsecapp:complexity}. However, our analysis shows that this overhead is moderate relative to the performance gains, and it could be further reduced by using more lightweight model approximations. Second, although we validate FIM in unstructured settings through Overcooked experiments, as shown in Appendix~\ref{subsecapp:overcooked}, it is not yet specifically optimized for image-based environments. In such settings, learning to identify and focus on semantically meaningful regions or objects in the observation space could further improve performance. Deeper integration with image-based representations, for example through object-centric or detection-based mechanisms, remains an important direction for future work.

\vspace{-0.5em}
\section{Conclusion}

In this paper, we address the challenge of efficient cooperation in sparse-reward MARL by proposing FIM, a framework that guides agent influence toward under-explored state dimensions and sustains coordinated focus through eligibility traces. By integrating entropy-based dimension weighting with structured intrinsic rewards based on counterfactual reasoning, FIM enables agents to induce targeted and persistent state transitions. Empirical results across Push-2-Box, SMAC, SMACv2, and GRF demonstrate that FIM significantly improves learning efficiency and coordination, outperforming state-of-the-art baselines. These findings highlight the potential of influence-guided learning to enable robust multi-agent cooperation in sparsely rewarded environments.
 
\newpage
\bibliographystyle{unsrtnat}
\bibliography{references}

%%%%%%%%%%%%%%%%%%%%%%%%%%%%%%%%%%%%%%%%%%%%%%%%%%%%%%%%%%%%

\newpage
\appendix

\section{Broader Impact}
\label{secapp:broader-impact}
This work advances cooperative multi-agent systems by introducing a framework that fosters coordinated behavior through influence-based intrinsic motivation. Enhanced cooperation among agents holds strong potential for positive societal impact in domains such as autonomous vehicle coordination, collaborative robotics, disaster response, and environmental monitoring. In these settings, the ability of agents to reason about and influence under-explored aspects collaboratively can lead to more robust, adaptive, and efficient team performance. As a foundational contribution, this research supports the development of AI systems that are better aligned with collective goals, promoting safer and more effective deployment in real-world multi-agent environments.

\section{Proof of Proposition 4.1}
\label{secapp:proof}
\begin{proof}
By the chain rule of entropy:
\begin{align}
    \mathcal{H}(s^d_0, s^d_1, \ldots, s^d_T)
    &= \mathcal{H}(s^d_0) + \sum_{t=0}^{T-1} \mathcal{H}(s^d_{t+1} \mid s^d_0, \ldots, s^d_t) \\
    &\leq \mathcal{H}(s^d_0) + \sum_{t=0}^{T-1} \mathcal{H}(s^d_{t+1} \mid s^d_t) \\
    &\leq \mathcal{H}(s^d_0) + T \cdot \mathcal{E}_d,
\end{align}
where the first inequality follows from the fact that conditioning reduces
entropy, and the second from the definition of $\mathcal{E}_d :=
\sup_{t \geq 0} \mathcal{H}(s^d_{t+1} \mid s^d_t)$. Therefore, to achieve
$\mathcal{H}(s^d_0, s^d_1, \ldots, s^d_T) \geq H^d_{\mathrm{target}}$, we
need:
\begin{equation}
    \mathcal{H}(s^d_0) + T \cdot \mathcal{E}_d \geq H^d_{\mathrm{target}},
\end{equation}
which gives:
\begin{equation}
    T \geq \frac{H^d_{\mathrm{target}} - \mathcal{H}(s^d_0)}{\mathcal{E}_d}.
\end{equation}
\end{proof}

\section{Implementation Details}
In this section, we provide practical details on how the proposed framework is implemented. First, we describe the empirical estimation of difference-based entropy $\mathcal{H}(d)$ in Appendix~\ref{subsecapp:sfi-imp-detail}. Next, we present the overall algorithmic details in  Appendix~\ref{subsecapp:full-imp-detail}.

\subsection{Empirical Estimation of $\mathcal{H}(d)$}
\label{subsecapp:sfi-imp-detail}

To estimate $\mathcal{H}(d)$, we construct the empirical distribution $\hat{p}\left(\Delta^d(s_t, s_{t+1})\right)$ by counting occurrences discretized to two decimal places from 100K episodes collected under the current behavior policy. The entropy is then computed as:
\begin{equation}
\mathcal{H}(d) \approx \mathbb{E}_{\boldsymbol{\beta}} \left[ -\log \hat{p}(\Delta^d(s_t, s_{t+1})) \right]
\end{equation}
To ensure comparability across environments, $\mathcal{H}(d)$ values are min-max normalized to the range $[0, 1]$ within each environment. To assess the validity of using marginal entropy as a surrogate, we empirically measured both the conditional and marginal entropies for each state dimension in GRF \texttt{academy\_2\_vs\_2}, as shown in Table~\ref{tab:entropy-comparison}. We found that the two quantities are numerically very close and, more importantly, induce almost identical rankings across dimensions. Since under-explored selection depends on this ranking, the marginal entropy provides a reliable surrogate in practice.

\newpage
\vspace{-2em}
\begin{table}[!t]
\caption{Comparison of $\mathcal{H}\left(\Delta^d(s_t, s_{t+1})\right)$ and $\mathcal{H}\left(\Delta^d(s_t, s_{t+1}) \mid s_t\right)$, averaged over state dimensions, in GRF \texttt{academy\_2\_vs\_2}}
\vspace{1em}
\centering
\begin{tabular}{@{}lcc@{}}
\toprule
\textbf{State Dimension} & $\mathcal{H}\left(\Delta^d(s_t, s_{t+1})\right)$ & $\mathcal{H}\left(\Delta^d(s_t, s_{t+1}) \mid s_t\right)$ \\ 
\midrule
Ally A position      & 0.73 & 0.72 \\
Ally A direction     & 0.82 & 0.80 \\
Ally B position      & 0.73 & 0.71 \\
Ally B direction     & 0.78 & 0.73 \\
Opponent GK position    & 0.27 & 0.21 \\
Opponent GK direction   & 0.32 & 0.26 \\
Opponent A position     & 0.74 & 0.70 \\
Opponent A direction    & 0.96 & 0.94 \\
Ball position        & 0.47 & 0.45 \\
Ball direction       & 0.24 & 0.23 \\
\bottomrule
\end{tabular}
\label{tab:entropy-comparison}
\vspace{-1em}
\end{table}

\subsection{Algorithmic Details of FIM}
\label{subsecapp:full-imp-detail}

Computing the influence $\mathrm{Inf}_t^d$ in Eq.~\eqref{eq:inf} relies on predicting next-state outcomes. FIM trains transition model $\hat{s}$ online, implemented as a three-layer MLP, by minimizing the mean squared error:
\begin{equation}
    \mathcal{L}_{\hat{s}} = \mathbb{E}_{s_t,\mathbf{a}_t,s_{t+1}}\!\left[
        \bigl\|\hat{s}(s_t, \mathbf{a}_t) - s_{t+1}\bigr\|^2
    \right]
    \label{eq:dynamics_loss}
\end{equation}

The FIM framework follows the CTDE paradigm, using QMIX to learn a joint action-value function. Each agent $i$ has $Q$-function $Q^i(\tau^i_t, a^i_t)$ based on its action-observation history $\tau^i_t$ and action $a^i_t$. These per-agent utilities are combined via a mixing network to produce a global joint $Q$-value, $Q_\theta^{\mathrm{tot}}(s_t, \mathbf{a}_t)$, where $\theta$ denotes the parameters of the mixing network. To stabilize learning, FIM employs a target mixing network $Q_{\theta^-}^{\mathrm{tot}}$, which is periodically updated by overwriting its parameters. The temporal difference (TD) loss is computed using a Bellman update:
\begin{equation}
\mathcal{L}_{\mathrm{TD}}(\theta) = \mathbb{E}_{s, \mathbf{a}, r, s'} \left[ \left( r_t + \gamma \max_{\mathbf{a}'} Q_{\theta^-}^{\mathrm{tot}}(s_{t+1}, \mathbf{a}') - Q_\theta^{\mathrm{tot}}(s_t, \mathbf{a}_t) \right)^2 \right]
\end{equation}
The parameters $\theta$ are updated by Adam optimizer, while the target network parameters $\theta^-$ are synchronized at fixed intervals. The complete training procedure of FIM is summarized in Algorithm~\ref{alg:FIM}.

\begin{algorithm}[!h]
\caption{FIM framework}
\label{alg:FIM}
\begin{algorithmic}[1]
\STATE \textbf{Initialize:} Q network, dynamics model $\hat{s}$
\STATE Collect trajectories under behavior policy
\STATE Approximate $\mathcal{H}(d)$ with the obtained trajectories based on Eq.~(8)
\STATE Compute $w_d$ for each $d \in \mathcal{D}$
\FOR{each training iteration}
    \FOR{each timestep $t$}
        \STATE Sample transition $(s_t,\mathbf{o}_t,\mathbf{a}_t,s_{t+1}, \mathbf{o}_{t+1})$ using $\boldsymbol{\pi}$,
        \STATE \hspace{1em} where $\mathbf{o}_t=(o^0_t,\cdots,o^{n-1}_t)$
        \FOR{each $d \in \mathcal{D}$}
            \STATE Compute influence $\mathrm{Inf}^d_t$ by Eq.~(3)
            \STATE Update eligibility trace $e^d_t$ by Eq.~(4)
        \ENDFOR
        \STATE Compute intrinsic reward $r^{\mathrm{FIM}}_{t}$ by Eq.~(9)
    \ENDFOR
    \STATE Update value function $Q^{\mathrm{tot}}$ and dynamics model $\hat{s}$
    \IF{interval of 0.5M timesteps}
        \STATE Re-estimate $\mathcal{H}(d)$ under current behavior policy based on Eq.~(8)
        \STATE Update $w_d$ for each $d \in \mathcal{D}$ by Eq.~(10)
    \ENDIF
\ENDFOR
\end{algorithmic}
\end{algorithm}

\newpage

\section{Environment Details}
\label{secapp:env-details}
\paragraph{Push-2-Box:} Push-2-Box is a cooperative multi-agent environment where two agents must jointly push one of two boxes toward a wall to obtain a reward, as depicted in Fig.~\ref{fig:push2box}. A box moves one cell if pushed by a single agent and two cells if pushed simultaneously. Thus, synchronized cooperation is essential for completing the task within the episode time limit. The environment terminates either when a box reaches the wall or when the episode length is exceeded. The \textbf{state space} consists of the $(x, y)$ positions of agents and boxes, resulting in an 8-dimensional state vector. Each agent receives the full environment state as observation. The \textbf{action space} is discrete, consisting of eight movement actions corresponding to up, down, left, right, top-right, right-down, down-left, and left-top directions. The \textbf{reward function} is described in Table~\ref{tab:push2box-reward}.

\vspace{1em}

\begin{minipage}[!b]{0.4\linewidth}
    \centering
    \includegraphics[width=0.5\linewidth]{fig/pushbox.pdf}
    \captionof{figure}{Push-2-Box}
    \label{fig:push2box}
\end{minipage}
\hfill
\begin{minipage}[!b]{0.6\linewidth}
    \centering
    \captionof{table}{Reward setting in Push-2-Box}
    \begin{tabular}{@{}ll@{}}
    \toprule
    \textbf{Event} & \textbf{Reward} \\ \midrule
    At least one box reaching the wall & +100 at episode end \\
    No box reaching the wall & -1 at episode end \\
    \bottomrule
    \end{tabular}
    \label{tab:push2box-reward}
\end{minipage}

\begin{figure}[!h]
  \centering
  \begin{subfigure}[T]{0.42\linewidth}
    \centering
    \includegraphics[width=\linewidth]{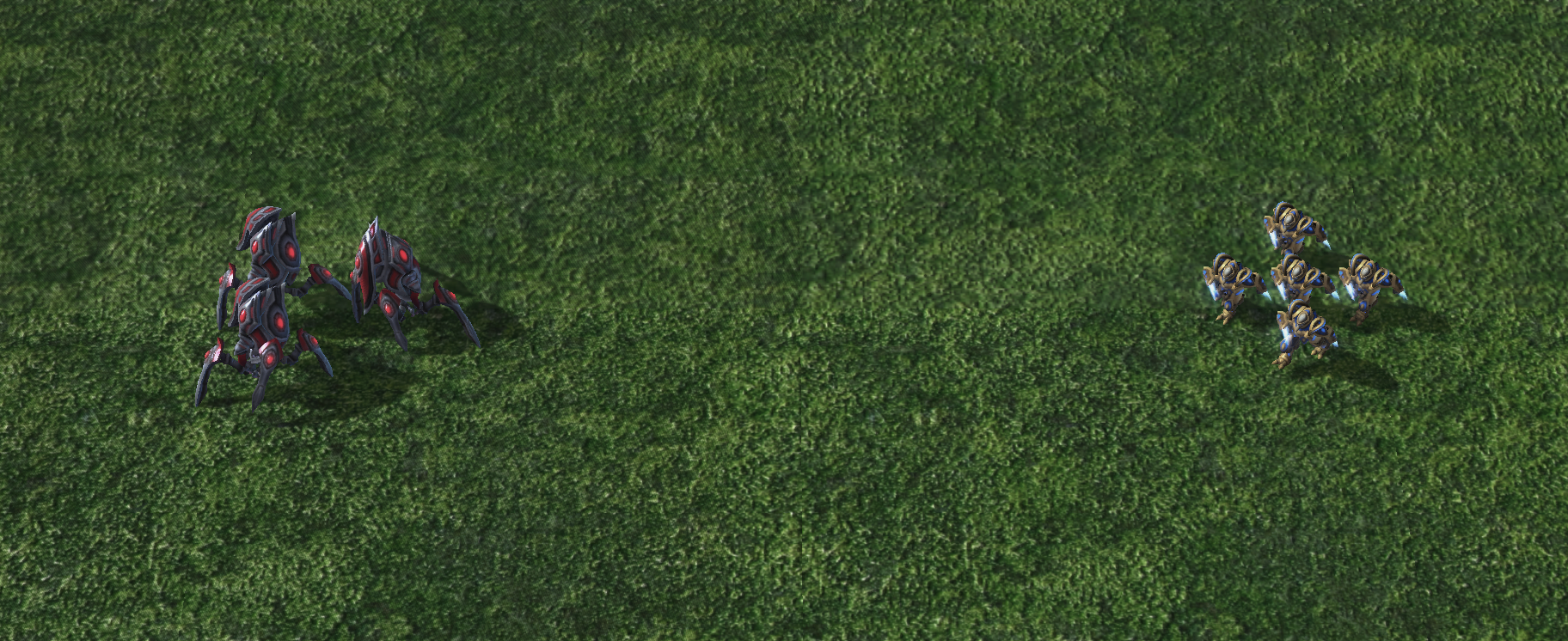}
    \caption{\texttt{3s\_vs\_5z}}
  \end{subfigure}
  \hfill
  \begin{subfigure}[T]{0.42\linewidth}
    \centering
    \includegraphics[width=\linewidth]{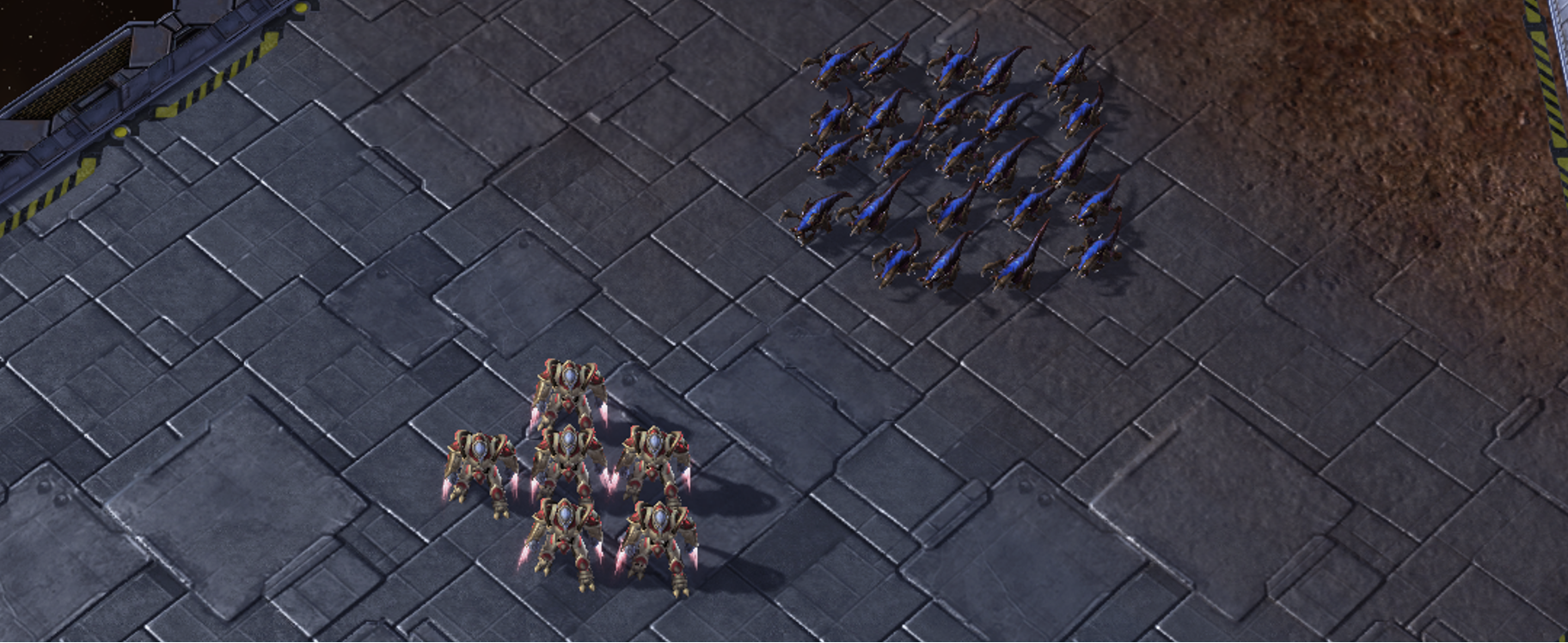}
    \caption{\texttt{corridor}}
  \end{subfigure}
  \begin{subfigure}[T]{0.42\linewidth}
    \centering
    \includegraphics[width=\linewidth]{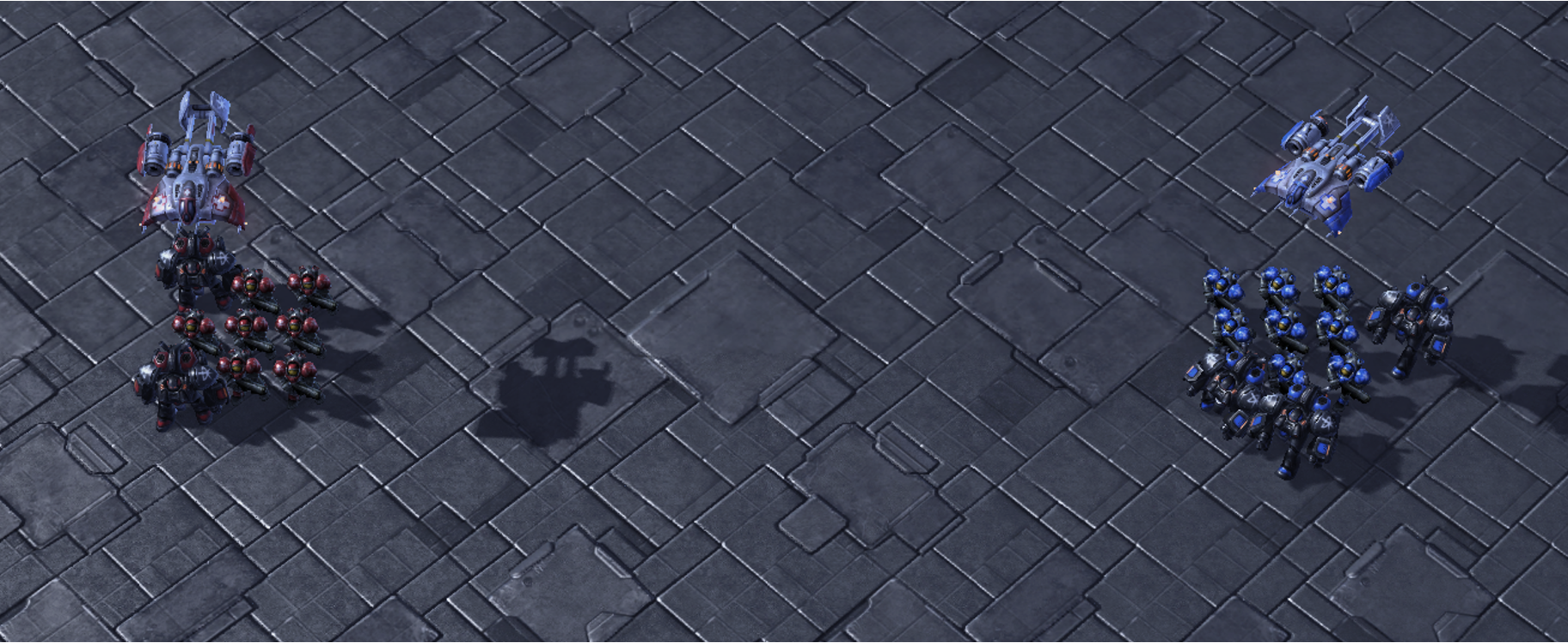}
    \caption{\texttt{MMM2}}
  \end{subfigure}
  \hfill
  \begin{subfigure}[T]{0.42\linewidth}
    \centering
    \includegraphics[width=\linewidth]{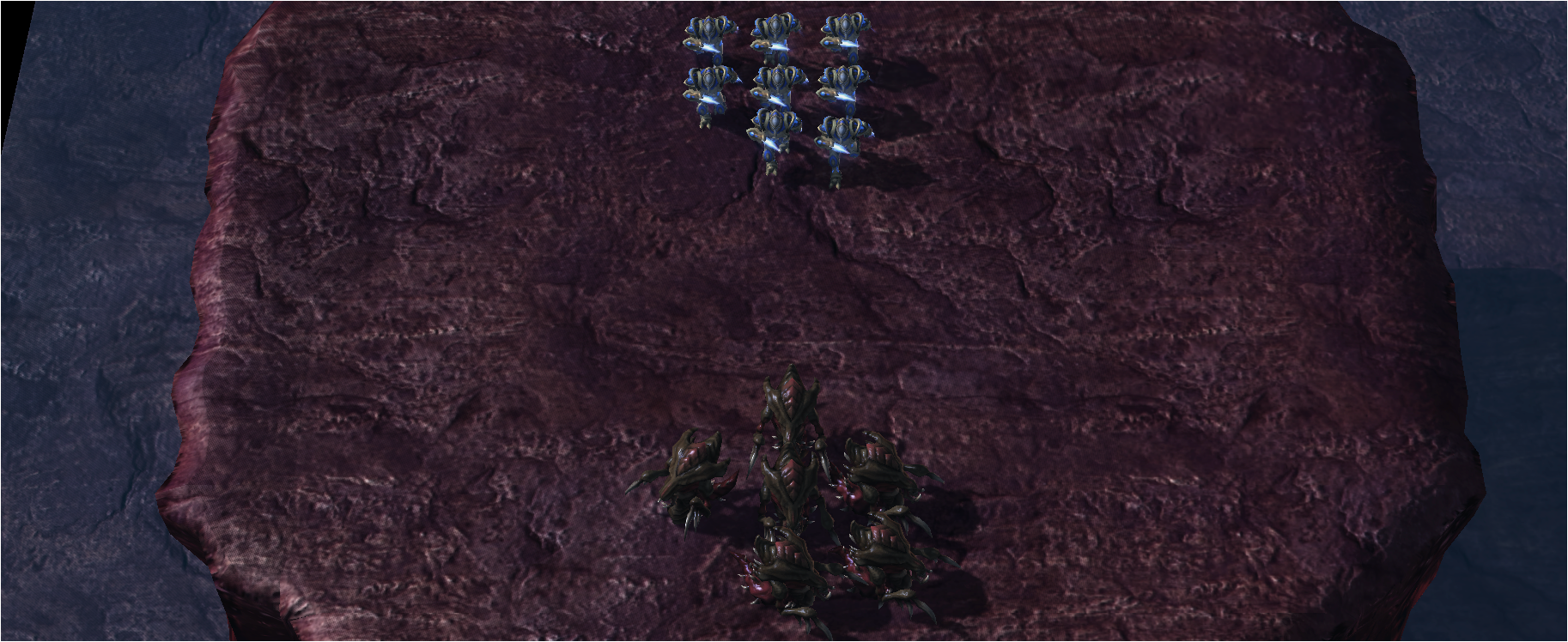}
    \caption{\texttt{6h\_vs\_8z}}
  \end{subfigure}  
  \begin{subfigure}[T]{0.42\linewidth}
    \centering
    \includegraphics[width=\linewidth]{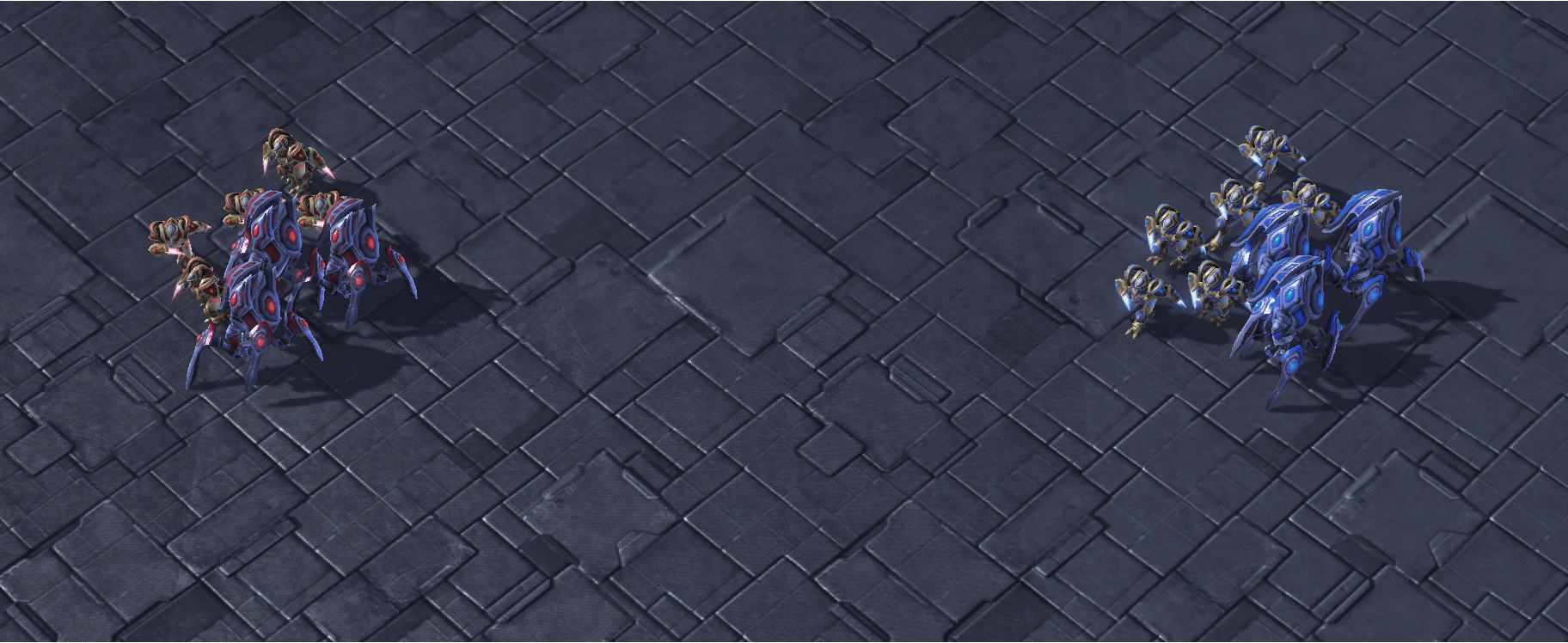}
    \caption{\texttt{3s5z\_vs\_3s6z}}
  \end{subfigure}    \caption{Visualization SMAC scenarios.}
  \label{fig:smac-vis}
\end{figure}

\paragraph{StarCraft Multi-Agent Challenge (SMAC \& SMACv2):} SMAC~\cite{samvelyan2019starcraft} is a benchmark for evaluating cooperative multi-agent reinforcement learning. Agents control individual StarCraft II units and coordinate to defeat enemy forces. Scenario visualizations are shown in Fig.\ref{fig:smac-vis}, and unit compositions are summarized in Table~\ref{tab:smac-conf}. The \textbf{state space} contains global information about all units (e.g., positions, health, and shields), while the \textbf{observation space} provides each agent with local features of nearby units, including relative $(x, y)$ positions and other attributes. The \textbf{action space} is discrete and consists of movement in four directions, attacking visible enemies, stopping, and a no-op action (only available to dead units). The \textbf{reward function} is summarized in Table~\ref{tab:smac-reward}, and our experiments focus on the sparse reward setting. We further evaluate on SMACv2~\cite{ellis2023smacv2}, an improved variant that introduces randomized unit compositions and initial positions, making coordination more challenging. The state, observation, and action spaces follow the same structure as SMAC, and we evaluate on three scenarios (\texttt{zerg\_5\_vs\_5}, \texttt{terran\_5\_vs\_5}, \texttt{protoss\_5\_vs\_5}), with unit compositions provided in Table~\ref{tab:smac-conf}.

\begin{table}[!h]
\caption{SMAC and SMACv2 scenario configuration}
\centering
\vspace{1em}
\begin{tabular}{@{}llllll@{}}
\toprule
\textbf{Scenario} & \textbf{Ally} & \textbf{Enemy} & \textbf{State Dim} & \textbf{Obs Dim} & \textbf{Action Dim} \\ \midrule
\texttt{3s\_vs\_5z} & 3 Stalkers & 5 Zealots & 68 & 48 & 11 \\ \midrule
\texttt{corridor} & 6 Zealots & 24 Zerglings & 282 & 156 & 30 \\ \midrule
\multirow{3}{*}{\texttt{MMM2}} & 1 Medivac, & 1 Medivac, & \multirow{3}{*}{322} & \multirow{3}{*}{176} & \multirow{3}{*}{18} \\
& 2 Marauders, & 3 Marauders, & & & \\
& 7 Marines & 8 Marines & & & \\ \midrule
\texttt{6h\_vs\_8z} & 6 Hydralisks & 8 Zealots & 140 & 78 & 14 \\ \midrule
\multirow{2}{*}{\texttt{3s5z\_vs\_3s6z}} & 3 Stalkers, & 3 Stalkers, & \multirow{2}{*}{230} & \multirow{2}{*}{136} & \multirow{2}{*}{15} \\
& 5 Zealots & 6 Zealots & & & \\ \midrule
\texttt{zerg\_5\_vs\_5}    & 5 Zerg units    & 5 Zerg units    & 65 & 82 & 11 \\ \midrule
\texttt{terran\_5\_vs\_5}  & 5 Terran units  & 5 Terran units  & 65 & 82 & 11 \\ \midrule
\texttt{protoss\_5\_vs\_5} & 5 Protoss units & 5 Protoss units & 75 & 92 & 11 \\
\bottomrule
\end{tabular}
\label{tab:smac-conf}
\vspace{-1em}
\end{table}

\newpage
\vspace{-0.5em}
\begin{table}[!t]
\caption{Sparse reward setting in SMAC and SMACv2}
\centering
\vspace{1em}
\begin{tabular}{@{}ll@{}}
\toprule
\textbf{Event} & \textbf{Reward} \\ \midrule
Winning the battle & +1 at episode end \\
Losing the battle & -1 at episode end \\ \bottomrule
\end{tabular}
\vspace{-0.5em}
\label{tab:smac-reward}
\end{table}

\vspace{-0.5em}
\begin{figure}[!h]
  \centering
  \begin{subfigure}[T]{0.31\linewidth}
    \centering
    \includegraphics[width=\linewidth]{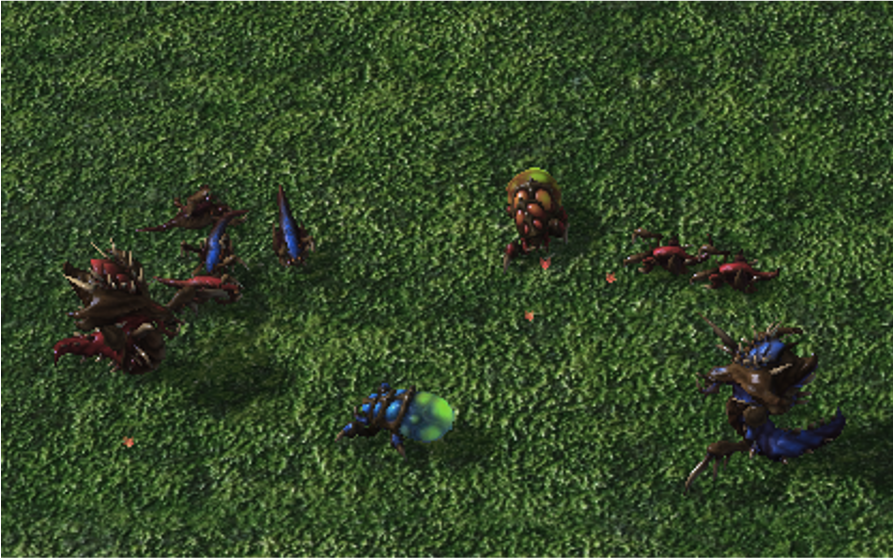}
    \caption{\texttt{zerg\_5\_vs\_5}}
  \end{subfigure}
  \hfill
  \begin{subfigure}[T]{0.31\linewidth}
    \centering
    \includegraphics[width=\linewidth]{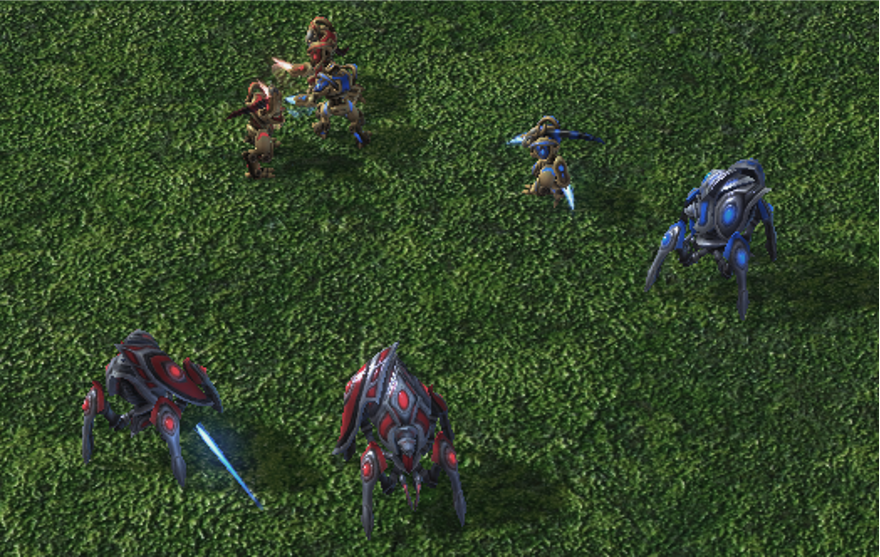}
    \caption{\texttt{protoss\_5\_vs\_5}}
  \end{subfigure}
  \hfill
  \begin{subfigure}[T]{0.31\linewidth}
    \centering
    \includegraphics[width=\linewidth]{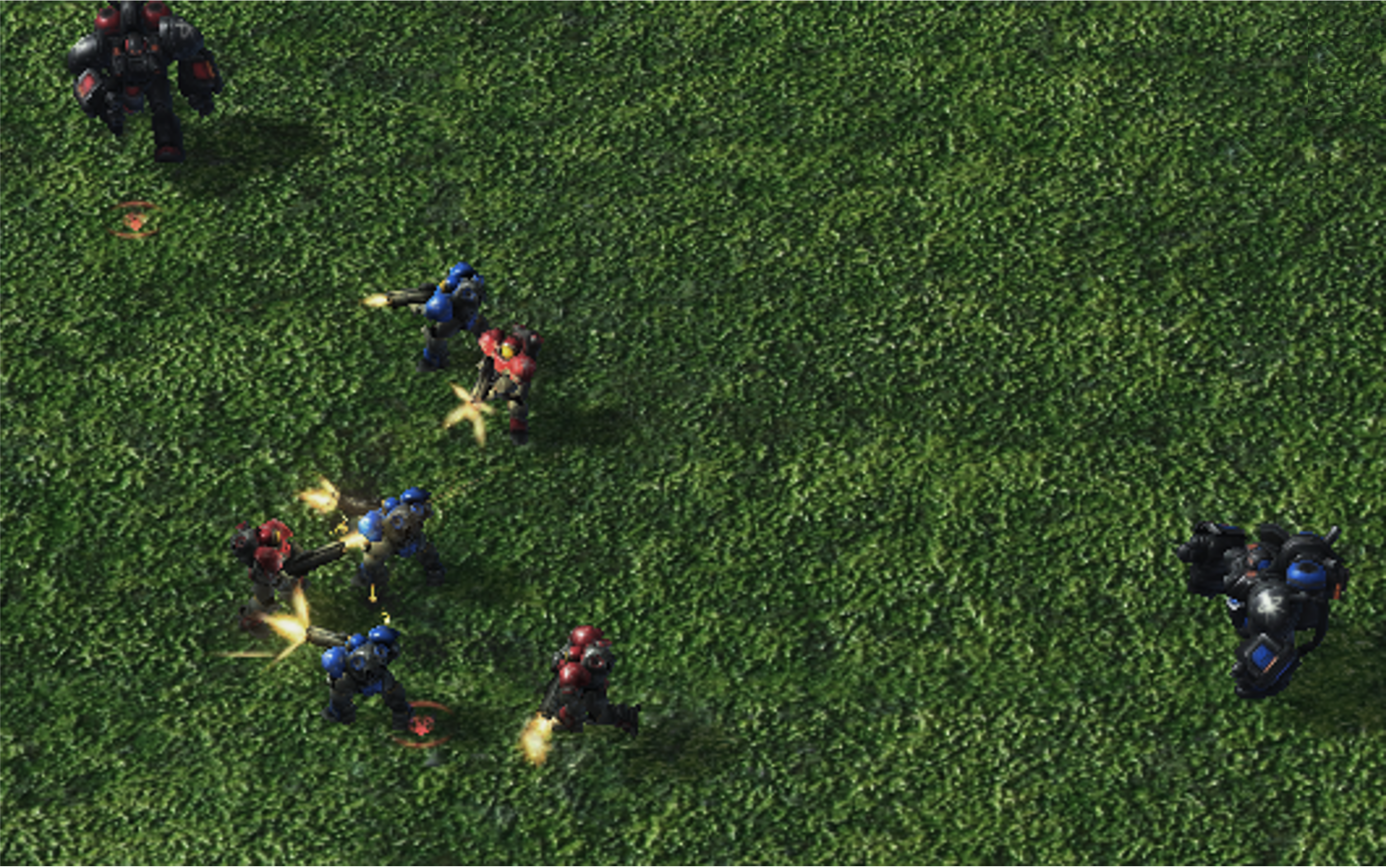}
    \caption{\texttt{terran\_5\_vs\_5}}
  \end{subfigure}
  \caption{Visualization SMACv2 scenarios.}
  \label{fig:smacv2-vis}
\end{figure}

\begin{figure}[!h]
  \centering
  \begin{subfigure}[t]{0.22\linewidth}
    \centering
    \includegraphics[width=1\linewidth]{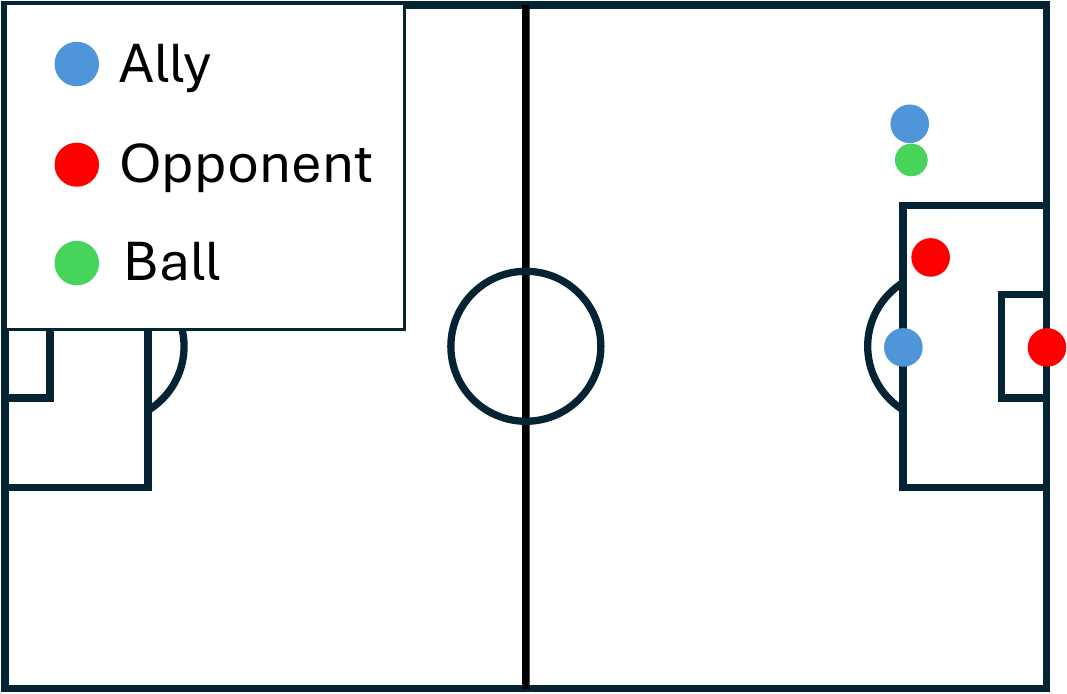}
    \caption{\texttt{academy\_2\_vs\_2}}
  \end{subfigure}
  \hfill
  \begin{subfigure}[t]{0.22\linewidth}
    \centering
    \includegraphics[width=1\linewidth]{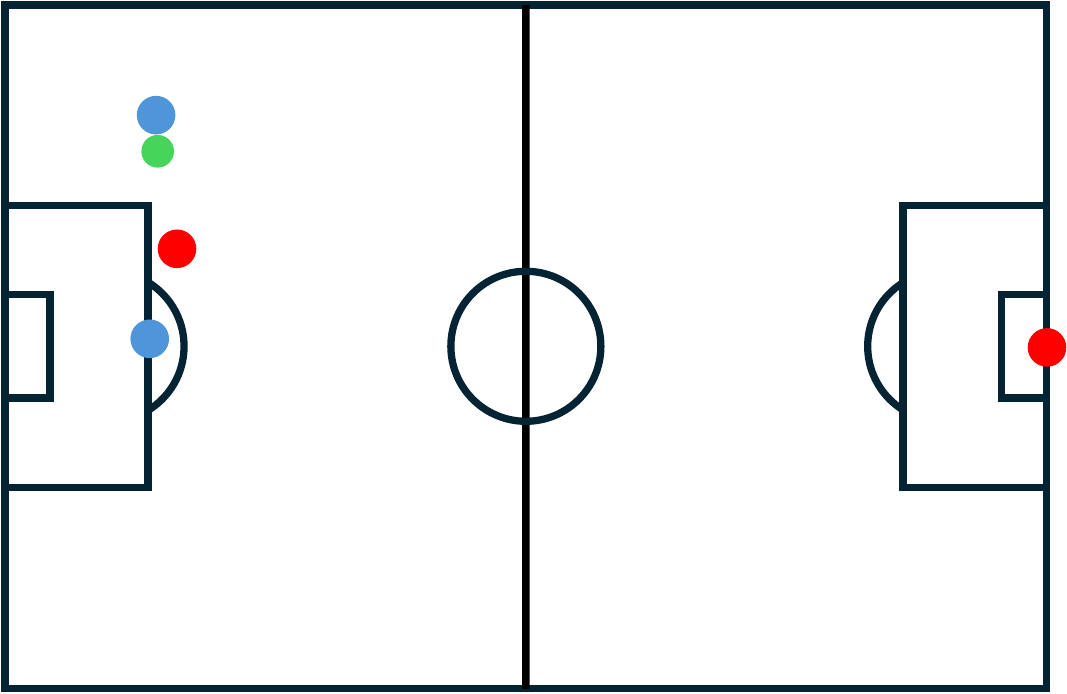}
    \caption{\texttt{academy\_2\_vs\_2\_} \texttt{full\_field}}
  \end{subfigure}
  \hfill
  \begin{subfigure}[t]{0.22\linewidth}
    \centering
    \includegraphics[width=1\linewidth]{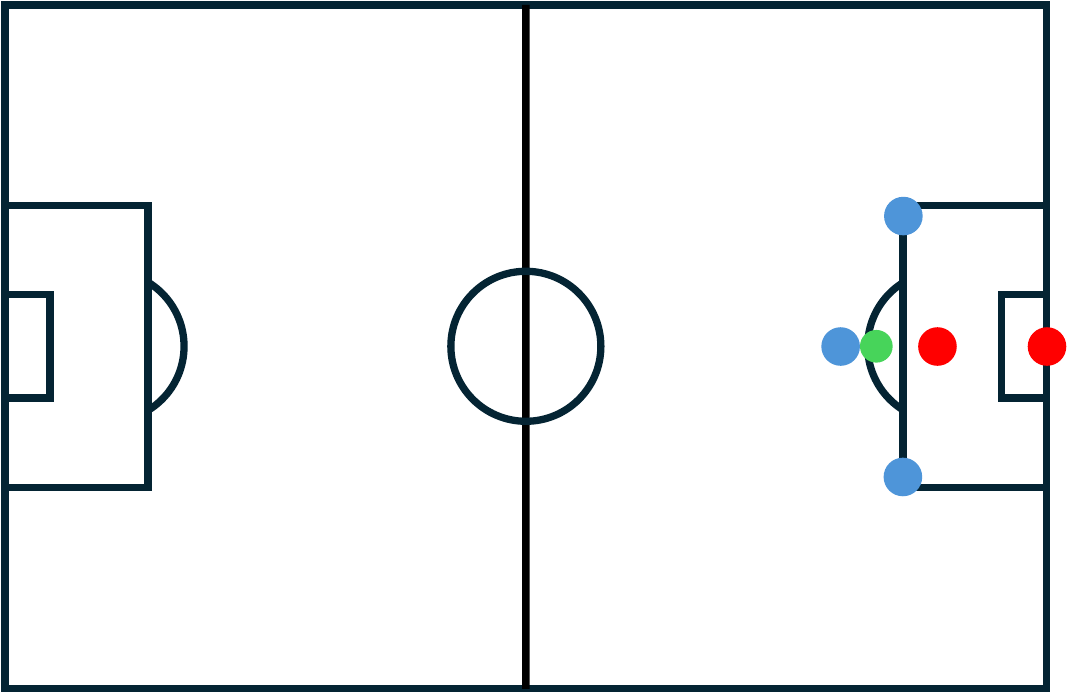}
    \caption{\texttt{academy\_3\_vs\_2}}
  \end{subfigure}
  \hfill
  \begin{subfigure}[t]{0.22\linewidth}
    \centering
    \includegraphics[width=1\linewidth]{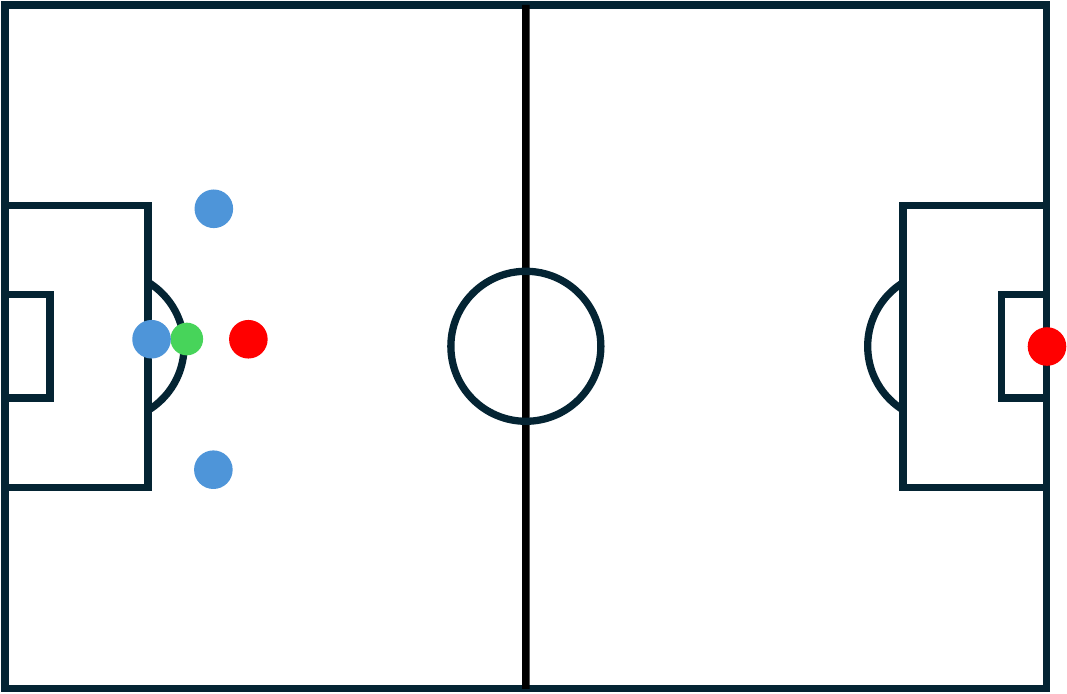}
    \caption{\texttt{academy\_3\_vs\_2\_} \texttt{full\_field}}
  \end{subfigure}
    \begin{subfigure}[t]{0.22\linewidth}
    \centering
    \includegraphics[width=1\linewidth]{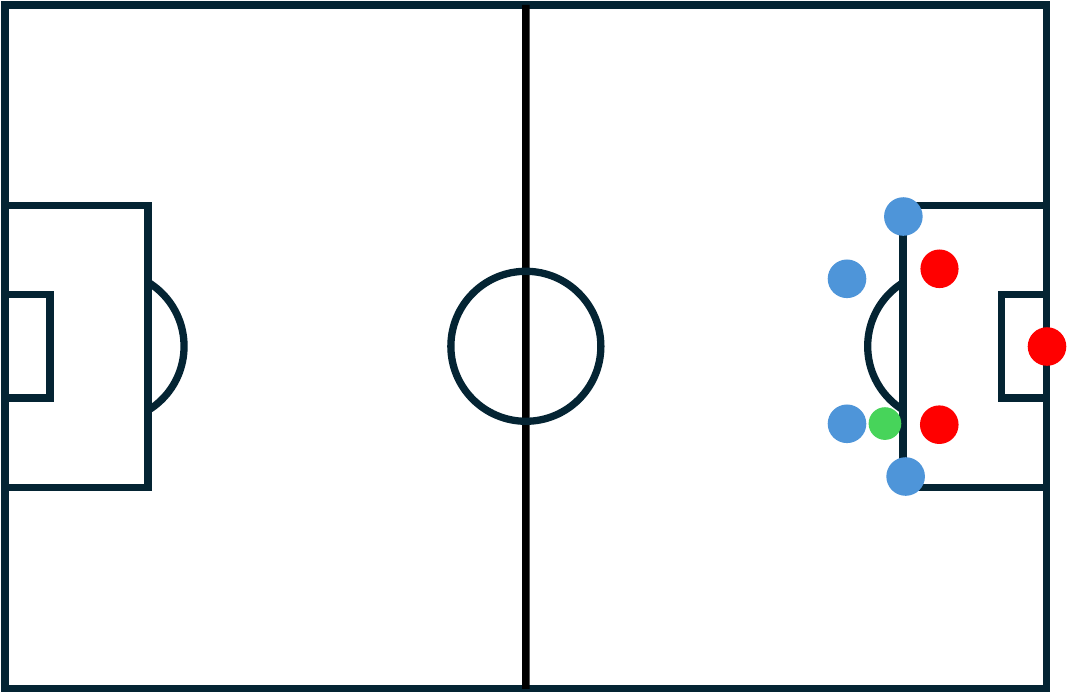}
    \caption{\texttt{academy\_4\_vs\_3}}
  \end{subfigure}
  \hfill
  \begin{subfigure}[t]{0.22\linewidth}
    \centering
    \includegraphics[width=1\linewidth]{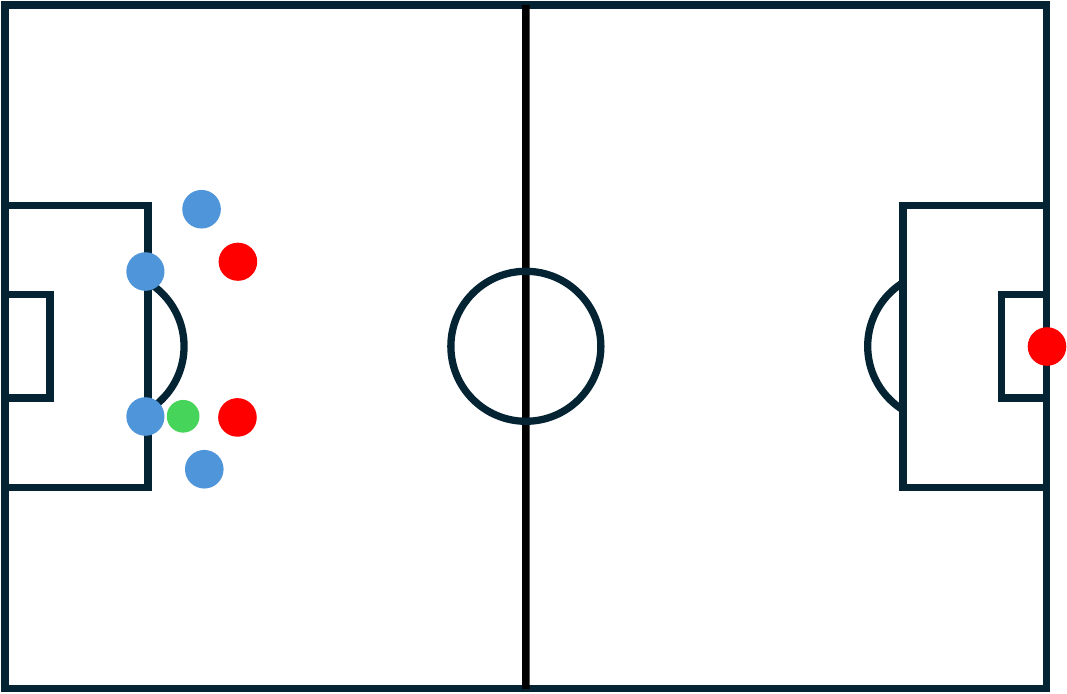}
    \caption{\texttt{academy\_4\_vs\_3\_} \texttt{full\_field}}
  \end{subfigure}
  \hfill
  \begin{subfigure}[t]{0.22\linewidth}
    \centering
    \includegraphics[width=1\linewidth]{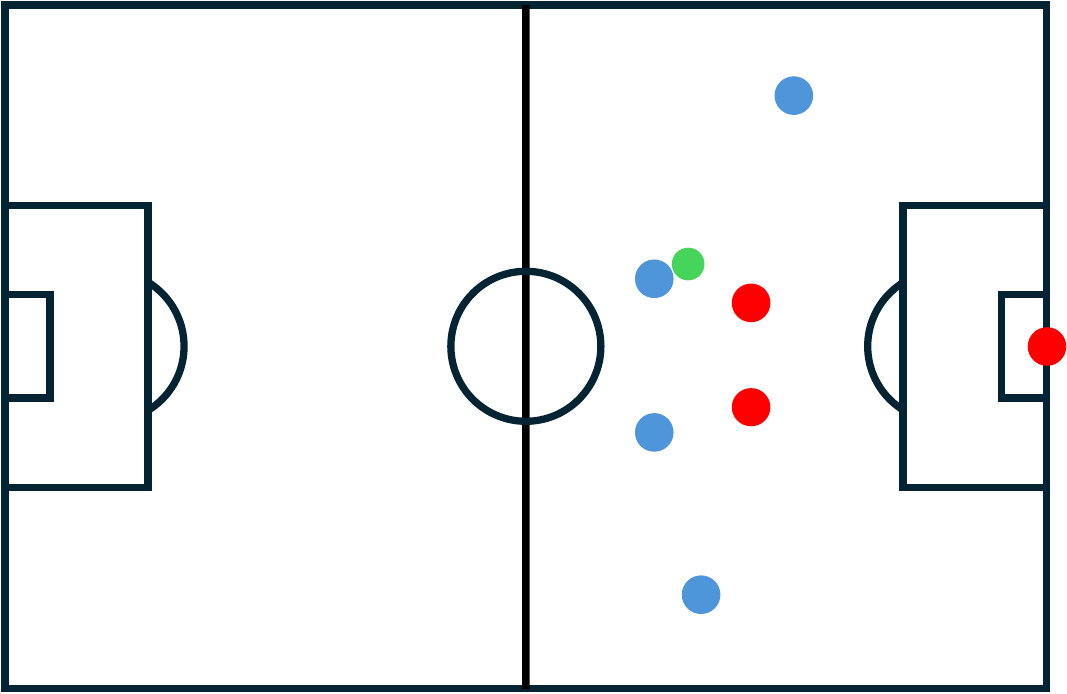}
    \caption{\texttt{academy\_counter} \texttt{attack}}
  \end{subfigure}
  \hfill
  \begin{subfigure}[t]{0.22\linewidth}
    \centering
    \includegraphics[width=1\linewidth]{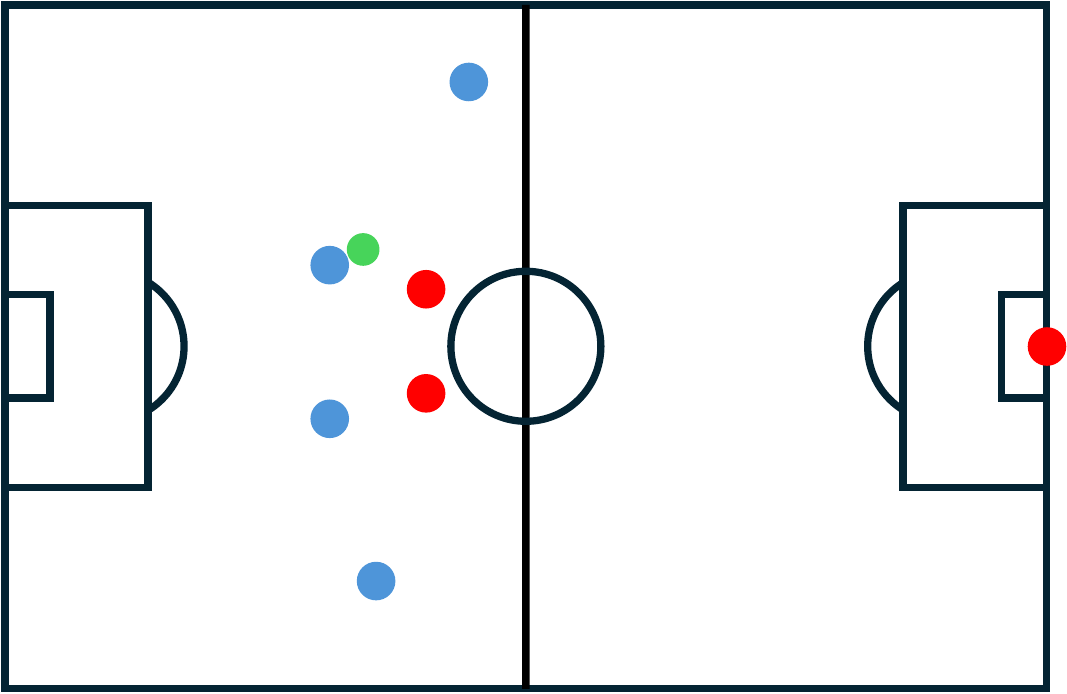}
    \caption{\texttt{academy\_counter} \texttt{attack\_full\_field}}
  \end{subfigure}
  \caption{Visualization of initial agent positions in GRF scenarios.}
  \label{fig:GRF-maps}
\end{figure}

\vspace{-1em}
\paragraph{Google Research Football (GRF):} GRF~\cite{kurach2020google} provides a soccer simulation, in which agents must cooperate to score goals against an opponent team controlled by a scripted AI. We adopt a sparse reward setting to evaluate cooperative behavior under severely limited feedback. The \textbf{state space} consists of player positions and velocities, as well as ball position and velocity. Each \textbf{observation space} for an agent includes local information about the ego player, other players, and ball-related features, all expressed relative to the agent’s current frame. The \textbf{action space} is discrete, covering movement in eight directions, sliding, passing, shooting, sprinting, and standing still. The \textbf{reward function} is described in Table~\ref{tab:grf-reward}. For brevity, several GRF scenarios are referred to using abbreviated names: \texttt{academy\_3\_vs\_2} refers to \texttt{academy\_3\_vs\_1\_with\_keeper}, \texttt{academy\_2\_vs\_2} to \texttt{academy\_run\_pass\_and\_shoot\_with\_keeper}, \texttt{academy\_counterattack} to \texttt{academy\_counterattack\_hard}, and \texttt{academy\_4\_vs\_3} to \texttt{academy\_4\_vs\_2\_with\_keeper} in the original GRF environment. As shown in Fig.~\ref{fig:GRF-maps}, we design full-field variants of these scenarios by repositioning players to opposite half of the court. Table~\ref{tab:grf-conf} provides
an overview of the unit configurations.

\newpage
\begin{table}[!h]
    \caption{Reward setting in GRF}
    \vspace{1em}
    \centering
    \begin{tabular}{@{}ll@{}}
    \toprule
    \textbf{Event} & \textbf{Reward} \\ \midrule
    Winning the match & +100 at episode end \\
    Losing the match & -1 at episode end \\
    \bottomrule
    \end{tabular}
    \label{tab:grf-reward}
\end{table}

\begin{table}[!h]
\caption{GRF scenario configuration}
\centering
\vspace{1em}
\begin{tabular}{@{}llllll@{}}
\toprule
\textbf{Scenario} & \textbf{Ally} & \textbf{Opponent} & \textbf{State Dim} & \textbf{Obs Dim} & \textbf{Action Dim} \\ \midrule
\multirow{2}{*}{\texttt{academy\_2\_vs\_2}} & \multirow{2}{*}{2 center back} & \multirow{2}{*}{\shortstack[l]{1 goalkeeper,\\1 center back}} & \multirow{2}{*}{22} & \multirow{2}{*}{22} & \multirow{2}{*}{19} \\
&&&&& \\
\midrule
\multirow{2}{*}{\shortstack[l]{\texttt{academy\_2\_vs\_2}\\\texttt{\_full\_field}}} & \multirow{2}{*}{2 center back} & \multirow{2}{*}{\shortstack[l]{1 goalkeeper,\\1 center back}} & \multirow{2}{*}{22} & \multirow{2}{*}{22} & \multirow{2}{*}{19} \\
&&&&& \\
\midrule
\multirow{2}{*}{\texttt{academy\_3\_vs\_2}} & \multirow{2}{*}{3 central midfield} & \multirow{2}{*}{\shortstack[l]{1 goalkeeper,\\1 center back}} & \multirow{2}{*}{26} & \multirow{2}{*}{26} & \multirow{2}{*}{19} \\
& & & & & \\ 
\midrule
\multirow{2}{*}{\shortstack[l]{\texttt{academy\_3\_vs\_2}\\\texttt{\_full\_field}}} & \multirow{2}{*}{3 central midfield} & \multirow{2}{*}{\shortstack[l]{1 goalkeeper,\\1 center back}} & \multirow{2}{*}{26} & \multirow{2}{*}{26} & \multirow{2}{*}{19} \\
& & & & & \\ 
\midrule
\multirow{2}{*}{\texttt{academy\_4\_vs\_3}} & \multirow{2}{*}{4 central midfield} & \multirow{2}{*}{\shortstack[l]{1 goalkeeper,\\2 center back}} & \multirow{2}{*}{34} & \multirow{2}{*}{34} & \multirow{2}{*}{19} \\
& & & & & \\ 
\midrule
\multirow{2}{*}{\shortstack[l]{\texttt{academy\_4\_vs\_3}\\\texttt{\_full\_field}}} & \multirow{2}{*}{4 central midfield} & \multirow{2}{*}{\shortstack[l]{1 goalkeeper,\\2 center back}} & \multirow{2}{*}{34} & \multirow{2}{*}{34} & \multirow{2}{*}{19} \\
& & & & & \\ 
\midrule
\multirow{4}{*}{\shortstack[l]{\texttt{academy}\\\texttt{\_counterattack}}} & \multirow{4}{*}{\shortstack[l]{1 central midfield,\\1 left midfield,\\1 right midfield,\\1 central front}} & \multirow{4}{*}{\shortstack[l]{1 goalkeeper,\\2 center back}} & \multirow{4}{*}{34} & \multirow{4}{*}{34} & \multirow{4}{*}{19} \\
& & & & & \\ 
& & & & & \\ 
& & & & & \\ 
\midrule
\multirow{4}{*}{\shortstack[l]{\texttt{academy}\\\texttt{\_counterattack}\\\texttt{\_full\_field}}} & \multirow{4}{*}{\shortstack[l]{1 central midfield,\\1 left midfield,\\1 right midfield,\\1 central front}} & \multirow{4}{*}{\shortstack[l]{1 goalkeeper,\\2 center back}} & \multirow{4}{*}{34} & \multirow{4}{*}{34} & \multirow{4}{*}{19} \\
& & & & & \\ 
& & & & & \\ 
& & & & & \\ 
\midrule
\bottomrule
\end{tabular}
\label{tab:grf-conf}
\end{table}

\newpage
\section{Experimental Details}
\vspace{-0.5em}
\label{secapp:exp-details}
FIM is implemented on top of the open-source framework from~\cite{hu2021rethinking}, which is also used to run QMIX~\cite{rashid2018qmix} and QPLEX~\cite{wangqplex}. LAIES~\cite{liu2023lazy}, RODE~\cite{wang2021rode}, MASER~\cite{jeon2022maser}, CDS~\cite{li2021celebrating}, and FoX~\cite{jo2024fox} are evaluated using the original code and settings provided by their respective authors. Experiments are conducted on an NVIDIA RTX 3090 GPU with an Intel Xeon Gold 6348 CPU (Ubuntu 20.04). Training completes within two days for Push-2-Box and SMAC, while each GRF scenario requires less than two days to reach 5 million timesteps. We begin by describing the baseline algorithms in Appendix~\ref{subsecapp:baseline} and outline the hyperparameter setup of FIM in Appendix~\ref{subsecapp:hyperparam}.

\vspace{-0.5em}
\subsection{Detailed Description of Baseline Algorithms}
\label{subsecapp:baseline}
\begin{itemize}
    \item \textbf{QMIX}~\cite{rashid2018qmix} factorizes the joint action-value into individual utilities combined by a monotonic mixing network. Code: \url{https://github.com/hijkzzz/pymarl2}
    \item \textbf{QPLEX}~\cite{wangqplex} extends QMIX with a duplex dueling architecture, decomposing joint value into individual value and advantage while enforcing the IGM principle. Code: \url{https://github.com/hijkzzz/pymarl2}
    \item \textbf{LAIES}~\cite{liu2023lazy} incentivizes agents to influence external task-relevant states via intrinsic rewards for both individual and joint impacts. Code: \url{https://github.com/liuboyin/LAIES}
    
    \item \textbf{RODE}~\cite{wang2021rode} employs hierarchical role-based policies where agents periodically select roles to guide low-level actions, enabling scalable specialization. Code: \url{https://github.com/TonghanWang/RODE}
    
    \item \textbf{MASER}~\cite{jeon2022maser} explores by assigning subgoals from past trajectories, rewarding agents for revisiting informative states. Code: \url{https://github.com/Jiwonjeon9603/MASER}
    
    \item \textbf{CDS}~\cite{li2021celebrating} encourages policy diversity by maximizing mutual information between agent identity and trajectory. Code: \url{https://github.com/lich14/CDS}
    
    \item \textbf{FoX}~\cite{jo2024fox} promotes exploration by maximizing entropy of agent formations and their mutual information with team structure. Code: \url{https://github.com/hyeon1996/FoX}

    \item \textbf{COMA}~\cite{foerster2018counterfactual} is a multi-agent policy gradient method that assigns credit using a centralized critic with counterfactual baselines. Code: \url{https://github.com/oxwhirl/pymarl2}

    \item \textbf{MA}$^2$\textbf{E}~\cite{kangma} addresses partial observability with masked autoencoder to infer global information from individual observations. Code: \url{https://github.com/cheesebro329/MA2E}
\end{itemize}

\subsection{Hyperparameter Setup of the Proposed FIM}
\label{subsecapp:hyperparam}

Scenario-specific hyperparameter of intrinsic reward weight $\alpha$ and update rate $\phi$ is provided in Table~\ref{tab:scenario-hyperparams}, while the softmax temperature in $\mathrm{Softmax}(-\mathcal{H}(d))$ is set to $0.1$, and the discount factor $\gamma$ is fixed at $0.99$ across all scenarios. The default hyperparameter settings of FIM, which are generally shared across scenarios, are summarized in Table~\ref{tab:hyperparams}. 

\begin{table}[!h]
\caption{Common hyperparameter setting of FIM}
\vspace{1em}
\centering
\begin{tabular}{lc}
\toprule
\textbf{Hyperparameters} & Value \\ \hline
Optimizer & Adam  \\
Replay buffer size & 5000 \\
Target update interval & 200 \\
Mini-batch size & 32 \\
Mixing network dim & 32 \\
Discount factor $\gamma$ & 0.99 \\
Learning rate & 0.0005 \\
Dynamics model $\hat{s}(\cdot)$ layer & 3 \\
Dynamics model $\hat{s}(\cdot)$ dim  & 128 \\
\bottomrule
\\[0.3em]
\end{tabular}
\label{tab:hyperparams}
\end{table}

\newpage
\begin{table}[!h]
\caption{Scenario specific hyperparameter setup of FIM}
\vspace{1em}
\centering
\begin{tabular}{lcc}
\toprule
\textbf{Scenario} & $\boldsymbol{\alpha}$ & $\boldsymbol{\phi}$ \\
\midrule
Push-2-Box & 5 & 0.05 \\
\midrule
\multicolumn{3}{l}{\textbf{StarCraft Multi-agent Challenge (Sparse)}} \\
\texttt{3s\_vs\_5z} & 10 & 0.05 \\
\texttt{corridor} & 20 & 0.05 \\
\texttt{MMM2} & 10 & 0.05 \\
\texttt{6h\_vs\_8z} & 20 & 0.05 \\
\texttt{3s5z\_vs\_3s6z} & 10 & 0.05 \\
\midrule
\multicolumn{3}{l}{\textbf{StarCraft Multi-agent Challenge v2 (Sparse)}} \\
\texttt{protoss\_5\_vs\_5} & 10 & 0.05 \\
\texttt{terran\_5\_vs\_5} & 10 & 0.05 \\
\texttt{zerg\_5\_vs\_5} & 10 & 0.05 \\
\midrule
\multicolumn{3}{l}{\textbf{Google Research Football (Sparse)}} \\
\texttt{academy\_2\_vs\_2} & 10 & 0.05 \\
\texttt{academy\_2\_vs\_2\_full\_field} & 10 & 0.05 \\
\texttt{academy\_3\_vs\_2} & 10 & 0.05 \\
\texttt{academy\_3\_vs\_2\_full\_field} & 10 & 0.05 \\
\texttt{academy\_4\_vs\_3} & 10 & 0.05 \\
\texttt{academy\_4\_vs\_3\_full\_field} & 10 & 0.05 \\
\texttt{academy\_counterattack} & 10 & 0.05 \\
\texttt{academy\_counterattack\_full\_field} & 10 & 0.05 \\
\bottomrule
\\[0.3em]
\end{tabular}
\vspace{-1.5em}
\label{tab:scenario-hyperparams}
\end{table}

\newpage
\section{Additional Experiments}
\vspace{-0.5em}
In this section, We present additional experiments to further validate the robustness and generality of FIM.  Specifically, we visualize the dimension weights $w_d$ in Appendix~\ref{subsecapp:w_d-vis} and evaluate FIM on Overcooked in Appendix~\ref{subsecapp:overcooked}.

\subsection{Visualization of SFI Weights $w_d$}
\vspace{-0.5em}
\label{subsecapp:w_d-vis}

\begin{figure}[!h]
  \centering
  \begin{subfigure}[T]{\linewidth}
    \centering
    \includegraphics[width=\linewidth]{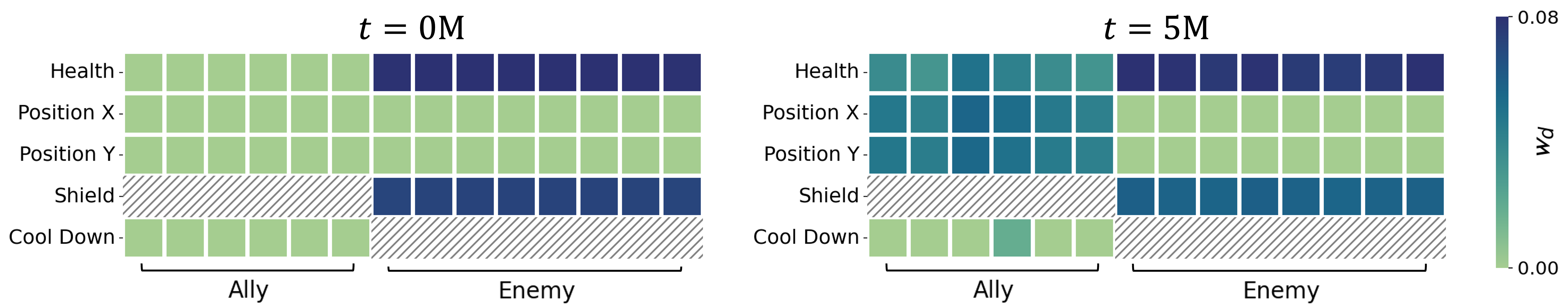}
    \caption{\texttt{6h\_vs\_8z}}
  \end{subfigure}
  \vfill
  \begin{subfigure}[T]{\linewidth}
    \centering
    \includegraphics[width=\linewidth]{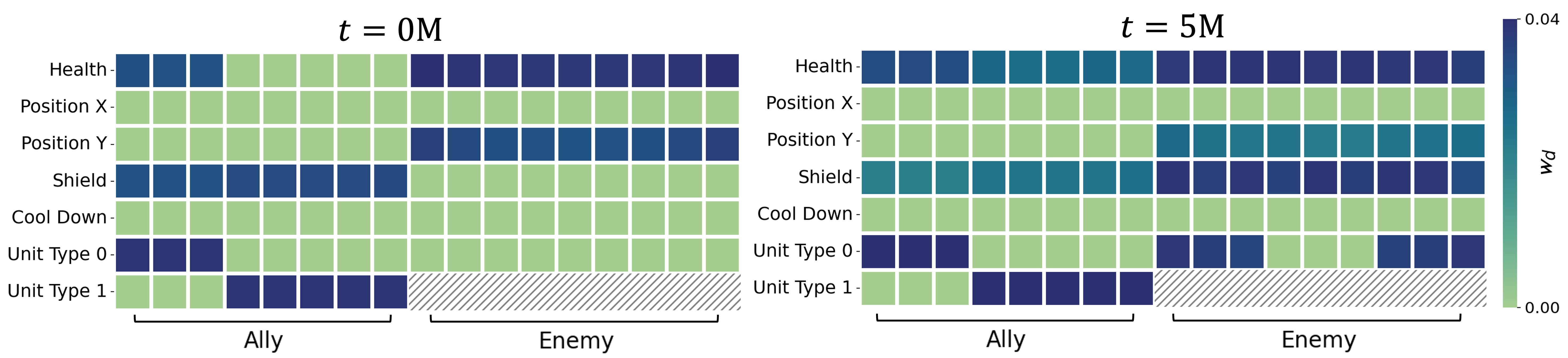}
    \caption{\texttt{3s5z\_vs\_3s6z}}
  \end{subfigure}
  \caption{SMAC $w_d$ visualization}
  \label{fig:smac-wd}
\end{figure}

\begin{figure}[!h]
  \centering
  \begin{subfigure}[T]{\linewidth}
    \centering
    \includegraphics[width=0.85\linewidth]{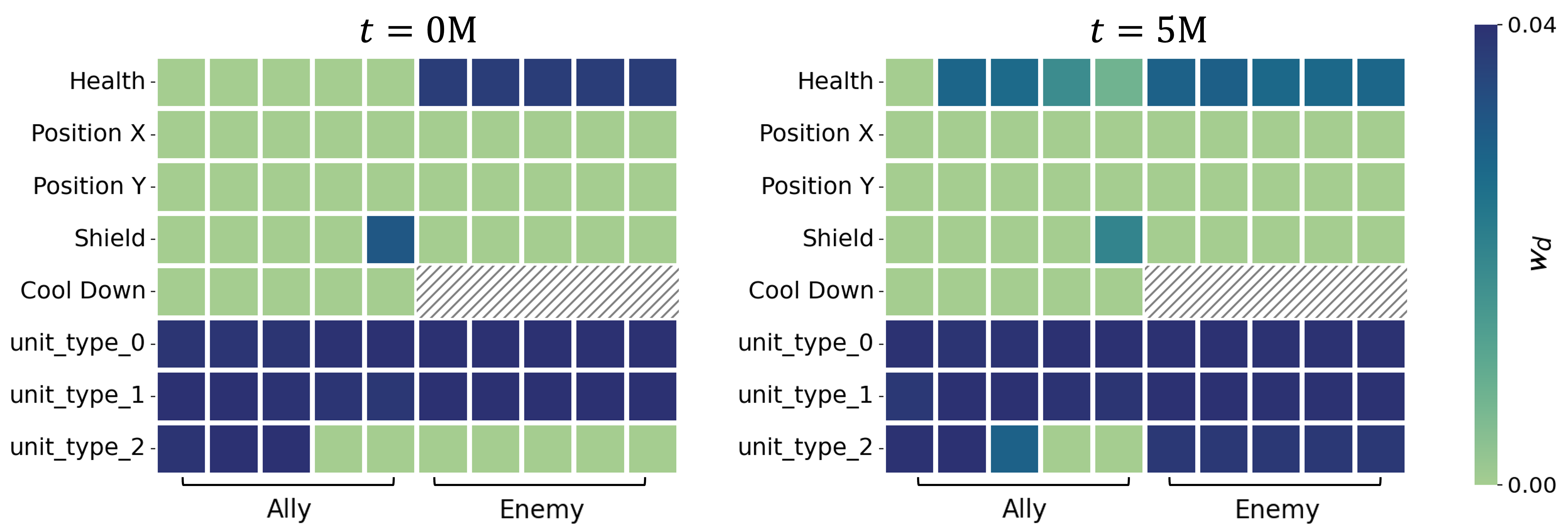}
    \caption{\texttt{protoss\_5\_vs\_5}}
  \end{subfigure}
  \vfill
  \begin{subfigure}[T]{\linewidth}
    \centering
    \includegraphics[width=0.85\linewidth]{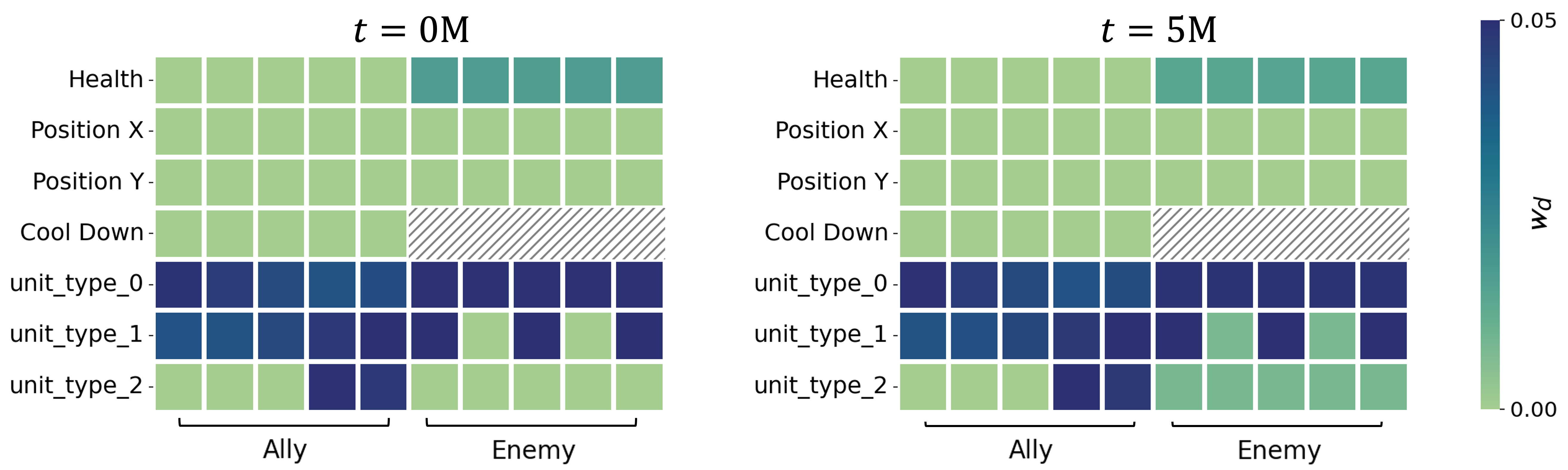}
    \caption{\texttt{terran\_5\_vs\_5}}
  \end{subfigure}
  \caption{SMACv2 $w_d$ visualization}
  \label{fig:smacv2-wd}
\end{figure}

Fig.~\ref{fig:smac-wd} and Fig.~\ref{fig:smacv2-wd} visualize $w_d$ changes as training proceeds in SMAC and SMACv2, respectively. Across both benchmarks, SFI consistently assigns the highest weights to enemy health dimensions throughout training, reflecting their status as under-explored state dimensions that change substantially only when multiple agents coordinate their attacks on the same target. This makes enemy health a natural focal point for joint influence, as reducing it requires sustained and synchronized action from the team. Ally-specific features such as ally health also receive progressively elevated weights as training advances. This is a consequence of agents' tendency to avoid direct combat in order to preserve their own health, which causes ally health dimensions to remain relatively stable and thus under-explored. By up-weighting these dimensions, SFI counteracts this avoidance behavior and encourages agents to engage more actively in battle, ultimately promoting more effective coordination.

\begin{figure}[!h]
  \centering
  \begin{subfigure}[T]{\linewidth}
    \centering
    \includegraphics[width=0.7\linewidth]{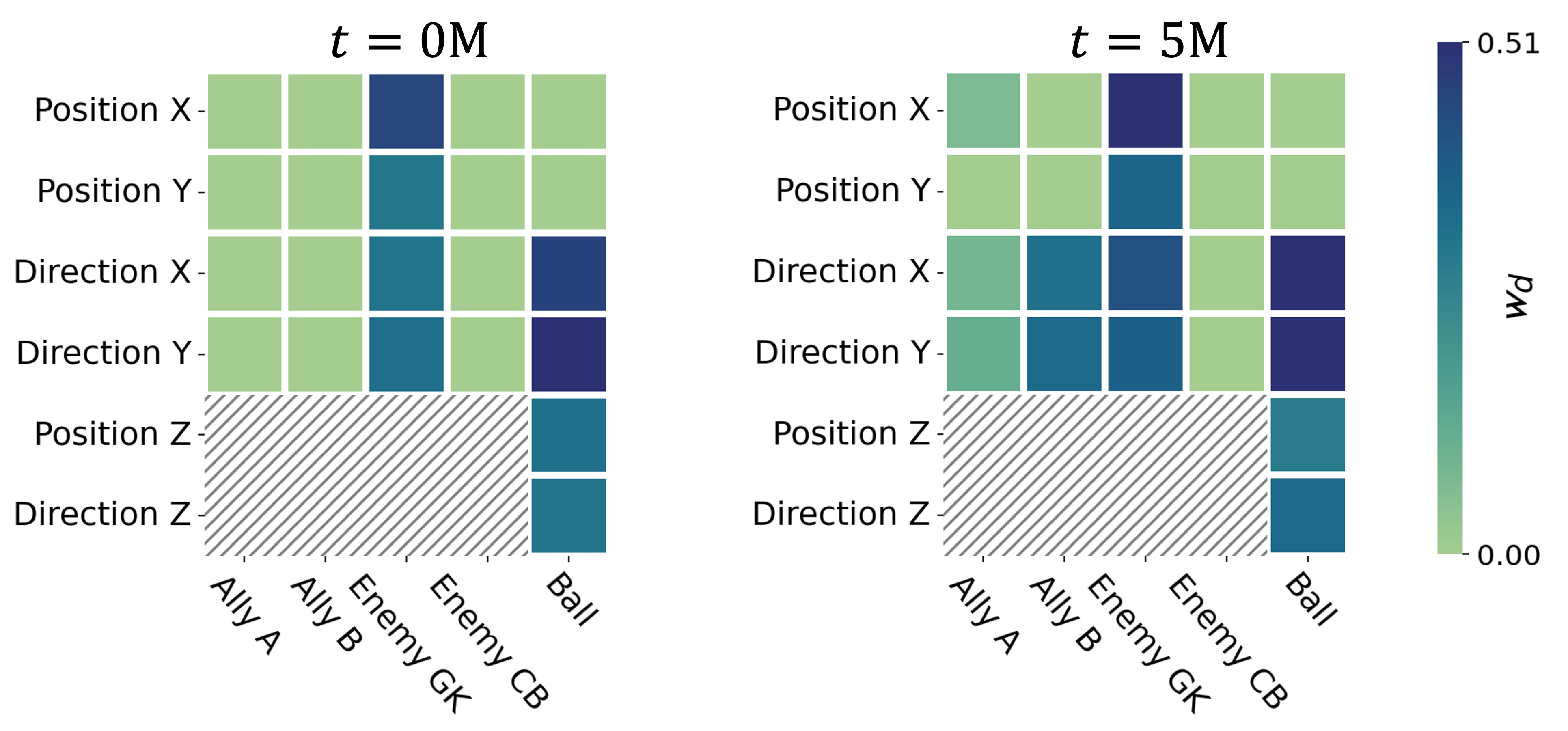}
    \caption{\texttt{academy\_2\_vs\_2}}
  \end{subfigure}
  \vfill
  \begin{subfigure}[T]{\linewidth}
    \centering
    \includegraphics[width=0.7\linewidth]{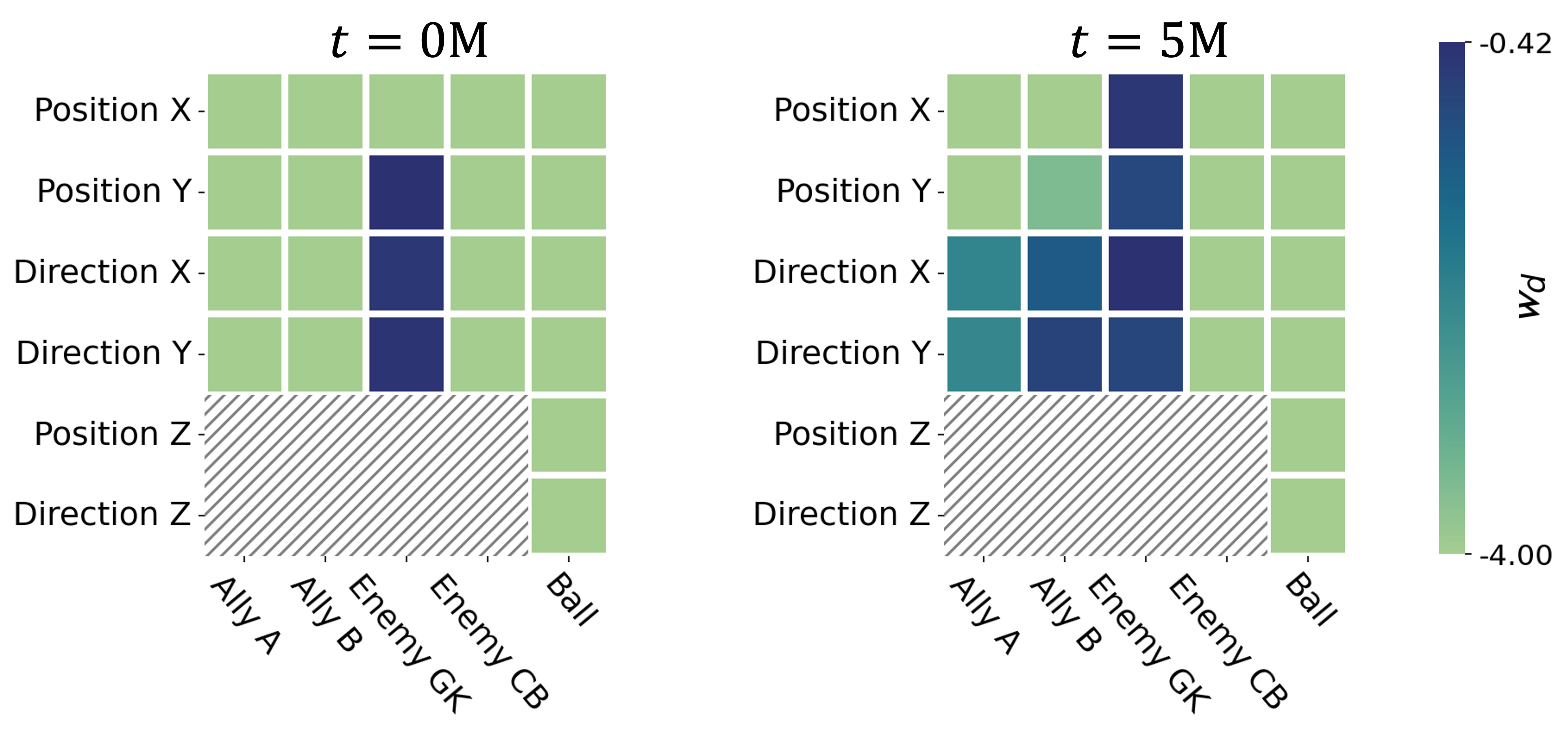}
    \caption{\texttt{academy\_2\_vs\_2\_full\_field}}
  \end{subfigure}
  \caption{GRF $w_d$ visualization}
  \label{fig:grf-wd}
\end{figure}

Fig.~\ref{fig:grf-wd} visualizes $w_d$ in GRF, where the opponent goalkeeper's position and direction dimensions are consistently assigned the highest weights throughout training. The goalkeeper's state exhibits persistently low entropy because it remains largely stationary under the joint behavior policy, shifting only in response to direct pressure from attacking agents. Consequently, SFI identifies these dimensions as primary targets, concentrating the team's cooperative effort on a state that is both stable and impactful. A detailed account of how this weight assignment translates into concrete agent behaviors is provided in Appendix~\ref{subsecapp:grf-strategy-analysis}.

\newpage

\newpage
\subsection{Evaluation of FIM on Overcooked}
\vspace{-0.5em}
\label{subsecapp:overcooked}
To evaluate FIM beyond structured vector states, we conduct an additional experiment on Overcooked (coord-ring, JaxMARL implementation).
Unlike structured representations that explicitly provide object coordinates, we construct the state as a height $\times$ width $\times$ components tensor, where the positions of components (e.g., agents, ingredients) are represented via binary encodings over spatial grids.
Even in this unstructured representation, certain dimensions exhibit notably low entropy.
In particular, the soup location remains under-explored since it cannot be moved until it is fully prepared.
As training progresses, agents explore diverse positions with the soup and eventually converge to efficient goal-directed trajectories.
As shown in Table~\ref{tab:overcooked}, FIM, with $\alpha=1$ and $\phi=0.05$, achieves approximately 10\% performance improvement over the IPPO baseline.

\begin{table}[!h]
\caption{Return score on Overcooked coord-ring (JaxMARL implementation).}
\vspace{1em}
\centering
\small
\begin{tabular}{lc}
\toprule
\textbf{Method} & \textbf{Return} \\
\midrule
IPPO & $195.7 \pm 5.3$ \\
IPPO + FIM & $218.5 \pm 4.7$ \\
\bottomrule
\end{tabular}
\label{tab:overcooked}
\end{table}

\newpage
\section{Additional Analysis and Ablations Studies}
\vspace{-0.5em}
In this section, we provide further analyses and ablation studies of FIM. First, we present FIM's strategy analysis in the GRF environment in Appendix~\ref{subsecapp:grf-strategy-analysis}. Next, we provide extended ablation studies that examine the independent contributions of SFI and AFI under various settings in Appendix~\ref{subsecapp:abl}. We then examine the relationship between dynamics model learning and performance improvements in Appendix~\ref{subsecapp:dynamics-model}, and analyze the computational cost of FIM relative to QMIX in Appendix~\ref{subsecapp:complexity}.

\vspace{-0.5em}
\subsection{Strategy Analysis in GRF}
\begin{figure}[!h]
  \centering
  \begin{subfigure}[b]{0.85\linewidth}
    \centering
    \includegraphics[width=\linewidth]{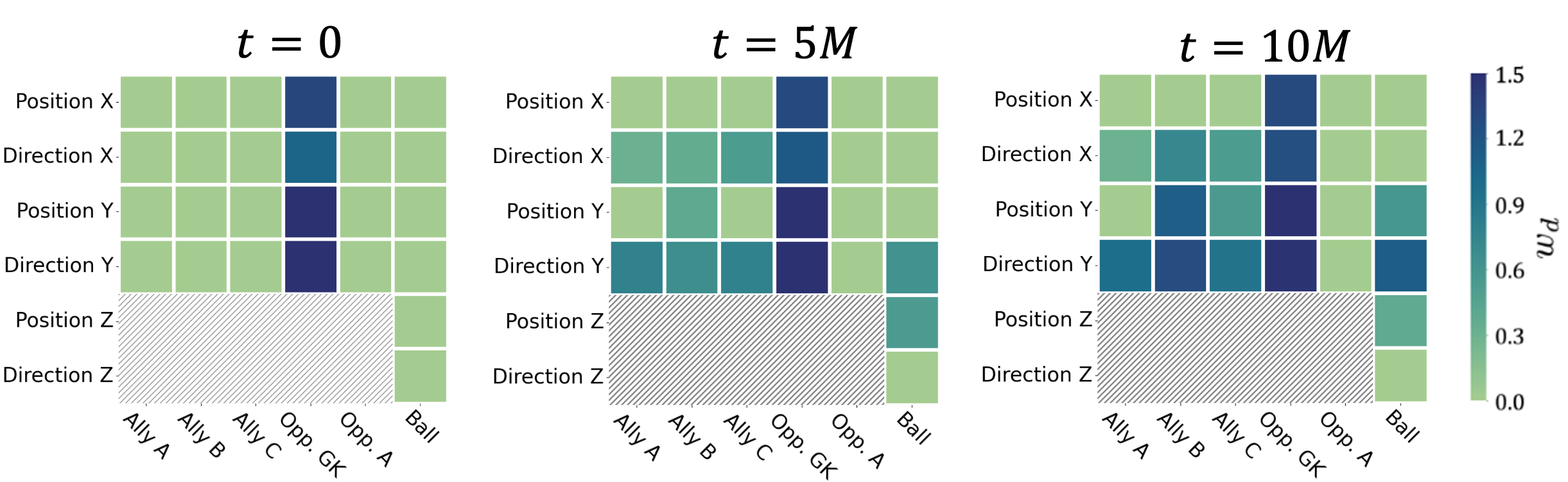} \caption{Dimensional weight $w_d$ across iterations}
  \end{subfigure}  
  \vfill
  \vspace{0.7em}
  \begin{subfigure}[b]{0.85\linewidth}
    \centering
    \includegraphics[width=\linewidth]{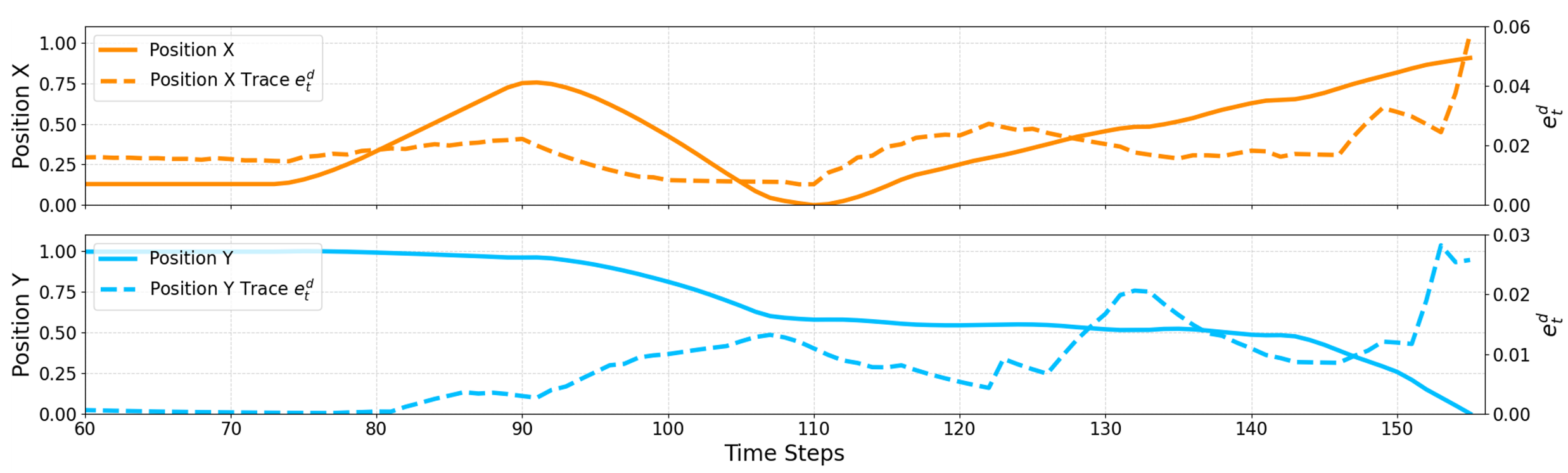}
    \caption{Changes in goalkeeper position and its trace $e^d_t$}
  \end{subfigure}
  \vfill
  \vspace{1em}
  \begin{subfigure}[b]{0.7\linewidth}
    \centering
    \includegraphics[width=\linewidth]{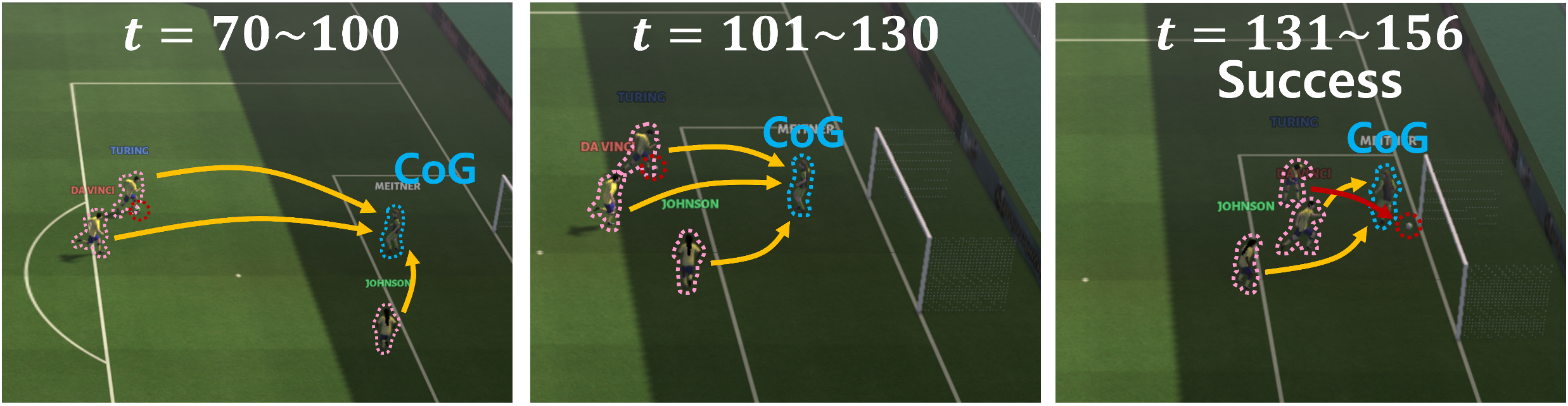}
    \caption{Rendered frames for highlighting agents' coordination}
  \end{subfigure}
  \caption{Analysis of FIM's strategy in GRF \texttt{academy\_3\_vs\_2\_full\_field}} 
  \label{fig:GRF-analysis}
\end{figure}

\vspace{-0.5em}
\label{subsecapp:grf-strategy-analysis}
In GRF, the dimensional weight $w_d$ consistently focus the opponent goalkeeper position and direction across the training, as illustrated in Fig.~\ref{fig:GRF-analysis}(a). These dimensions exhibit low entropy throughout the training, since the goalkeeper typically remains stationary and only shifts position when a ball-carrying agent approaches the goalpost. This characteristic makes the goalkeeper’s state both stable and strategically significant, as displacing it creates scoring opportunities and thus serves as a valuable proxy objective in sparse reward settings. Accordingly, FIM guides agents to influence the goalkeeper’s position.

As shown in Fig.~\ref{fig:GRF-analysis}(b)-(c), this insight is reflected in the agent trajectory. Around $t \approx 70$ within the trajectory, agents begin to receive intrinsic rewards by subtly influencing the goalkeeper’s position, even while positioned far from the goal area. By $t \approx 100$, the accumulated eligibility traces further incentivize agents to continue exerting influence over the goalkeeper, enabling a gradual progression toward the goal. Near $t \approx 130$, the goalkeeper briefly moves out of position, and the attacking agent capitalizes on this opportunity to score. Notably, FIM guides agents to approach the goal proactively and maintain persistent influence over the goalkeeper’s positioning, which is under-explored region for the joint behavior policy.
\vspace{-0.5em}

\vspace{-1em}
\subsection{Extended Ablation Studies}
\label{subsecapp:abl}

To further evaluate the robustness of FIM, we conduct experiments on four challenging scenarios: SMAC \texttt{3s\_vs\_5z}, SMAC \texttt{3s5z\_vs\_3s6z}, GRF \texttt{academy\_3\_vs\_2}, and GRF \texttt{academy\_3\_vs\_2\_full\_field}, focusing on (i) alternatives to the SFI state focusing mechanism, (ii) the impact of each module in FIM through a component ablation study, (iii) the reward scaling factor $\alpha$, and (iv) the update rate $\phi$.

\paragraph{Alternatives to the SFI State Selection Mechanism:} We investigate alternative strategies to the SFI state focusing mechanism for weighting state dimensions to influence: \textit{uniform-weighting influence}, \textit{external state focusing influence} (EFI), and \textit{least-change state focusing influence} (LFI). In all cases, the intrinsic reward is computed as in FIM, with each variant differing only in weighting the state dimensions in $\mathcal{D}$. The uniform-weighting variant set $w_d=1$ for all $d \in \mathcal{D}$, which is equivalent to Eq.~\ref{eq:afi_reward}. EFI only weights external state features, following the approach of LAIES~\cite{liu2023lazy}: enemy health, shield, and positions in SMAC; and opponent and ball positions and directions in GRF. LFI weights state dimensions with the smaller average temporal change $|\tilde{s}^d_{t+1} - \tilde{s}^d_t|$ via softmax. In SMAC, this typically includes enemy health and ally positions, while in GRF, it often corresponds to ally direction features due to their relatively small-scale temporal changes.

\begin{figure}[!h]
  \centering
  \begin{subfigure}[t]{0.24\linewidth}
    \centering
    \includegraphics[width=\linewidth]{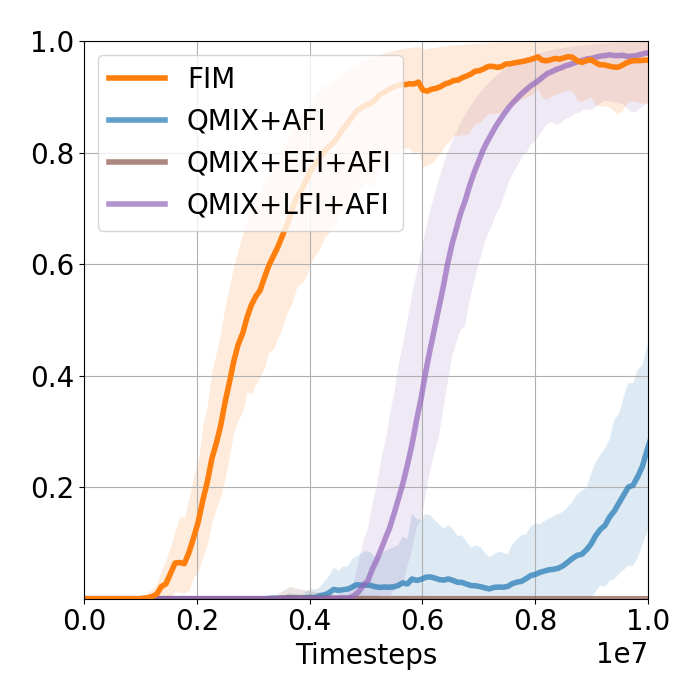}
    \vspace{-2em}
    \caption{\texttt{3s\_vs\_5z}}
  \end{subfigure}
  \hfill
  \begin{subfigure}[t]{0.24\linewidth}
    \centering
    \includegraphics[width=\linewidth]{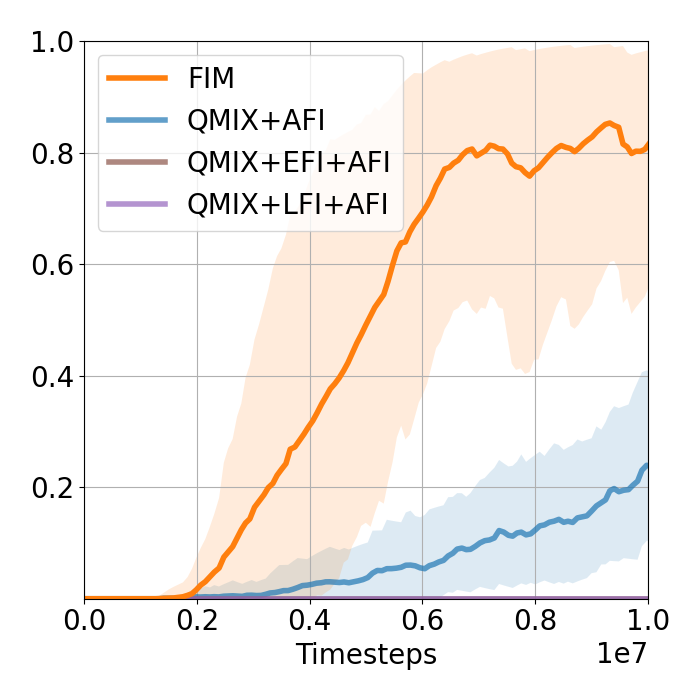}
    \vspace{-2em}
    \caption{\texttt{3s5z\_vs\_3s6z}}
  \end{subfigure}
  \hfill
  \begin{subfigure}[t]{0.24\linewidth}
    \centering
    \includegraphics[width=\linewidth]{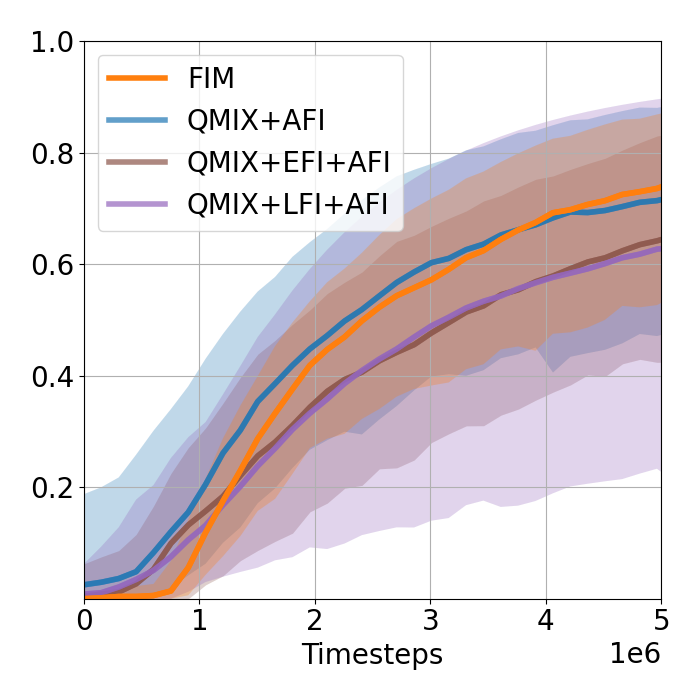}
    \vspace{-2em}
    \caption{\texttt{academy\_3\_vs\_2}}
  \end{subfigure}
  \hfill
  \begin{subfigure}[t]{0.24\linewidth}
    \centering
    \includegraphics[width=\linewidth]{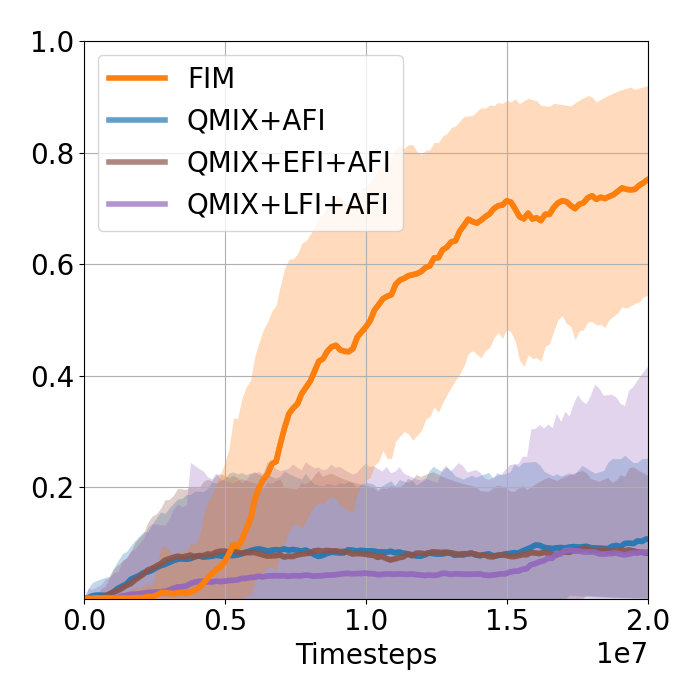}
    \vspace{-2em}
    \caption{\texttt{academy\_3\_vs\_2\_fu} \texttt{ll\_field}}
  \end{subfigure}
  \vspace{-0.5em}
  \caption{Alternatives to SFI}
  \label{fig:sfi-alternatives}
\end{figure}

\vspace{-0.25em}
As shown in Fig.~\ref{fig:sfi-alternatives}, while some SFI variants show comparable performance in (a) and (c), the state dimensions weighted by SFI consistently lead to the highest overall performance. When variants include easily influenced features such as ally position, agents tend to exploit these trivial dimensions, leading to reward hacking and suboptimal behavior. These findings underscore the effectiveness of FIM’s entropy-based weighting, which highlights stable and causally meaningful under-explored dimensions.
\vspace{-1em}

\begin{figure}[!h]
  \centering
  \begin{subfigure}[t]{0.24\linewidth}
    \centering
    \includegraphics[width=\linewidth]{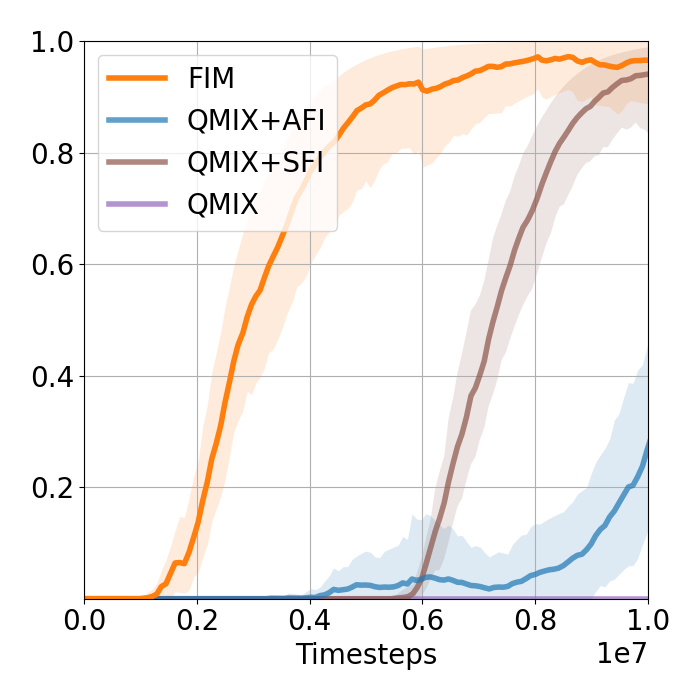}
    \vspace{-2em}
    \caption{\texttt{3s\_vs\_5z}}
  \end{subfigure}
  \hfill
  \begin{subfigure}[t]{0.24\linewidth}
    \centering
    \includegraphics[width=\linewidth]{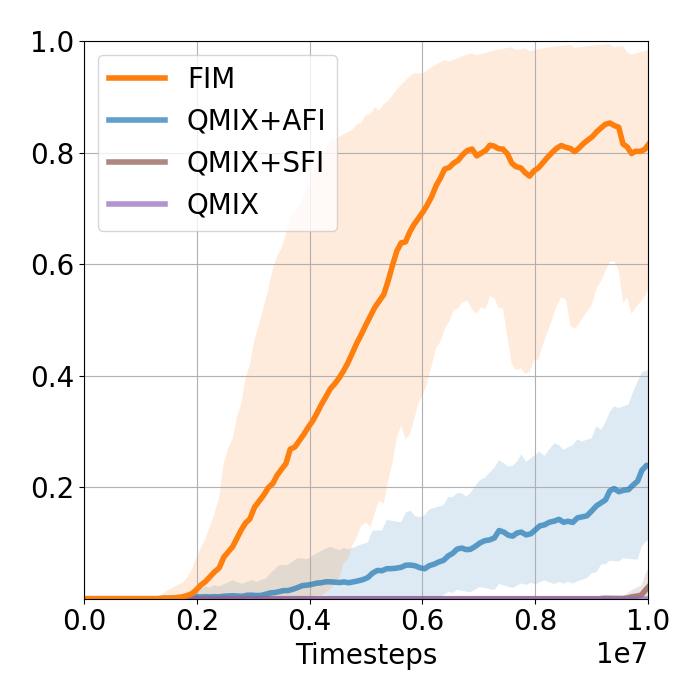}
    \vspace{-2em}
    \caption{\texttt{3s5z\_vs\_3s6z}}
  \end{subfigure}
  \hfill
  \begin{subfigure}[t]{0.24\linewidth}
    \centering
    \includegraphics[width=\linewidth]{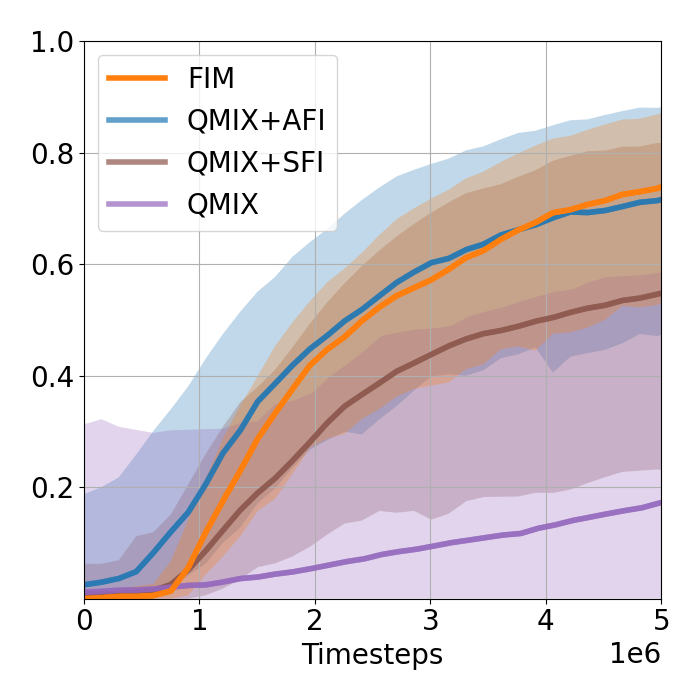}
    \vspace{-2em}
    \caption{\texttt{academy\_3\_vs\_2}}
  \end{subfigure}
  \hfill
  \begin{subfigure}[t]{0.24\linewidth}
    \centering
    \includegraphics[width=\linewidth]{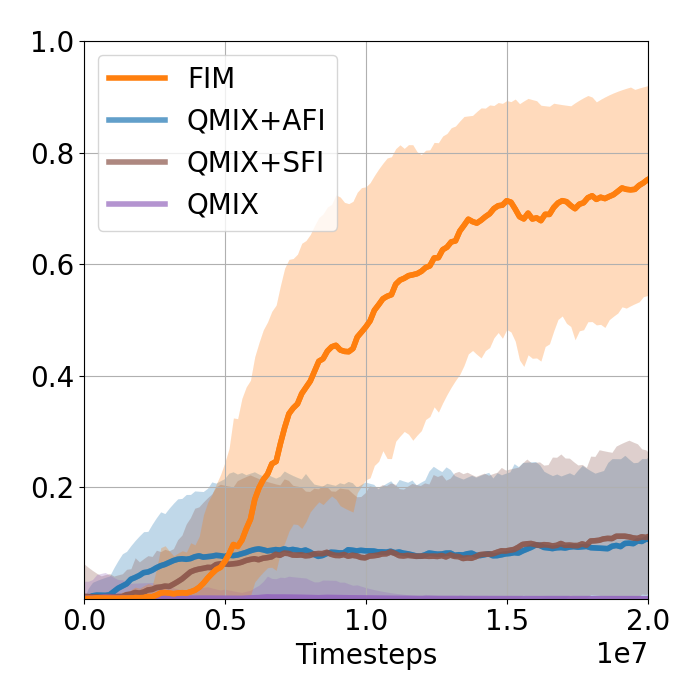}
    \vspace{-2em}
    \caption{\texttt{academy\_3\_vs\_2\_fu} \texttt{ll\_field}}
  \end{subfigure}
  \vspace{-1em}
  \caption{Component evaluation}
  \label{fig:comp-eval}
\end{figure}

\paragraph{Component Evaluation:} Fig.~\ref{fig:comp-eval} compares four variants: vanilla QMIX, QMIX with state focusing influence (SFI), QMIX with agent focusing influence (AFI), and the full FIM framework that integrates both components. SFI improves performance by narrowing exploration to under-explored state dimensions, while AFI promotes deeper exploration by sustaining coordinated influence toward shared targets. Although each component contributes independently, only their combination yields the highest success rates.

\vspace{-1em}

\begin{figure}[!h]
  \centering
  \begin{subfigure}[t]{0.24\linewidth}
    \centering
    \includegraphics[width=\linewidth]{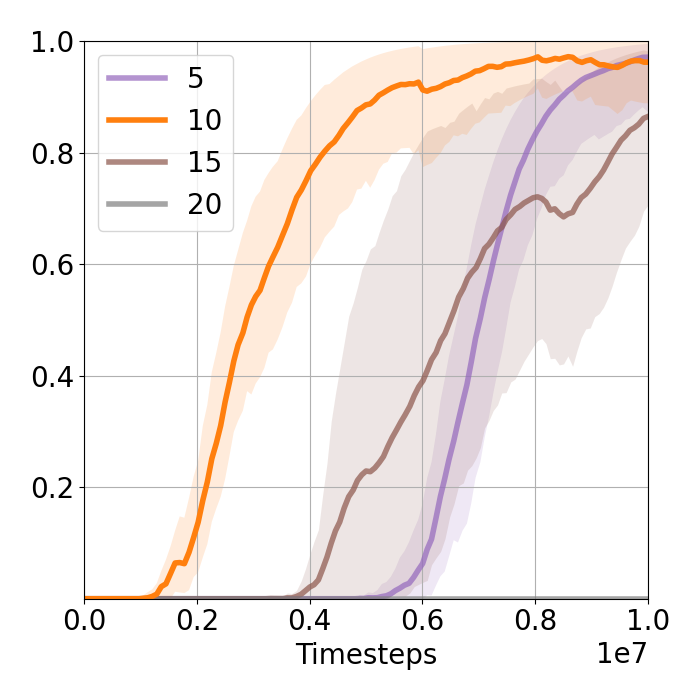}
    \vspace{-2em}
    \caption{\texttt{3s\_vs\_5z}}
  \end{subfigure}
  \hfill
  \begin{subfigure}[t]{0.24\linewidth}
    \centering
    \includegraphics[width=\linewidth]{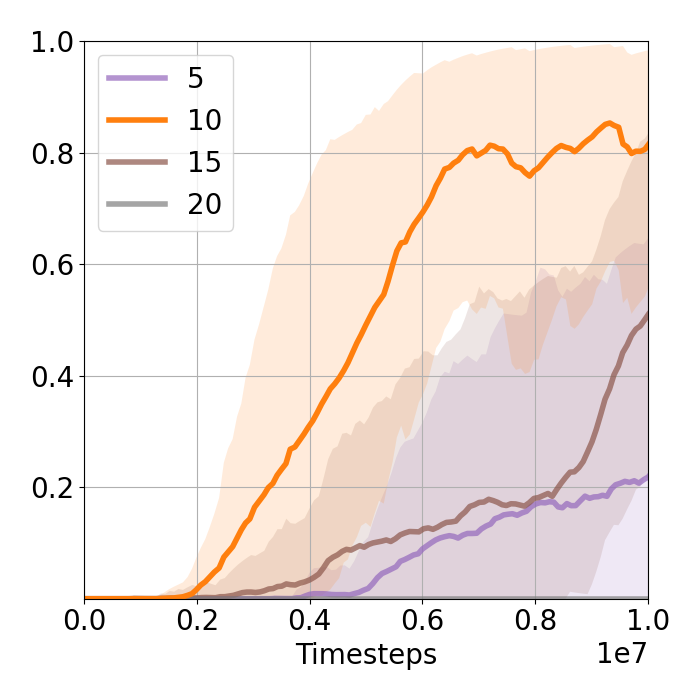}
    \vspace{-2em}
    \caption{\texttt{3s5z\_vs\_3s6z}}
  \end{subfigure}
  \hfill
  \begin{subfigure}[t]{0.24\linewidth}
    \centering
    \includegraphics[width=\linewidth]{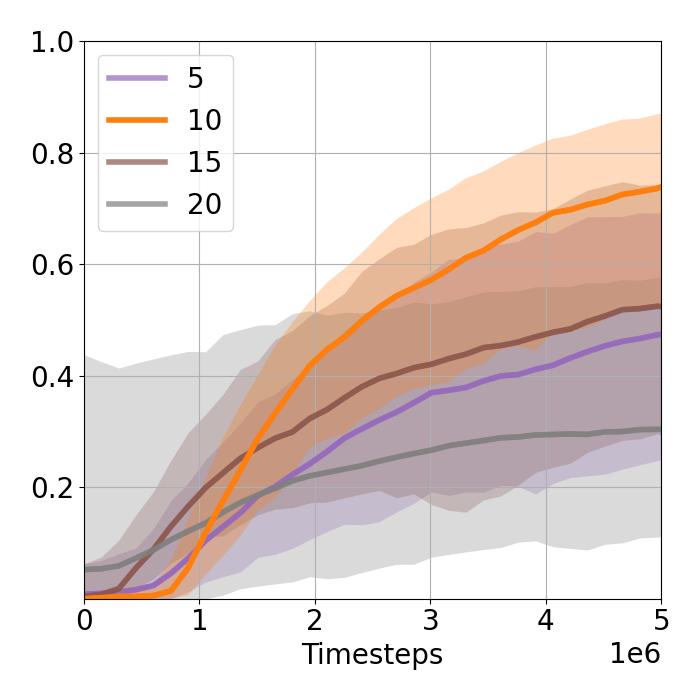}
    \vspace{-2em}
    \caption{\texttt{academy\_3\_vs\_2}}
  \end{subfigure}
  \hfill
  \begin{subfigure}[t]{0.24\linewidth}
    \centering
    \includegraphics[width=\linewidth]{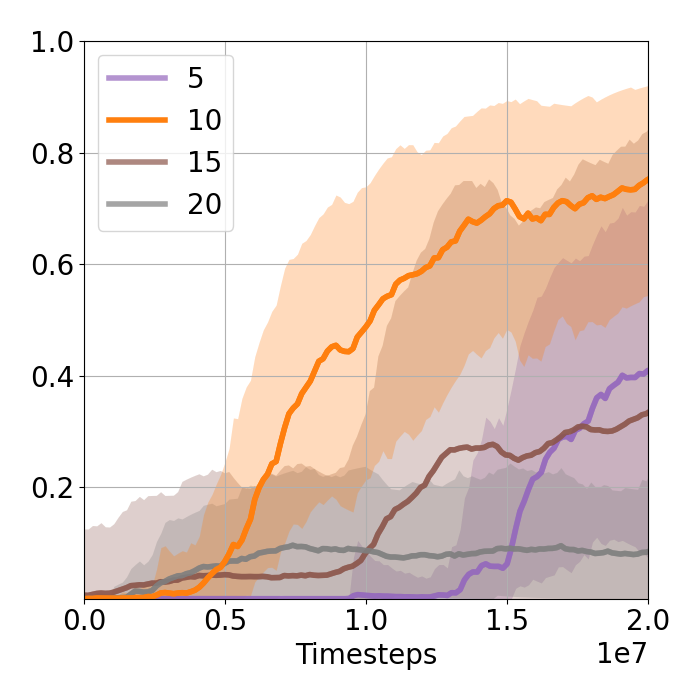}
    \vspace{-2em}
    \caption{\texttt{academy\_3\_vs\_2\_fu} \texttt{ll\_field}}
  \end{subfigure}
  \vspace{-1em}
  \caption{Reward scaling factor $\alpha$}
  \label{fig:alpha}
\end{figure}

\newpage
\vspace{-0.5em}
\paragraph{Reward scaling factor $\alpha$:} We examine how the reward scaling factor $\alpha$ affects performance by testing values in $\alpha \in \{5, 10, 15, 20\}$, as shown in Fig.~\ref{fig:alpha}. When $\alpha$ is too small, the intrinsic reward signal becomes negligible, preventing agents from effectively learning the influence strategy promoted by FIM. Conversely, setting $\alpha$ too large causes agents to over-prioritize intrinsic rewards, ignoring critical environmental feedback. Thus, we select $\alpha=10$ as default value that is well aligned with the extrinsic reward scale, guiding exploration without overwhelming the task objective.

\vspace{-1em}

\begin{figure}[!h]
  \centering
  \begin{subfigure}[t]{0.24\linewidth}
    \centering
    \includegraphics[width=\linewidth]{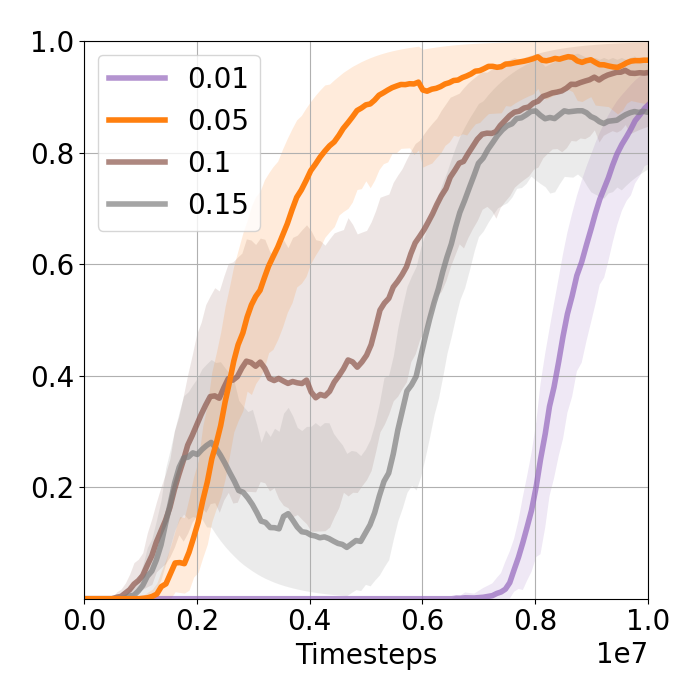}
    \vspace{-2em}
    \caption{\texttt{3s\_vs\_5z}}
  \end{subfigure}
  \hfill
  \begin{subfigure}[t]{0.24\linewidth}
    \centering
    \includegraphics[width=\linewidth]{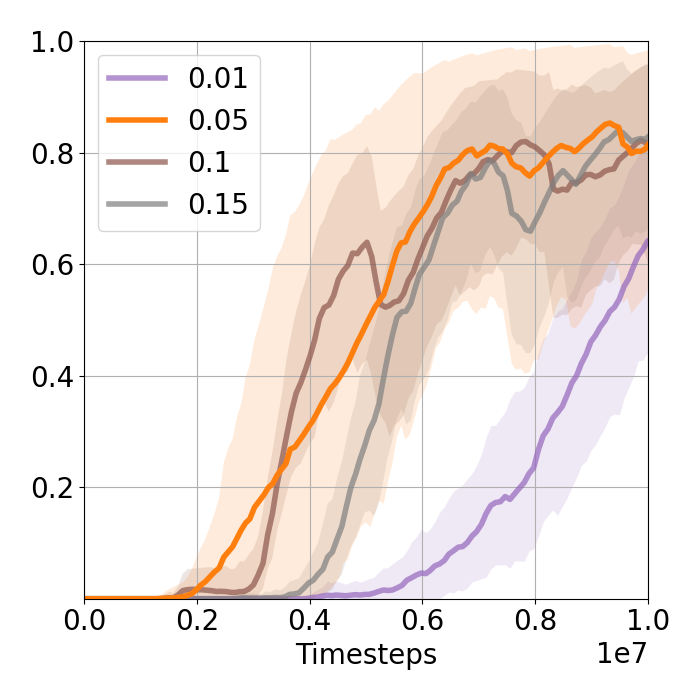}
    \vspace{-2em}
    \caption{\texttt{3s5z\_vs\_3s6z}}
  \end{subfigure}
  \hfill
  \begin{subfigure}[t]{0.24\linewidth}
    \centering
    \includegraphics[width=\linewidth]{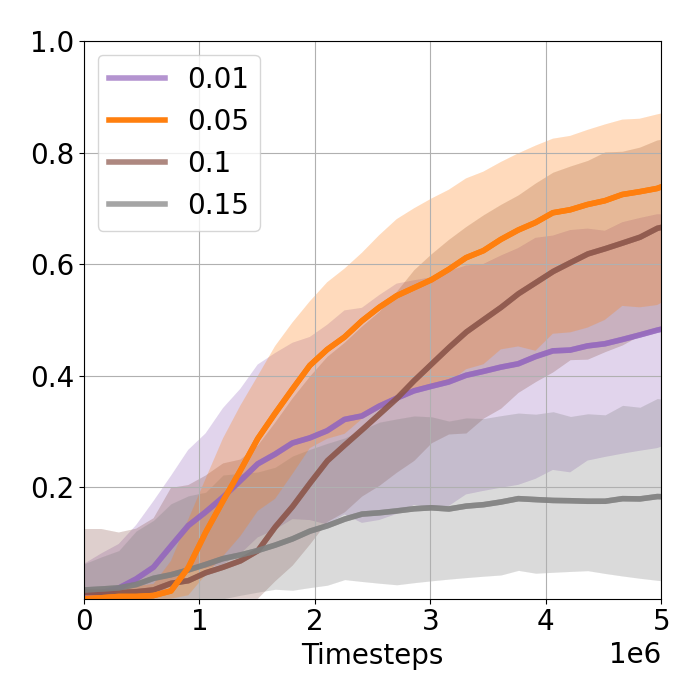}
    \vspace{-2em}
    \caption{\texttt{academy\_3\_vs\_2}}
  \end{subfigure}
  \hfill
  \begin{subfigure}[t]{0.24\linewidth}
    \centering
    \includegraphics[width=\linewidth]{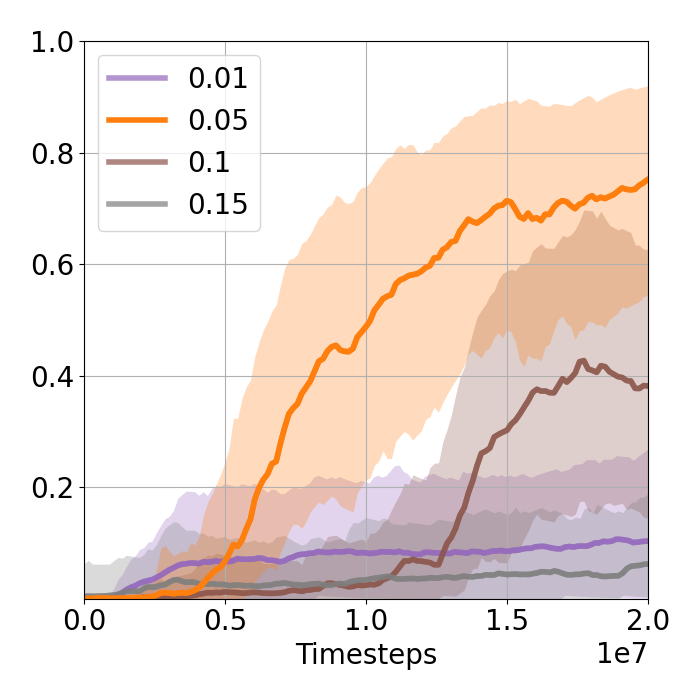}
    \vspace{-2em}
    \caption{\texttt{academy\_3\_vs\_2\_fu} \texttt{ll\_field}}
  \end{subfigure}
  \vspace{-1em}
  \caption{Update rate $\phi$}
  \label{fig:phi}
\end{figure}

\paragraph{Update rate $\phi$:} We examine how the under-explore update rate $\phi$ affects performance by testing values in $\phi \in \{0.01, 0.05, 0.1, 0.15\}$, as shown in Fig.~\ref{fig:phi}. When $\phi$ is too small, the dimension weights $w_d$ adapt too slowly to the evolving behavior policy, failing to reflect which dimensions are currently under-explored and thus delaying effective coordination. Conversely, setting $\phi$ too large causes frequent weight shifts that destabilize the focus target, preventing agents from sustaining coordinated influence on any single dimension. Thus, we select $\phi = 0.05$ as the default, which strikes the best balance between stable and responsive adaptation to the current policy.

\vspace{-1em}

\newpage
\subsection{Analysis of the Learned Dynamics Model}
\label{subsecapp:dynamics-model}
Since the intrinsic reward in FIM is computed from the predictions of the learned dynamics model $\hat{s}$, its accuracy directly influences the reward signal. While a high mean-squared error (MSE) might seem detrimental, our results suggest that prediction inaccuracies can also serve a constructive role by implicitly encouraging exploration of regions with complex or less predictable dynamics. In this sense, model error may act as a form of curiosity, resonating with ideas from curiosity-driven exploration in model-based RL~\cite{pathak2017curiosity}.

\begin{figure}[!h]
  \centering
  \begin{subfigure}[t]{0.35\linewidth}
    \centering
    \includegraphics[width=0.8\linewidth]{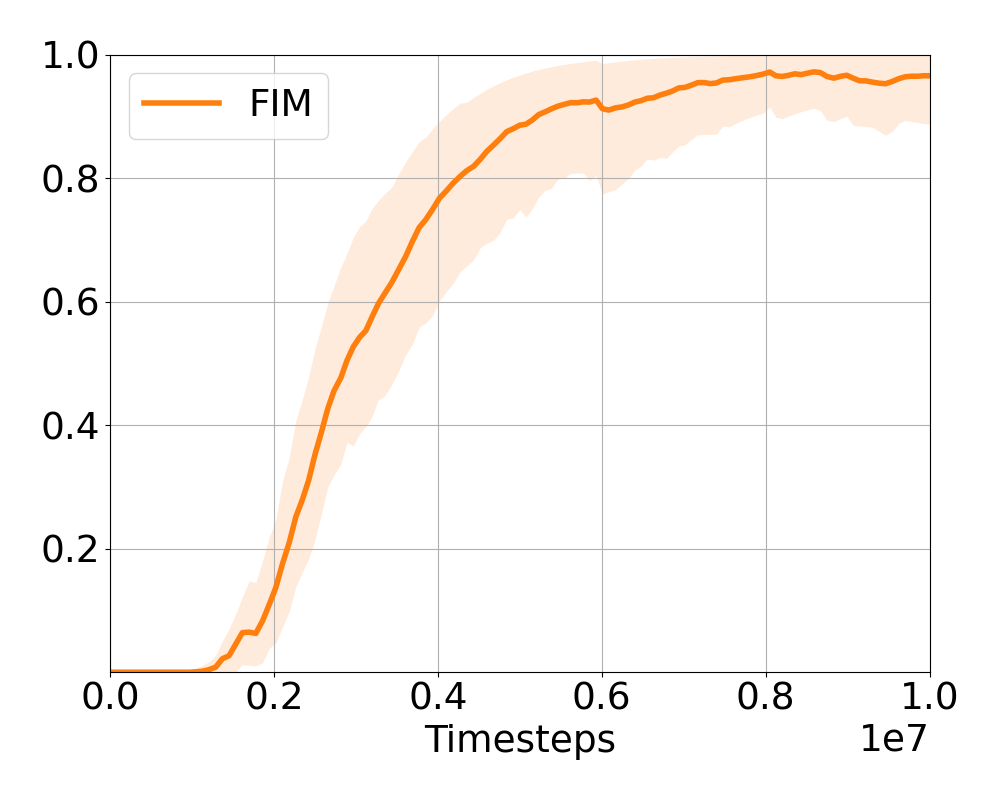}
    \vspace{-0.7em}
    \caption{FIM Performance}
  \end{subfigure}
  \hspace{1.7em}
  \begin{subfigure}[t]{0.35\linewidth}
    \centering
    \includegraphics[width=0.8\linewidth]{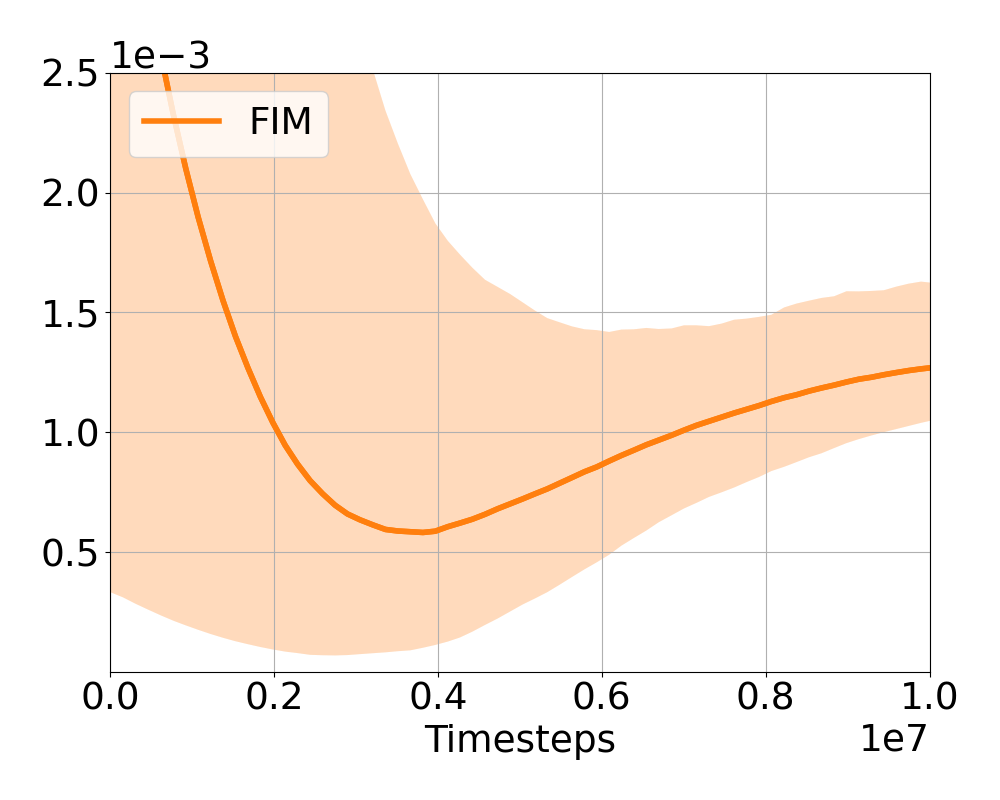}
    \vspace{-0.7em}
    \caption{Mean squared error loss of $\hat{s}$}
  \end{subfigure}
  \caption{Comparison of FIM performance and mean squared error loss of $\hat{s}$ in \texttt{3s\_vs\_5z}}
  \label{fig:comp-pred-mse}
\end{figure}

To examine this effect empirically, we analyzed the SMAC \texttt{3s\_vs\_5z} scenario. As shown in Fig.~\ref{fig:comp-pred-mse}, the forward model’s MSE gradually increased during training, likely reflecting exposure to more diverse transitions. Notably, this trend coincided with a steady improvement in win rate, suggesting that moderate prediction error did not destabilize learning but rather correlated with productive exploration, ultimately supporting performance gains.

\subsection{Comparison of Computational Complexity}
\vspace{-0.5em}
\label{subsecapp:complexity}

FIM computes intrinsic rewards by estimating each agent’s influence through counterfactual marginalization over its action set $\mathcal{A}$ for every dimension in $\mathcal{D}$. This results in a space complexity of $O(|\mathcal{N}| \cdot |\mathcal{A}| \cdot |\mathcal{D}|)$ per timestep, while the time complexity remains $O(1)$ due to GPU parallelization. FIM uses a lightweight three-layer multilayer perceptron (MLP) as the forward transition model and does not alter the main Q-network architecture, keeping computational overhead minimal. We compare FIM against QMIX with dense rewards (QMIX-DR), since sparse-reward QMIX often converges to tie-seeking behaviors that avoid conflict~\cite{liu2023lazy}, resulting in minimal policy updates and unrealistically low computational cost. As shown in Table~\ref{tab:computation}, FIM's average computation time per 1 million timesteps is comparable to QMIX-DR. In \texttt{3s5z\_vs\_3s6z}, FIM also requires fewer timesteps to reach a 60\% success rate, demonstrating strong efficiency. These results emphasize FIM’s ability to enhance agent behavior without compromising computational cost.

\begin{table}[!h]
    \caption{Average computation time (in minutes) per 1 million timesteps (top row) and the number of timesteps (in millions) required to reach a 60\% success rate (bottom row).}
    \vspace{1em}
    \centering
    \begin{tabular}{@{}lll@{}}
        \toprule
        \textbf{Scenario} & \textbf{FIM} & \textbf{QMIX-DR} \\
        \midrule
        \multirow{2}{*}{\texttt{3s\_vs\_5z}} 
        & 72.40 min & 70.65 min \\
        & 3.24M & 4.21M \\
        \midrule
        \multirow{2}{*}{\texttt{3s5z\_vs\_3s6z}} 
        & 126.43 min & 123.23 min \\
        & 5.32M & 13.05M \\
        \midrule
        \\[0.5em]
    \end{tabular}
    \vspace{-1em}
    \label{tab:computation}
\end{table}

%%%%%%%%%%%%%%%%%%%%%%%%%%%%%%%%%%%%%%%%%%%%%%%%%%%%%%%%%%%%

\newpage
\section*{NeurIPS Paper Checklist}

\begin{enumerate}

\item {\bf Claims}
    \item[] Question: Do the main claims made in the abstract and introduction accurately reflect the paper's contributions and scope?
    \item[] Answer: 
    \answerYes{} 
    \item[] Justification: The abstract and introduction (Section~\ref{sec:introduction}) clearly and accurately reflect the paper’s main contributions and scope.
    \item[] Guidelines:
    \begin{itemize}
        \item The answer \answerNA{} means that the abstract and introduction do not include the claims made in the paper.
        \item The abstract and/or introduction should clearly state the claims made, including the contributions made in the paper and important assumptions and limitations. A \answerNo{} or \answerNA{} answer to this question will not be perceived well by the reviewers. 
        \item The claims made should match theoretical and experimental results, and reflect how much the results can be expected to generalize to other settings. 
        \item It is fine to include aspirational goals as motivation as long as it is clear that these goals are not attained by the paper. 
    \end{itemize}

\item {\bf Limitations}
    \item[] Question: Does the paper discuss the limitations of the work performed by the authors?
    \item[] Answer: \answerYes{} % Replace by \answerYes{}, \answerNo{}, or \answerNA{}.
    \item[] Justification: Justification: Section~\ref{sec:limitation} discusses two limitations of FIM: additional computational overhead from training the dynamics model, and limited verification for image-based environments.
    \item[] Guidelines:
    \begin{itemize}
        \item The answer \answerNA{} means that the paper has no limitation while the answer \answerNo{} means that the paper has limitations, but those are not discussed in the paper. 
        \item The authors are encouraged to create a separate ``Limitations'' section in their paper.
        \item The paper should point out any strong assumptions and how robust the results are to violations of these assumptions (e.g., independence assumptions, noiseless settings, model well-specification, asymptotic approximations only holding locally). The authors should reflect on how these assumptions might be violated in practice and what the implications would be.
        \item The authors should reflect on the scope of the claims made, e.g., if the approach was only tested on a few datasets or with a few runs. In general, empirical results often depend on implicit assumptions, which should be articulated.
        \item The authors should reflect on the factors that influence the performance of the approach. For example, a facial recognition algorithm may perform poorly when image resolution is low or images are taken in low lighting. Or a speech-to-text system might not be used reliably to provide closed captions for online lectures because it fails to handle technical jargon.
        \item The authors should discuss the computational efficiency of the proposed algorithms and how they scale with dataset size.
        \item If applicable, the authors should discuss possible limitations of their approach to address problems of privacy and fairness.
        \item While the authors might fear that complete honesty about limitations might be used by reviewers as grounds for rejection, a worse outcome might be that reviewers discover limitations that aren't acknowledged in the paper. The authors should use their best judgment and recognize that individual actions in favor of transparency play an important role in developing norms that preserve the integrity of the community. Reviewers will be specifically instructed to not penalize honesty concerning limitations.
    \end{itemize}

\item {\bf Theory assumptions and proofs}
    \item[] Question: For each theoretical result, does the paper provide the full set of assumptions and a complete (and correct) proof?
    \item[] Answer: \answerYes{} % Replace by \answerYes{}, \answerNo{}, or \answerNA{}.
    \item[] Justification: For the theoretical result, we state the assumption and provide complete proof in Appendix~\ref{secapp:proof}.
    \item[] Guidelines:
    \begin{itemize}
        \item The answer \answerNA{} means that the paper does not include theoretical results. 
        \item All the theorems, formulas, and proofs in the paper should be numbered and cross-referenced.
        \item All assumptions should be clearly stated or referenced in the statement of any theorems.
        \item The proofs can either appear in the main paper or the supplemental material, but if they appear in the supplemental material, the authors are encouraged to provide a short proof sketch to provide intuition. 
        \item Inversely, any informal proof provided in the core of the paper should be complemented by formal proofs provided in appendix or supplemental material.
        \item Theorems and Lemmas that the proof relies upon should be properly referenced. 
    \end{itemize}

    \item {\bf Experimental result reproducibility}
    \item[] Question: Does the paper fully disclose all the information needed to reproduce the main experimental results of the paper to the extent that it affects the main claims and/or conclusions of the paper (regardless of whether the code and data are provided or not)?
    \item[] Answer: \answerYes{}.
    \item[] Justification: The paper ensures reproducibility by providing the full code along with detailed hyperparameter settings in Table~\ref{tab:hyperparams}.
    \item[] Guidelines:
    \begin{itemize}
        \item The answer \answerNA{} means that the paper does not include experiments.
        \item If the paper includes experiments, a \answerNo{} answer to this question will not be perceived well by the reviewers: Making the paper reproducible is important, regardless of whether the code and data are provided or not.
        \item If the contribution is a dataset and\slash or model, the authors should describe the steps taken to make their results reproducible or verifiable. 
        \item Depending on the contribution, reproducibility can be accomplished in various ways. For example, if the contribution is a novel architecture, describing the architecture fully might suffice, or if the contribution is a specific model and empirical evaluation, it may be necessary to either make it possible for others to replicate the model with the same dataset, or provide access to the model. In general. releasing code and data is often one good way to accomplish this, but reproducibility can also be provided via detailed instructions for how to replicate the results, access to a hosted model (e.g., in the case of a large language model), releasing of a model checkpoint, or other means that are appropriate to the research performed.
        \item While NeurIPS does not require releasing code, the conference does require all submissions to provide some reasonable avenue for reproducibility, which may depend on the nature of the contribution. For example
        \begin{enumerate}
            \item If the contribution is primarily a new algorithm, the paper should make it clear how to reproduce that algorithm.
            \item If the contribution is primarily a new model architecture, the paper should describe the architecture clearly and fully.
            \item If the contribution is a new model (e.g., a large language model), then there should either be a way to access this model for reproducing the results or a way to reproduce the model (e.g., with an open-source dataset or instructions for how to construct the dataset).
            \item We recognize that reproducibility may be tricky in some cases, in which case authors are welcome to describe the particular way they provide for reproducibility. In the case of closed-source models, it may be that access to the model is limited in some way (e.g., to registered users), but it should be possible for other researchers to have some path to reproducing or verifying the results.
        \end{enumerate}
    \end{itemize}

\item {\bf Open access to data and code}
    \item[] Question: Does the paper provide open access to the data and code, with sufficient instructions to faithfully reproduce the main experimental results, as described in supplemental material?
    \item[] Answer: \answerYes{}
    \item[] Justification: The code is openly accessible with clear instructions for reproducing the main experimental results.
    \item[] Guidelines:
    \begin{itemize}
        \item The answer \answerNA{} means that paper does not include experiments requiring code.
        \item Please see the NeurIPS code and data submission guidelines (\url{https://neurips.cc/public/guides/CodeSubmissionPolicy}) for more details.
        \item While we encourage the release of code and data, we understand that this might not be possible, so \answerNo{} is an acceptable answer. Papers cannot be rejected simply for not including code, unless this is central to the contribution (e.g., for a new open-source benchmark).
        \item The instructions should contain the exact command and environment needed to run to reproduce the results. See the NeurIPS code and data submission guidelines (\url{https://neurips.cc/public/guides/CodeSubmissionPolicy}) for more details.
        \item The authors should provide instructions on data access and preparation, including how to access the raw data, preprocessed data, intermediate data, and generated data, etc.
        \item The authors should provide scripts to reproduce all experimental results for the new proposed method and baselines. If only a subset of experiments are reproducible, they should state which ones are omitted from the script and why.
        \item At submission time, to preserve anonymity, the authors should release anonymized versions (if applicable).
        \item Providing as much information as possible in supplemental material (appended to the paper) is recommended, but including URLs to data and code is permitted.
    \end{itemize}

\item {\bf Experimental setting/details}
    \item[] Question: Does the paper specify all the training and test details (e.g., data splits, hyperparameters, how they were chosen, type of optimizer) necessary to understand the results?
    \item[] Answer: \answerYes{}
    \item[] Justification: The paper provides all necessary training and testing details, including hyperparameter settings, ablation studies, and optimizer specifications.
    \item[] Guidelines:
    \begin{itemize}
        \item The answer \answerNA{} means that the paper does not include experiments.
        \item The experimental setting should be presented in the core of the paper to a level of detail that is necessary to appreciate the results and make sense of them.
        \item The full details can be provided either with the code, in appendix, or as supplemental material.
    \end{itemize}

\item {\bf Experiment statistical significance}
    \item[] Question: Does the paper report error bars suitably and correctly defined or other appropriate information about the statistical significance of the experiments?
    \item[] Answer: \answerYes{}
    \item[] Justification: All performance plots report the mean across 5 random seeds with shaded areas denoting standard deviation, as stated in Section~\ref{sec:exp}.
    \item[] Guidelines:
    \begin{itemize}
        \item The answer \answerNA{} means that the paper does not include experiments.
        \item The authors should answer \answerYes{} if the results are accompanied by error bars, confidence intervals, or statistical significance tests, at least for the experiments that support the main claims of the paper.
        \item The factors of variability that the error bars are capturing should be clearly stated (for example, train/test split, initialization, random drawing of some parameter, or overall run with given experimental conditions).
        \item The method for calculating the error bars should be explained (closed form formula, call to a library function, bootstrap, etc.)
        \item The assumptions made should be given (e.g., Normally distributed errors).
        \item It should be clear whether the error bar is the standard deviation or the standard error of the mean.
        \item It is OK to report 1-sigma error bars, but one should state it. The authors should preferably report a 2-sigma error bar than state that they have a 96\% CI, if the hypothesis of Normality of errors is not verified.
        \item For asymmetric distributions, the authors should be careful not to show in tables or figures symmetric error bars that would yield results that are out of range (e.g., negative error rates).
        \item If error bars are reported in tables or plots, the authors should explain in the text how they were calculated and reference the corresponding figures or tables in the text.
    \end{itemize}

\item {\bf Experiments compute resources}
    \item[] Question: For each experiment, does the paper provide sufficient information on the computer resources (type of compute workers, memory, time of execution) needed to reproduce the experiments?
    \item[] Answer: \answerYes{}
    \item[] Justification: Section~\ref{secapp:exp-details} specifies the GPU/CPU setup and Section~\ref{subsecapp:complexity} discusses the time of execution.
    \item[] Guidelines:
    \begin{itemize}
        \item The answer \answerNA{} means that the paper does not include experiments.
        \item The paper should indicate the type of compute workers CPU or GPU, internal cluster, or cloud provider, including relevant memory and storage.
        \item The paper should provide the amount of compute required for each of the individual experimental runs as well as estimate the total compute. 
        \item The paper should disclose whether the full research project required more compute than the experiments reported in the paper (e.g., preliminary or failed experiments that didn't make it into the paper). 
    \end{itemize}
    
\item {\bf Code of ethics}
    \item[] Question: Does the research conducted in the paper conform, in every respect, with the NeurIPS Code of Ethics \url{https://neurips.cc/public/EthicsGuidelines}?
    \item[] Answer: \answerYes{}
    \item[] Justification: The research uses standard simulation benchmarks without involving human subjects or sensitive data and adheres to NeurIPS ethical guidelines throughout.
    \item[] Guidelines:
    \begin{itemize}
        \item The answer \answerNA{} means that the authors have not reviewed the NeurIPS Code of Ethics.
        \item If the authors answer \answerNo, they should explain the special circumstances that require a deviation from the Code of Ethics.
        \item The authors should make sure to preserve anonymity (e.g., if there is a special consideration due to laws or regulations in their jurisdiction).
    \end{itemize}

\item {\bf Broader impacts}
    \item[] Question: Does the paper discuss both potential positive societal impacts and negative societal impacts of the work performed?
    \item[] Answer: \answerYes{}
    \item[] Justification: Societal impacts are discussed in Appendix~\ref{secapp:broader-impact} and no negative impacts are foreseen.
    \item[] Guidelines:
    \begin{itemize}
        \item The answer \answerNA{} means that there is no societal impact of the work performed.
        \item If the authors answer \answerNA{} or \answerNo, they should explain why their work has no societal impact or why the paper does not address societal impact.
        \item Examples of negative societal impacts include potential malicious or unintended uses (e.g., disinformation, generating fake profiles, surveillance), fairness considerations (e.g., deployment of technologies that could make decisions that unfairly impact specific groups), privacy considerations, and security considerations.
        \item The conference expects that many papers will be foundational research and not tied to particular applications, let alone deployments. However, if there is a direct path to any negative applications, the authors should point it out. For example, it is legitimate to point out that an improvement in the quality of generative models could be used to generate Deepfakes for disinformation. On the other hand, it is not needed to point out that a generic algorithm for optimizing neural networks could enable people to train models that generate Deepfakes faster.
        \item The authors should consider possible harms that could arise when the technology is being used as intended and functioning correctly, harms that could arise when the technology is being used as intended but gives incorrect results, and harms following from (intentional or unintentional) misuse of the technology.
        \item If there are negative societal impacts, the authors could also discuss possible mitigation strategies (e.g., gated release of models, providing defenses in addition to attacks, mechanisms for monitoring misuse, mechanisms to monitor how a system learns from feedback over time, improving the efficiency and accessibility of ML).
    \end{itemize}
    
\item {\bf Safeguards}
    \item[] Question: Does the paper describe safeguards that have been put in place for responsible release of data or models that have a high risk for misuse (e.g., pre-trained language models, image generators, or scraped datasets)?
    \item[] Answer: \answerNA{}
    \item[] Justification: The paper poses no significant risk of misuse, as it does not release models or datasets with dual-use concerns.
    \item[] Guidelines:
    \begin{itemize}
        \item The answer \answerNA{} means that the paper poses no such risks.
        \item Released models that have a high risk for misuse or dual-use should be released with necessary safeguards to allow for controlled use of the model, for example by requiring that users adhere to usage guidelines or restrictions to access the model or implementing safety filters. 
        \item Datasets that have been scraped from the Internet could pose safety risks. The authors should describe how they avoided releasing unsafe images.
        \item We recognize that providing effective safeguards is challenging, and many papers do not require this, but we encourage authors to take this into account and make a best faith effort.
    \end{itemize}

\item {\bf Licenses for existing assets}
    \item[] Question: Are the creators or original owners of assets (e.g., code, data, models), used in the paper, properly credited and are the license and terms of use explicitly mentioned and properly respected?
    \item[] Answer: \answerYes{}
    \item[] Justification: All used environments and baselines are properly cited in Section~\ref{sec:exp}.
    \item[] Guidelines:
    \begin{itemize}
        \item The answer \answerNA{} means that the paper does not use existing assets.
        \item The authors should cite the original paper that produced the code package or dataset.
        \item The authors should state which version of the asset is used and, if possible, include a URL.
        \item The name of the license (e.g., CC-BY 4.0) should be included for each asset.
        \item For scraped data from a particular source (e.g., website), the copyright and terms of service of that source should be provided.
        \item If assets are released, the license, copyright information, and terms of use in the package should be provided. For popular datasets, \url{paperswithcode.com/datasets} has curated licenses for some datasets. Their licensing guide can help determine the license of a dataset.
        \item For existing datasets that are re-packaged, both the original license and the license of the derived asset (if it has changed) should be provided.
        \item If this information is not available online, the authors are encouraged to reach out to the asset's creators.
    \end{itemize}

\item {\bf New assets}
    \item[] Question: Are new assets introduced in the paper well documented and is the documentation provided alongside the assets?
    \item[] Answer: \answerYes{}
    \item[] Justification: The released code are documented and provided with usage instructions.
    \item[] Guidelines:
    \begin{itemize}
        \item The answer \answerNA{} means that the paper does not release new assets.
        \item Researchers should communicate the details of the dataset\slash code\slash model as part of their submissions via structured templates. This includes details about training, license, limitations, etc. 
        \item The paper should discuss whether and how consent was obtained from people whose asset is used.
        \item At submission time, remember to anonymize your assets (if applicable). You can either create an anonymized URL or include an anonymized zip file.
    \end{itemize}

\item {\bf Crowdsourcing and research with human subjects}
    \item[] Question: For crowdsourcing experiments and research with human subjects, does the paper include the full text of instructions given to participants and screenshots, if applicable, as well as details about compensation (if any)? 
    \item[] Answer: \answerNA{}
    \item[] Justification: The research does not involve any human subjects or crowdsourcing.
    \item[] Guidelines:
    \begin{itemize}
        \item The answer \answerNA{} means that the paper does not involve crowdsourcing nor research with human subjects.
        \item Including this information in the supplemental material is fine, but if the main contribution of the paper involves human subjects, then as much detail as possible should be included in the main paper. 
        \item According to the NeurIPS Code of Ethics, workers involved in data collection, curation, or other labor should be paid at least the minimum wage in the country of the data collector. 
    \end{itemize}

\item {\bf Institutional review board (IRB) approvals or equivalent for research with human subjects}
    \item[] Question: Does the paper describe potential risks incurred by study participants, whether such risks were disclosed to the subjects, and whether Institutional Review Board (IRB) approvals (or an equivalent approval/review based on the requirements of your country or institution) were obtained?
    \item[] Answer: \answerNA{}
    \item[] Justification: The paper does not involve experiments with human participants requiring IRB or equivalent approval.
    \item[] Guidelines:
    \begin{itemize}
        \item The answer \answerNA{} means that the paper does not involve crowdsourcing nor research with human subjects.
        \item Depending on the country in which research is conducted, IRB approval (or equivalent) may be required for any human subjects research. If you obtained IRB approval, you should clearly state this in the paper. 
        \item We recognize that the procedures for this may vary significantly between institutions and locations, and we expect authors to adhere to the NeurIPS Code of Ethics and the guidelines for their institution. 
        \item For initial submissions, do not include any information that would break anonymity (if applicable), such as the institution conducting the review.
    \end{itemize}

\item {\bf Declaration of LLM usage}
    \item[] Question: Does the paper describe the usage of LLMs if it is an important, original, or non-standard component of the core methods in this research? Note that if the LLM is used only for writing, editing, or formatting purposes and does \emph{not} impact the core methodology, scientific rigor, or originality of the research, declaration is not required.
    %this research? 
    \item[] Answer: \answerNA{}
    \item[] Justification: LLMs were not used in the development of the core methodology of this work.
    \item[] Guidelines:
    \begin{itemize}
        \item The answer \answerNA{} means that the core method development in this research does not involve LLMs as any important, original, or non-standard components.
        \item Please refer to our LLM policy in the NeurIPS handbook for what should or should not be described.
    \end{itemize}

\end{enumerate}

\end{document}